\documentclass[runningheads]{llncs}

\usepackage{eccv}

\usepackage{eccvabbrv}

\usepackage{graphicx}
\usepackage{booktabs}
\usepackage{tabularx}
\usepackage{multirow}
\usepackage{array}
\usepackage{hyperref}
\usepackage{lipsum}
\usepackage{enumitem}

\usepackage{mdframed} 
\usepackage{xcolor}   
\usepackage[accsupp]{axessibility}  

\usepackage{orcidlink}
\usepackage{tikz}
\usepackage{xcolor}

\newcommand{\textnoindent}[1]{\noindent\textbf{#1.}}

\begin{document}
\title{Physics-Free Spectrally Multiplexed Photometric Stereo under Unknown Spectral Composition} 

\titlerunning{SpectraM-PS}

\author{Satoshi Ikehata\inst{1,2}\orcidlink{0000-0002-6061-7956} \and
Yuta Asano\inst{1}\orcidlink{0000-0002-0898-7802}}

\authorrunning{S. Ikehata and Y. Asano}

\institute{National Institute of Informatics, Tokyo, Japan \and
Tokyo Institute of Technology, Tokyo, Japan}

\maketitle
\begin{abstract}
  In this paper, we present a groundbreaking spectrally multiplexed photometric stereo approach for recovering surface normals of dynamic surfaces without the need for calibrated lighting or sensors, a notable advancement in the field traditionally hindered by stringent prerequisites and spectral ambiguity. By embracing spectral ambiguity as an advantage, our technique enables the generation of training data without specialized multispectral rendering frameworks. We introduce a unique, physics-free network architecture, SpectraM-PS, that effectively processes multiplexed images to determine surface normals across a wide range of conditions and material types, without relying on specific physically-based knowledge. Additionally, we establish the first benchmark dataset, SpectraM14, for spectrally multiplexed photometric stereo, facilitating comprehensive evaluations against existing calibrated methods. Our contributions significantly enhance the capabilities for dynamic surface recovery, particularly in uncalibrated setups, marking a pivotal step forward in the application of photometric stereo across various domains.
\end{abstract}

\begin{figure}[!h]
	\begin{center}
		\includegraphics[width=120mm]{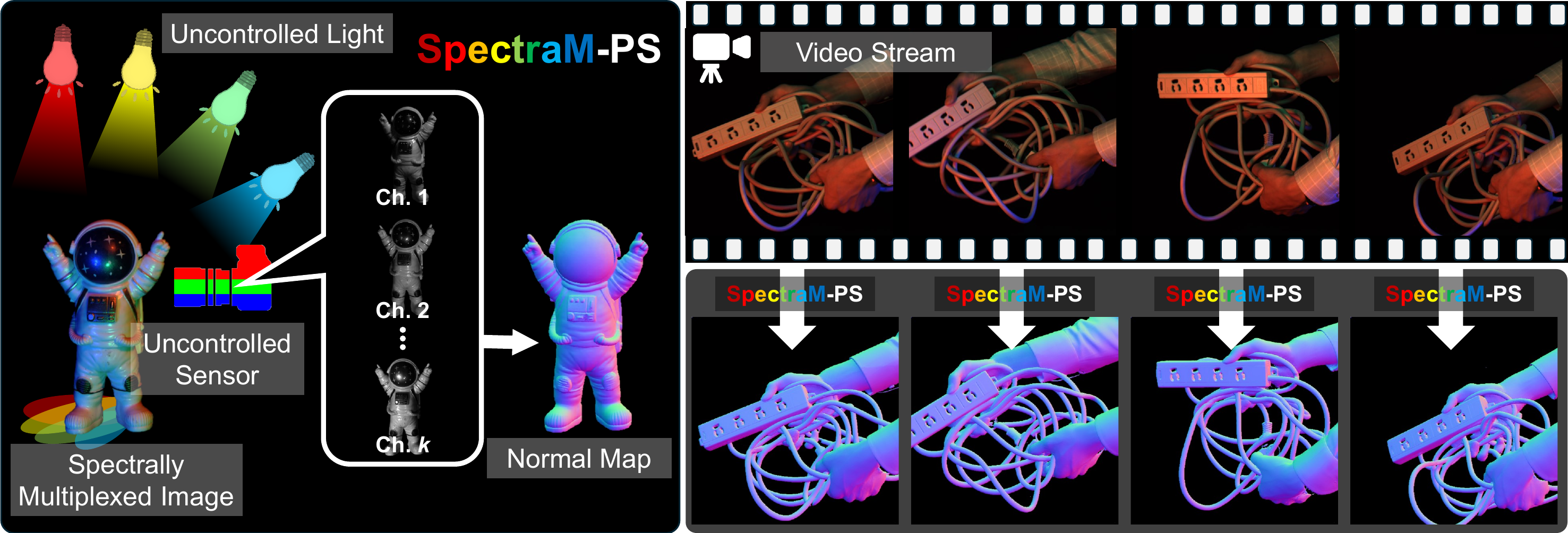}
	\end{center}
	\vspace{-10pt}
	\caption{(Left) Illustration of our SpectraM-PS. Our method recovers a surface normal map from a spectrally multiplexed image. The spectral/spatial composition for generating the observations is unknown. There is potential for a mismatch between the sensor's spectral sensitivity and the light source's spectral distribution, which may lead to crosstalk. (Right) By applying our method to individual frames of a video, the normal map of dynamic surfaces can be recovered.}
	\vspace{-13pt}
	\label{fig:teaser}
\end{figure}
\section{Introduction}
\label{sec:intro}
Recovering detailed normals of dynamic surfaces is essential for monitoring various processes: in manufacturing, it helps in tracking wear and tear of machine parts; in agriculture, it allows for the observation of crop growth through changes in leaf geometry; and in sports engineering, it aids in improving equipment design and safety by analyzing how surfaces deform upon impact.

Photometric Stereo (PS)~\cite{Woodham1980,Silver1980} derives object surface normals from observations under different lighting conditions at a fixed viewpoint. Despite decades of progress, the requirement for objects to stay stationary during lighting changes challenges the recovery of dynamic surfaces, essential for analyzing temporal surface deformations. PS researches have employed {\it spectral multiplexing} for dynamic surface recovery~\cite{Kontsevich1994,Hernandez2007a,Vogiatzis2012,Chakrabarti2016,Ozawa2018,Ju2020}—a technique originally used in telecommunications and spectroscopy~\cite{ishio1984}. This technique utilizes the varying wavelengths of light to multiplex and subsequently demultiplex signals within a single sensor, thereby increasing the capacity for information transmission.


Historically, spectrally multiplexed photometric stereo is often referred to as color photometric stereo~\cite{Hernandez2007a,Decker2009,Anderson2011,Ju2018,Ju2020}, specifically when objects are illuminated with monochromatic red, green, and blue lights from {\it various angles}, captured in the camera's RGB channels. Each channel is then treated as an observation under a distinct lighting for PS analysis. This technique has been further extended to not only RGB but also any number of spectral bands and is specifically referred to as multispectral photometric stereo~\cite{Guo2021a,Guo2022,Lv2023}. These techniques enable dynamic surface recovery by processing each temporal multi-channel frame separately.  

Despite their potential in dynamic surface recovery, current spectrally multiplexed photometric stereo methods face stringent prerequisites that limit their practicality. These include the necessity for precisely calibrated directional lighting in controlled environments~\cite{Hernandez2007a,Decker2009,Anderson2011,Guo2022,Lv2023} and sensors with aligned spectral sensitivities~\cite{Ju2018,Ju2020}. Furthermore, they make strong assumptions about the surface, requiring it to be convex, integrable, Lambertian, and exhibit uniform chromaticity~\cite{Kontsevich1994,Hernandez2007a}. By contrast, recent PS methods without spectral multiplexing support non-Lambertian surfaces~\cite{Ikehata2014a,Shi2012a}, spatially-varying materials~\cite{Ikehata2018,Chen2018}, and the use of uncalibrated lighting~\cite{Shi2010,Chen2019,Ikehata2022,Ikehata2023}. This disparity arises from the challenge where identical observations are produced by different spectral compositions of light, surface and sensor~\cite{Nayatani1972,Hill1999}, a phenomenon absent in conventional PS due to constant spectral compositions of them across images. Recently, Guo~\etal~\cite{Guo2021a,Guo2022} thoroughly explored how the spectral ambiguity renders spectrally multiplexed photometric stereo ill-posed, necessitating severely unpractical conditions on light, surface, and sensor to resolve the ambiguity. 

In this work, we propose a spectrally multiplexed photometric stereo method that recovers normals directly from multiplexed observations produced by \textit{unknown composition} of lights, surface, and sensor (See~\cref{fig:teaser}-left), drawing inspiration from recent data-driven photometric stereo methods~\cite{Ikehata2022,Ikehata2023}. While prior works~\cite{Guo2021a,Lv2023} have considered the spectral ambiguity harmful and something that must be resolved, we demonstrate that it can even be beneficial for a data-driven approach as it compacts the input space and allows for the generation of training data without a multispectral rendering framework. Trained on spectrally composed observations, our generic, physics-free architecture directly maps a single multiplexed image with an order-agnostic, arbitrary number of channels to object surface normals without the need for calibrating lights and sensors, and without imposing severe constraints on surface reflectance and geometry. By applying our method to individual frames of a video, dynamic surface recovery via spectrally multiplexed photometric stereo in uncalibrated, uncontrolled scenarios is achieved as illustrated in~\cref{fig:teaser} (right).

While numerous benchmarks exist for conventional photometric stereo~\cite{Shi2016, Ren2022, Wang2023}, not a single benchmark is available for spectrally multiplexed PS. Therefore, we have created the first real benchmark dataset, namely {\it SpectraM}, for this task. For comparative evaluations with calibrated methods such as~\cite{Guo2021a, Lv2023}, we carefully calibrated directional light sources of different wavelengths, including their directions. We implemented five different difficulty settings by varying the type of light sources (RGB vs NIR) and whether individual light sources were illuminated independently or simultaneously, catering to both ideal conditions without channel crosstalk and more realistic conditions with channel crosstalk.

Our contributions are summarized as follows: (1) We pioneer the use of spectrally multiplexed photometric stereo for recovering dynamic surfaces in uncalibrated setups, employing a data-driven approach to overcome spectral ambiguity, a significant barrier in prior work. (2) We introduce a unique, physics-free neural network, {\it SpectraM-PS} (\textbf{Spectra}lly \textbf{M}ultiplexed \textbf{PS}), that recovers surface normals from a spectrally multiplexed image, capable of handling images with any number of order-agnostic channels. (3) We demonstrate how spectral ambiguity restricts the input space for training data generation, offering a strategy for efficient dataset creation without the need for multispectral rendering. (4) We create the first evaluation benchmark, {\it SpectraM}, for this domain, showing our method's superiority over current calibrated spectrally multiplexed photometric stereo techniques. 
\section{Related Work}
\label{sec:related work}
\textnoindent{Temporally Multiplexed Photometric Stereo (Conventional)} From a communication perspective, conventional photometric stereo, as originally proposed by Woodham~\cite{Woodham1980}, employs a {\it time multiplexing} strategy to recover static surfaces. This method involves temporally varying lighting conditions while capturing images from a fixed viewpoint. Since the same light sources and sensor always provide observations, image differences stem solely from changes in light direction and intensity. This approach simplifies addressing complex conditions such as cast shadows~\cite{Ikehata2012,Ikehata2014b}, non-Lambertian surfaces~\cite{Goldman2005,Ikehata2014a}, non-convex surfaces~\cite{Ikehata2018}, and uncalibrated lighting~\cite{Shi2010,Chen2019,Kaya2021}. Recently, learning-based methods~\cite{Santo2017, Ikehata2018, Li2019, Zheng2019, Logothetis2021, Chen2018, Yakun2021, Yao2020, Liu2021, Ikehata2021, Chen2019, Chen2020, Kaya2021, Tiwari2022, Ikehata2022, Taniai2018, Li2022a, Li2022b} have emerged as an effective alternative, addressing challenges faced by traditional, physics-based approaches~\cite{Wu2010, Ikehata2012, Goldman2005, Hertzmann2005, Hayakawa1994, Alldrin2007a, Feng2013, Basri2007}. These data-driven methods regress normal maps from observations utilizing techniques such as observation map regression~\cite{Ikehata2018,Ikehata2022icip}, set pooling~\cite{Chen2018,Chen2020}, graph neural networks~\cite{Yao2020}, Transformer~\cite{Ikehata2021}, and neural rendering for inverse rendering optimization~\cite{Taniai2018,Li2022a,Li2022b}. Notably, the introduction of universal photometric stereo methods~\cite{Ikehata2022, Ikehata2023} has enabled the handling of unknown, spatially-varying lighting in a purely data-driven framework. Inspired by these advancements, our work aims to regress normals from observations under unknown light, surface, and sensor conditions.

\vspace{3pt}\textnoindent{Spectrally Multiplexed Photometric Stereo} Despite its potential for dynamic surface recovery, spectrally multiplexed photometric  stereo~\cite{Drew1992,Kontsevich1994} has remained less explored than its mainstream counterpart, primarily due to notable limitations.

\textbf{Lighting Constraints:} Existing methods necessitate multiple directional lights in controlled settings, contrasting the flexibility of temporally multiplexed techniques that adapt to diverse lighting conditions~\cite{Mecca2013, Mo2018, Ikehata2023}. They typically require pre-calibrated light directions and distinct light source spectra to prevent channel crosstalk. In contrast, our approach accommodates uncontrolled lighting scenarios without the need for predefined or calibrated setups.

\textbf{Surface and Sensor Constraints:} Prior works assume significant limitations on surfaces, such as Lambertian, convex, and uniform properties~\cite{Drew1992,Kontsevich1994,Hernandez2007a}. Recent advances like Lv~\etal~\cite{Lv2023} extend to non-Lambertian surfaces but still require uniform materials. Sensor requirements typically involve narrow-band spectral responses and a fixed number of channels, limiting flexibility. Our approach leverages a data-driven model, training neural networks on synthetic data to handle complex surfaces and varied spectral sensor responses.

\textbf{Data-driven Methods:} To our knowledge, there are few data-driven methods for this task~\cite{Ju2018, Ju2020, Lv2023}. Previous studies, such as those by Ju \etal~\cite{Ju2018, Ju2020}, require identical spectral and spatial lighting conditions during both training and testing, greatly restricting their practicality. ELIE-Net~\cite{Lv2023} permits variability in training and test setups; however, strong assumptions on both light and surfaces prohibitively limit its applications. Our model, on the other hand, eschews explicit lighting models in favor of learning direct input-output relationships, allowing for accurate predictions under varied and unknown spectral compositions and supporting materials with spatially diverse properties. Furthermore, unlike ELIE-Net's reliance on spectral BRDF datasets, our training approach utilizes assets akin to those employed in conventional photometric stereo.

\section{Problem Statement}
\label{sec:unimps}

Given a single image $I \in \mathbb{R}^{h \times w \times k}$ captured by a static $k$-channel orthographic sensor, along with an optional object mask $M \in \mathbb{R}^{h \times w}$, the objective of spectrally multiplexed photometric stereo is to recover the surface normals of the object, $N \in \mathbb{R}^{h \times w \times 3}$\footnote{It should be noted that unlike conventional PS, reflectance recovery generally falls outside the scope of spectrally multiplexed PS due to its inherently ill-posed nature.}. The object is supposed to be illuminated by multiple light sources, each with unique spatial and spectral properties. Previous studies have typically assumed an equal number of light sources and sensor channels, with each light source's wavelength precisely matching the spectral response of a single channel, thereby precluding any channel crosstalk, and with the directions of lights predetermined. In contrast, we do not presuppose the spatial distribution of illumination nor require the spectrum of each light source to be exclusively aligned with the spectral responses of the sensor channels, thus permitting channel crosstalk. This distinction is elaborated in subsequent sections.

\section{Method}
\label{sec:method}
\begin{figure}[!t]
	\begin{center}
		\includegraphics[width=120mm]{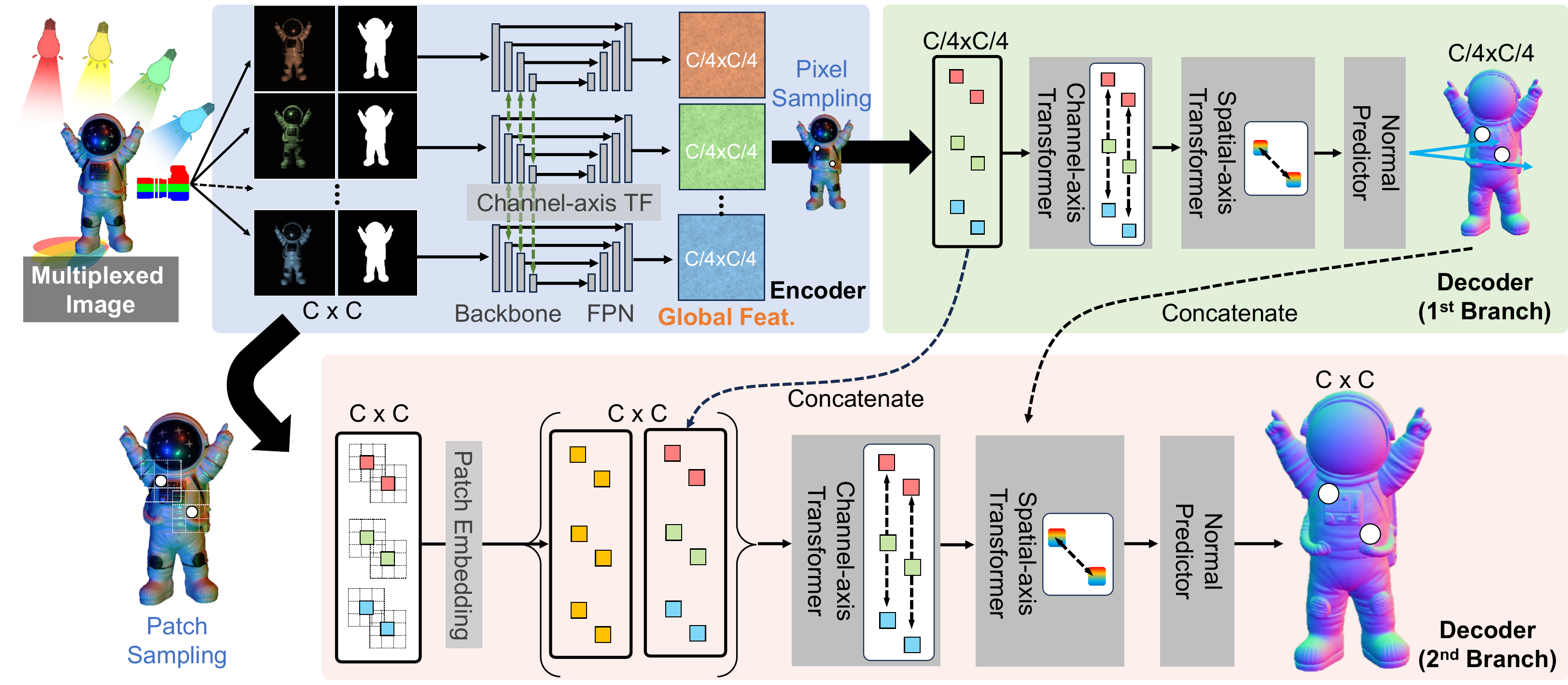}
	\end{center}
	\vspace{-10pt}
	\caption{SpectraM-PS involves decomposing a spectrally multiplexed image into independent channels. The Global Feature Encoder extracts a feature map from each channel. The surface vector is then recovered by the Dual-scale Surface Normal Decoder at each pixel. We adopt a dual-scale approach to preserve the entire shape, while employing patch-embedding techniques to enhance local surface details.}
	\label{fig:architecture}
	\vspace{-5pt}
\end{figure}
We propose and tackle the problem of spectrally multiplexed photometric stereo from a single image with multiple channels, produced by an unknown spectral/spatial composition of the sensor, light, and surface. To build such a method, we train neural networks to directly infer the normal map from an image.

Our method addresses two challenges: (1) a physics-free architecture that accepts a varying number of spectral channels and is agnostic to their order, and (2) an effective approximation of the spectrally multiplexed image for efficient training. We consider (2) to be of significant importance, yet it remains largely unexplored. Synthesizing spectrally multiplexed images in a physically accurate manner is prohibitively challenging, owing to the increased complexity of their parameter spaces and the scarcity of 3D assets with detailed spectral properties, as well as the complex nature of light-surface interactions across different wavelengths. To address this issue, developing an efficient approximation method for rendering spectrally multiplexed images using common RGB image rendering techniques is crucial.

\subsection{Physics-free Spectrally Multiplexed PS Network (SpectraM-PS)}
The architecture of SpectraM-PS is illustrated in~\cref{fig:architecture}. Drawing inspiration from established Transformer-based photometric stereo networks~\cite{Ikehata2021,Ikehata2022,Ikehata2023}, we integrate an encoder to first extract the global features and a decoder to estimate per-pixel surface normals. The architecture derives normals solely from the input image and mask, without prior light information. This indicates that the architecture focuses the network's learning objective on the relationship between input and output without relying on physics-based principles, unlike prior works.

In our model, all the interactions among features from different sensor channels are employed by n\"aive Transformer~\cite{Vaswani2017} in similar to~\cite{Ikehata2021,Ikehata2022,Ikehata2023}. Transformer functions by mapping input features to query, key, and value vectors of equal dimensions. These vectors are processed through a multi-head self-attention mechanism, utilizing a softmax layer, followed by a feed-forward network comprising two linear layers. Both the input and output layers maintain identical dimensionality, with the inner layer having twice the dimension of the input. Each layer is surrounded by a residual connection, succeeded by layer normalization~\cite{Xiong2020}. The advantage of employing Transformers in photometric stereo networks lies in their capability to facilitate complex interactions among intermediate features, a task unachievable with simple operations like pooling~\cite{Chen2018,Chen2019,Chen2020,Lv2023} and observation map~\cite{Ikehata2018,Ikehata2022icip}. Additionally, the token-based attention mechanism allows for different number of input tokens (\ie, sensor channels) between training and test phases and ensures that the results are independent on the order of tokens. 

Building on the established Transformer-based architecture~\cite{Ikehata2023} for temporally multiplexed photometric stereo, we extend its scope to a spectrally multiplexed one. To accommodate a variable number of channels and eliminate dependency on their order, an input spectrally multiplexed image is first {\it split} into individual channels, each of which is concatenated with an object mask (If no mask is provided, replace with a matrix of ones.) and then input into the same encoder of a neural network. This approach is distinctly different from traditional methods that encode an input image as it is in neural networks~\cite{Ju2018,Ju2020}. Then, at \textbf{Preprocessing}, we normalize each channel by dividing it by a random value between its maximum and mean. Each channel and mask are resized or cropped to a resolution ($c\times c$) that is a multiple of $32$ to be input into the multi-scale encoder. \textbf{Global Feature Encoder} first applies a backbone network (\ie, ConvNeXt-T~\cite{convnext}) to individualy encode the concatenation of each channel and mask, then uses Transformer layers for channel-axis (\ie, sensor channel) feature communication across scales (the number of Transformer layers is \{0, 1, 2, 4\} at \{1/4, 1/8, 1/16, 1/32\} scales, hidden dimensions are same with input dimensions), and finally, a feature pyramid network~\cite{Xiao2018} for integrating features at different levels. Note that the design of encoder is almost the same as~\cite{Ikehata2022,Ikehata2023}, except that images are replaced by sensor channels, so details are omitted.

Given global features $\in\mathcal{R}^{k\times c/4 \times c/4\times 256}$, our novel \textbf{Dual-scale Surface Normal Decoder} adopts a dual-scale strategy for predicting point-wise surface normals at $m$ (\ie, 2048) sampled locations at the original resolution within the object mask. The first branch recovers low-frequency surface normals at the feature map resolution ($\frac{c}{4}\times \frac{c}{4}$). Concretely, all global features corresponding to each sample location are processed by five channel-axis Transformer layers (with a 256 hidden dimension) and are pooled via Pooling-by-Multihead-Attention (PMA)\cite{Lee2019} using an additional channel-axis Transformer layer (with a 384 hidden dimension). To enhance spatial communication, two \textit{spatial}-axis Transformer layers (with a 384 hidden dimension) inspired by Ikehata\cite{Ikehata2023} are employed (\ie, Transformer is employed among samples at different locations), with a final MLP (384$\rightarrow$192$\rightarrow$3) predicting the low-frequency normals at sampled locations. The second branch focuses on high-resolution normal recovery, using patch embedding for local context at the same $m$ locations, with $w\times w$ patches ($w=21$) processed by an MLP (with a 256 hidden dimension) and two layer norms. These patches, concatenated with bilinearly interpolated global features, pass through five channel-axis Transformer blocks (with a 256 hidden dimension), PMA (with a 384 hidden dimension), and are merged with the first branch output normals into 387-dimensional vectors. Two additional spatial-axis Transformer layers (with a 384 hidden dimension) enable non-local interactions, culminating in a final MLP (384$\rightarrow$192$\rightarrow$3) for high-resolution normals, normalized to unit vectors. The complete normal map is formed by merging all the vectors from different sample sets. 

It should be noted that while SDM-UniPS~\cite{Ikehata2023} targets temporally multiplexed PS with tens of images, and its decoder performs normal estimation purely on a pixel basis. In contrast, spectrally multiplexed PS deals with fewer channels (\eg, three with RGB sensors), making a pixel-basis architecture less effective. Therefore, we use {\it patch embedding} at the patch-basis decoder to capture fine details with a {\it dual-scale architecture} for preserving overall shape. Without a dual-scale design, the recovery of surface normals becomes overly influenced by local image textures captured through patch embedding. This leads to a failure in preserving the entire shape, resulting in a significant reduction in accuracy. Our motivation is supported by~\cref{fig:dataset} (left), where SDM-UniPS~\cite{Ikehata2023} fails to recover fine details with six temporally multiplexed images, while our architecture produces a more plausible normal map. 
\subsection{Efficient Training Strategy Utilizing Spectral Ambiguity}
\label{sec:training_data}
\begin{figure}[t]
	\begin{center}
        \includegraphics[width=120mm]{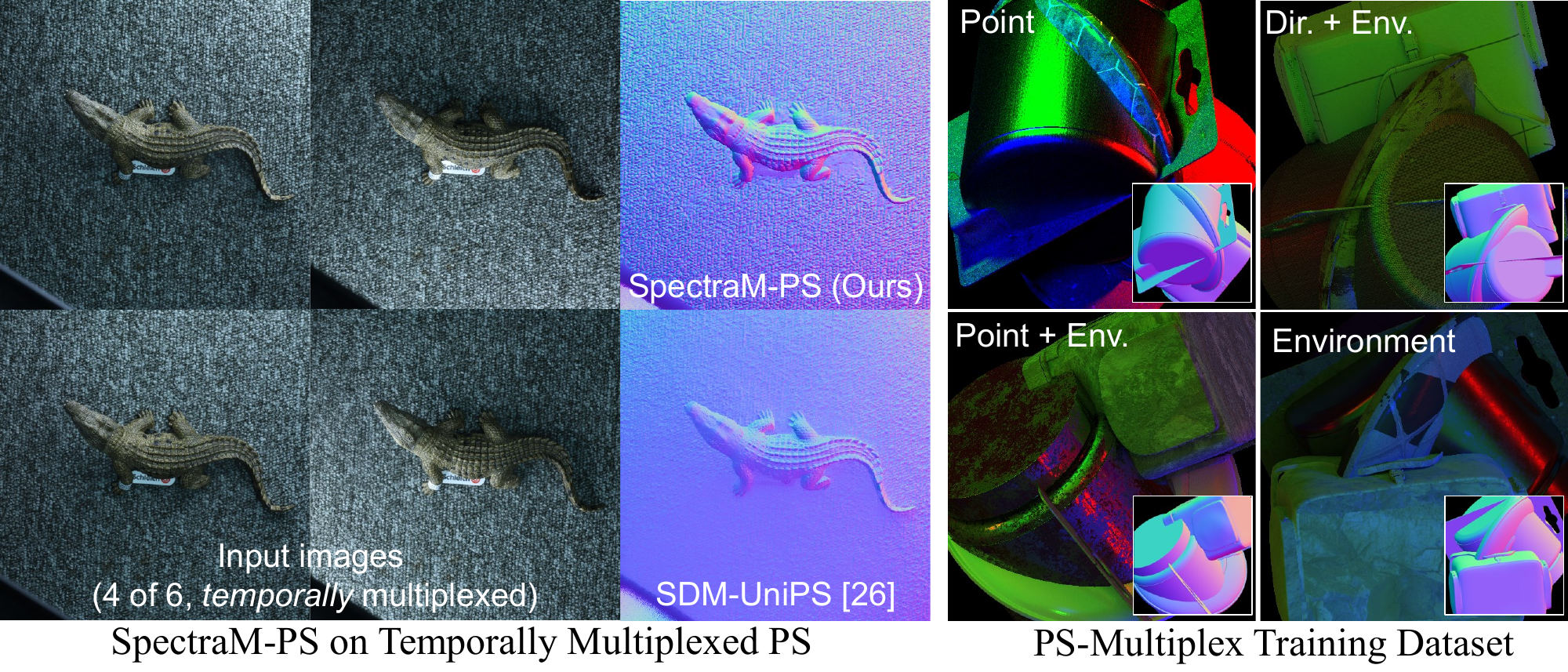}
	\end{center}
	\vspace{-10pt}
	\caption{(Left) Comparison of SpectraM-PS and SDM-UniPS~\cite{Ikehata2023} on six {\it temporally} multiplexed PS images. Due to the patch-wise basis of SpectraM-PS, fine details are better recovered. (Right) Illustration of different lighting conditions in PS-Multiplex.}
	\vspace{-10pt}
	\label{fig:dataset}
\end{figure}
Aligning the training and test data domains in neural networks is essential for optimal model performance~\cite{Ben2010,Tzeng2017}. However, rendering spectrally multiplexed data poses challenges due to the scarcity of multispectral Bidirectional Reflectance Distribution Functions (BRDFs). In reality, ELIE-Net~\cite{Lv2023} was trained using only 51 measured isotropic spectral BRDFs. On the other hand, given the availability of various large isotropic BRDF databases~\cite{Mat3DTextures2023,AdobeStock,MatAmbientCG2023,MatMERLBRDF2014,MatPoliigon2023}, we seek to explore utilizing these datasets for training our model, leveraging the fact that our network does not distinguish images based on their physically-based principles. In this section, we highlight how RGB images serve as a practical approximation, simplifying the complexity inherent in multispectral imaging. 

We begin the discussion by characterizing multispectral imaging. Assuming that the surface doesn't emit light and only reflections on surface are considered, the image formation model is described as follows~\cite{Kajiya1986}:
\begin{equation}
\begin{split}
    I_{(s,p)} &= \int_{\Omega}(\omega_{i}^{\mathsf{T}} n_p)\int_{0}^{\infty}S_s(\lambda)   f_{p}(\omega_{i},\omega_{o}, \lambda) L_{p}(\omega_{i}, \lambda)  \mathrm{d} \lambda \mathrm{d} \omega_{i}.\label{eq:ms_rendering_multi}
\end{split}
\end{equation}

In this equation, \(I_{(s, p)}\) denotes the incoming spectral radiance at the sensor \(s\) (or $s$-th channel) from a surface point \(p\). The term \(f_p\) represents BRDF, \(L_{p}\) the incident light intensity at the surface point, and \(\lambda\) the wavelength of the incident light. The symbols \(\omega_{i}\) and \(\omega_{o}\) denote the directions of incident and reflected light, respectively. \(S_s(\lambda)\) refers to the spectral sensitivity of the sensor \(s\) at wavelength \(\lambda\), \(n_p\) is the surface normal, and \(\Omega\) represents the hemisphere over which incident light directions are possible. The integral sums over all incident directions and wavelengths. It is important to note that the incident light intensity \(L_p\) depends not only on the direct contribution from light sources but also on the visibility of light (e.g., attached and cast shadows) and indirect illuminations.

\cref{eq:ms_rendering_multi} illustrates the concept of spectral ambiguity, showing that an infinite number of combinations of \(S_s(\lambda)\), \(f_{p}(\omega_{i}, \omega_{o}, \lambda)\), and \(L_{p}(\omega_{i}, \lambda)\) can result in the same spectral radiance, including narrowband compositions. In other words, with spectral ambiguity, a single observation \(I_{(s,p)}\) can encompass the observations for all spectral compositions that satisfy the equation (\ie, metamerism~\cite{Nayatani1972,Hill1999}). This perspective justifies the theory of substituting multispectral images, which possess a broad parameter space, with narrowband RGB images. It is worth mentioning that channel crosstalk primarily affects the incident light intensity, consequently distorting the product of \( S_s(\lambda) \cdot f_{p}(\omega_{i}, \omega_{o}, \lambda) \cdot L_{p}(\omega_{i}, \lambda) \) in \cref{eq:ms_rendering_multi}. This implies that observations influenced by spectral crosstalk can still be equivalently represented using a narrowband setup under spectral ambiguity. In the experiments, we demonstrate that our model, trained on three narrowband observations can be applied to multiplexed data with channel crosstalk. To realize this approximation, we rendered a large number of three-channel narrowband images using the path-tracing algorithm in Blender~\cite{Blender}, where up to 10-bounce reflections are permitted, based on common 3D assets~\cite{AdobeStock} for RGB rendering. Following the rendering pipeline described in~\cite{Ikehata2023}, we rendered objects by combining three different lighting models: directional, point, and environmental (five combinatorial settings in total as shown in~\cref{fig:dataset}). To simulate spectrally multiplexed images, we defined R, G, and B light sources and illuminated the surface in a multiplexed manner. It is important to note that the rendered RGB images are decomposed into three grayscale images, each of which was independently fed into the network; therefore, any wavelength-dependent information is masked. For material diversity, we adopted the method from~\cite{Ikehata2023}, categorizing 897 AdobeStock texture maps into three groups: 421 diffuse, 219 specular, and 257 metallic textures. Four objects from a set of 410 3D AdobeStock models were randomly selected and textured with these materials. This structured approach led to the rendering of 106,374 multiplexed images along with their ground truth surface normal maps, forming the `PS-Multiplex' dataset.

\section{SpectraM14 Benchmark Dataset}
\begin{figure}[t!]
	\begin{center}
		\includegraphics[width=120mm]{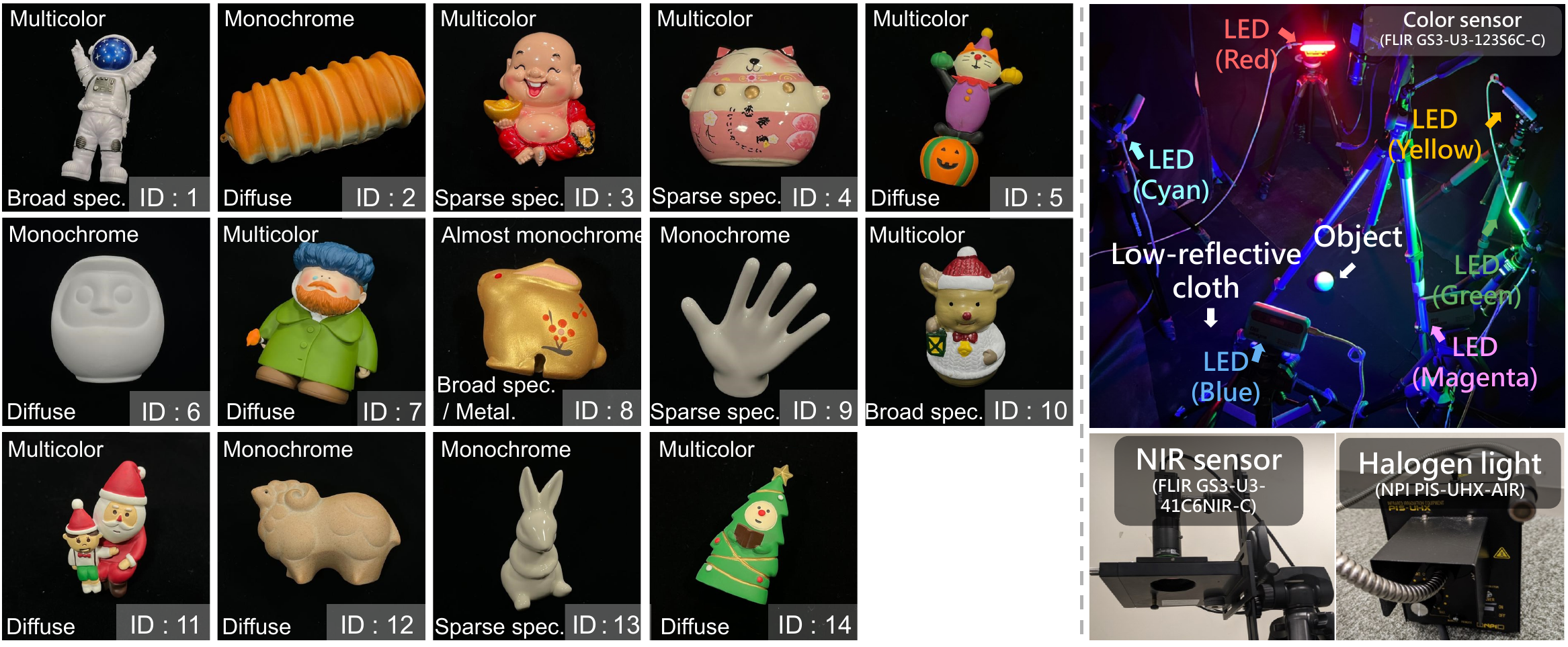}
	\end{center}
	\vspace{-15pt}
	\caption{Objects in SpectraM14.}
	\label{fig:realdata}
	\begin{center}
		\includegraphics[width=120mm]{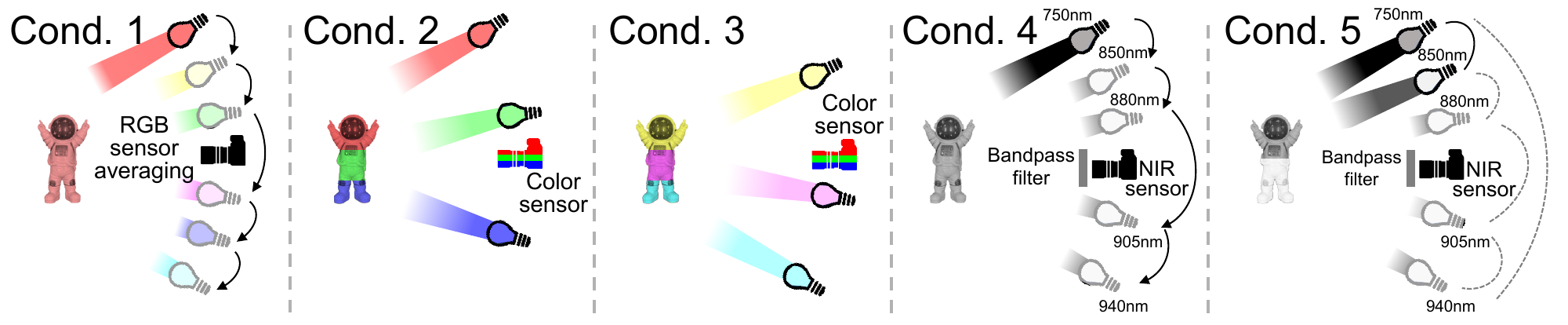}
	\end{center}
	\vspace{-15pt}
	\caption{Illustration of six conditions in SpectraM14.}
	\label{fig:setup}
\end{figure}
Due to the lack of a benchmark for spectrally multiplexed PS, the first comprehensive evaluation dataset, named SpectraM14, is created. This dataset includes 14 objects, each exhibiting a range of optical properties such as monochromatic or multicolored appearances and diffuse or specular reflections, as depicted in~\cref{fig:realdata}. Our benchmark encompasses tasks under five distinct conditions, as described later. 

\vspace{3pt}\textnoindent{Imaging Setup}
To acquire our dataset, we utilized a color camera (FLIR GS3-U3-123S6C-C) and an NIR camera (FLIR GS3-U3-41C6NIR-C), both equipped with a 50mm lens. For the NIR camera, we used narrowband filters with wavelengths of 750nm, 850nm, 880nm, 905nm, and 940nm, and the acquired images were manually merged. Objects were placed 0.8m from the camera to approximate orthographic projection. Following conventional PS benchmarks~\cite{Shi2016,Ren2022,Wang2023}, data capture occurred in a controlled, dark environment with the scene draped in black cloth to mitigate interreflection. The camera's ISO sensitivity was minimized to enhance image quality. The imaging area was further isolated using low-reflectance cloths to suppress inter-reflection. For each illumination condition, we collected six images under varying exposures to produce HDR input images. For the evaluations throughout this paper, the images are cropped using an object mask and resized to 512px × 512px. 

\vspace{3pt}\textnoindent{Lighting}
Six LED and three halogen light sources, positioned roughly 1 meter from the object, provided illumination. We used the ``Weeylite S05 RGB Pocket Lamp'' and the ``NPI PIS-UHX-AIR'' for lighting. This setup enabled the use of red, green, blue, yellow, magenta, cyan, and NIR lighting, with spectra validated using a Hamamatsu Photonics Multichannel Analyzer C10027-01.

\vspace{3pt}\textnoindent{Calibration and Ground Truth Data}
We measured the directions of lights using specular reflections from a mirror sphere. Light intensity was standardized across the visible spectrum by averaging RGB values from reflected light on a white target. The ground truth normals were captured with a SHINING 3D EinScan-SE scanner.

\vspace{3pt}\textnoindent{Evaluation Procedure} 
The design philosophy of this benchmark is to assess the robustness and adaptability of spectrally multiplexed PS methods under realistic lighting conditions, accounting for variations in channel numbers and the presence of spectral crosstalk. For a comprehensive evaluation, we designed tasks under five distinct conditions as shown in~\cref{fig:setup}: \textbf{Condition 1}: Color sensor, no crosstalk condition: Six colors of light (red, green, blue, cyan, yellow, magenta) were each independently illuminated and observed with an RGB sensor. Afterward, the channels of RGB were averaged. \textbf{Condition 2}: Color sensor, weak crosstalk condition: Three colors of light (red, green, blue) were simultaneously illuminated and observed through each channel of the RGB sensor. \textbf{Condition 3}: Color sensor, strong crosstalk condition: Three colors of light (cyan, yellow, magenta) were simultaneously illuminated and observed through each channel of the RGB sensor. \textbf{Condition 4}: NIR sensor, no crosstalk condition: Light at wavelengths of 750 nm, 850 nm, 880 nm, 905 nm, and 940 nm were each independently illuminated and observed with a monochrome sensor corresponding to each wavelength. \textbf{Condition 5}: NIR sensor, spatially-varying lighting condition: New images were created by averaging two images taken under the conditions mentioned above. The combinations were (750 nm, 850 nm), (850 nm, 880 nm), (880 nm, 905 nm), (905 nm, 940 nm), and (940 nm, 750 nm).

\section{Experiment}
In this section, we evaluate our method on our SpectraM14. Our method is compared with one SOTA optimization-based method~\cite{Guo2022} and one SOTA learning-based method~\cite{Lv2023}. The former introduces a closed-form solution for spectrally multiplexed photometric stereo applied to monochromatic surfaces with spatially varying (SV) albedo. The latter presents a Spectral Reflectance Decomposition (SRD) model, which disentangles spectral reflectance into geometric and spectral components for surface normal recovery under non-Lambertian spectral reflectance conditions. Unlike the compared methods, our approach does not assume a specific lighting setup, whereas both methods presume the presence of calibrated single directional light sources.
\label{sec:experiemnt}

\vspace{3pt}\textnoindent{Training details} SpectraM-PS was trained from scratch on the PS-Multiplex dataset until convergence using the AdamW optimizer, with a step decay learning rate schedule that reduced the learning rate by a factor of 0.8 every ten epochs. We applied learning rate warmup during the first epoch and used a batch size of 16, an initial learning rate of 0.0001, and a weight decay of 0.05. Each batch consisted of three input training multiplexed images with three channels each. The training loss was computed using the Mean Squared Error (MSE) loss function to measure \(\ell_2\) errors between the predicted surface normal vectors and the ground truth surface normal vectors. We measured the reconstruction accuracy of our method by computing the mean angular errors (MAE) between the predicted and true surface normal maps, expressed in degrees.

\vspace{3pt}\textnoindent{Computational Cost} The inference time of PS methods varies with the number of pixels and channels in the input image. For Condition 2 and 3 with a \(512 \times 512 \times 3\) image, the mean and standard deviation of inference times (in sec) over 14 objects in SpectraM14 benchmark were: our method (3.42/0.85), Lv~\etal.~\cite{Lv2023} (0.46/0.24) and Guo~\etal.~\cite{Guo2022} (2.38/1.10). Our architecture leads to higher computational costs; however, none of the methods were suitable for real-time processing (\eg, 15 fps requires 0.06 sec/frame).

\vspace{3pt}\textnoindent{Ablation Study}
We firstly validate the individual technical contributions of our training dataset (\ie, PS-Multiplex) and the physics-free architecture (\ie, SpectraM-PS) using a synthetic evaluation dataset. 
Firstly, we validate the efficacy of our training dataset, PS-Multiplex, by adapting an existing universal photometric stereo architecture designed for the conventional task (\ie, SDM-UniPS~\cite{Ikehata2023}) to the spectrally multiplexed photometric stereo task. Since both ours and SDM-UniPS take multiple observations and an object mask as input, this adaptation straightforwardly involves training the model on PS-Multiplex by treating each channel of an image as an individual image. Subsequently, we compare this model against our proposed SpectraM-PS to demonstrate the efficacy of our dual-scale design with local patch embedding.

For evaluating the contribution of our architecture (SpectraM-PS) and training dataset (PS-Multiplex), we additionally rendered three-channel spectrally multiplexed images representing six distinct surface material categories: (a) uniform, Lambertian; (b) piece-wise uniform, Lambertian; (c) non-uniform, Lambertian; (d) uniform, non-Lambertian; (e) piece-wise uniform, non-Lambertian; and (f) non-uniform, non-Lambertian. In uniform materials, every point on the surface within a scene exhibits the same material properties. For piece-wise uniform materials, each object in a scene is composed of the same material, yet different objects possess distinct materials. Non-uniform materials feature unique PBR textures assigned to each object. The rendering process for these images was identical to that used for the PS-Multiplex datasets in each category. We generated 100 scenes for each surface material category, and MAEs (stds) are averaged over them. The results are presented in Tab.\textcolor{red}{1}. In summary, SDM-UniPS~\cite{Ikehata2023} trained on our PS-Multiplex dataset demonstrates proper adaptation to the spectrally multiplexed photometric stereo task. Nonetheless, our SpectraM-PS method significantly enhanced reconstruction accuracy, showcasing an architecture-level improvement over SDM-UniPS for the spectrally multiplexed photometric stereo task, where the number of input channels is typically much fewer than that of input images for conventional PS.
\begin{table}[!t]
	\begin{center}
 \caption{Ablation analysis of the contributions of SpectraM-PS and PS-Multiplex.}
 \vspace{-10pt}	
 \includegraphics[width=120mm]{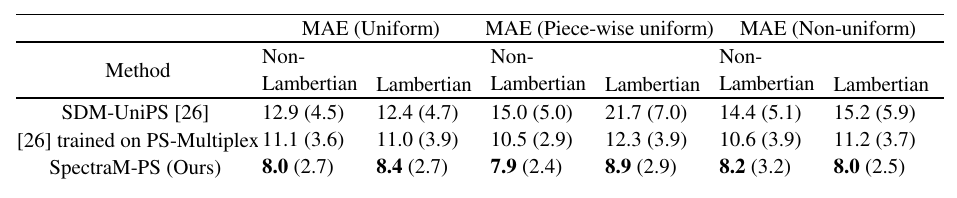}
	\end{center}	
	\label{table:ablation_study}
	\vspace{-30pt}
\end{table}

\begin{figure}[!t]
	\begin{center}
 \includegraphics[width=120mm]{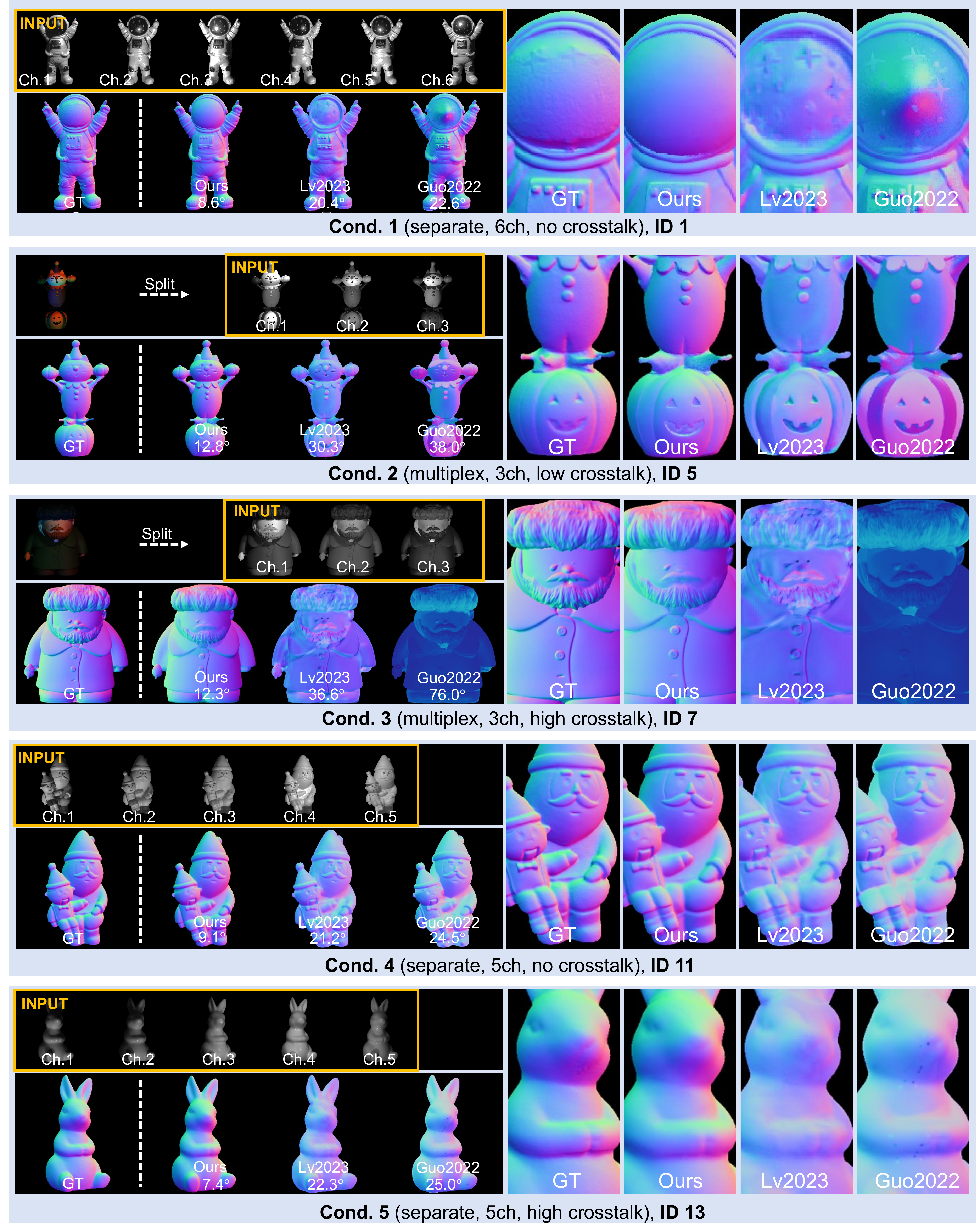}
	\end{center}
	\vspace{-10pt}
	\caption{Evaluation on SpectraM14. Full results are available in the supplementary.}
	\label{fig:spectram14}
	\vspace{-10pt}
\end{figure}

\vspace{3pt}\textnoindent{Comparative Evaluation on SpectraM14}
\begin{table}[!h]
{\scriptsize
\begin{center}
\centering
\caption{Comparison in condition 1. The values are mean angular errors in degrees.}
\vspace{-12pt}
\label{tab:comparison1}
\setlength{\tabcolsep}{2pt} 
\begin{tabular}{c|*{14}{c}|c}
\toprule
  \multirow{2}{*}{Method} & \multicolumn{13}{c}{Object ID}& \\
 & 1 & 2 & 3 & 4 & 5 & 6 & 7 & 8 & 9 & 10 & 11 & 12 & 13 & 14 & Ave.\\
\hline
Ours & \textbf{8.6} & \textbf{11.2} & \textbf{13.0} & \textbf{7.3} & \textbf{12.2} & \textbf{6.1} & \textbf{11.5} & \textbf{5.0} & \textbf{5.7} & \textbf{7.5} & \textbf{10.3} & \textbf{5.1} & \textbf{6.1} & \textbf{12.2} & \textbf{8.9} \\
Lv~\etal~\cite{Lv2023} & 20.4 & 17.0 & 21.1 & 13.9 & 23.1 & 10.7 & 21.2 & 16.6 & 10.9 & 15.9 & 19.0 & 16.4 & 13.2 & 18.6 & 17.1 \\
Guo~\etal~\cite{Guo2022} & 22.6 & 15.2 & 20.7 & 13.4 & 27.1 & 7.2 & 31.3 & 24.8 & 8.0 & 18.2 & 24.5 & 11.4 & 10.2 & 29.3 & 18.9 \\
\bottomrule
\end{tabular}
\end{center}
\caption{Comparison in condition 2. The values are mean angular errors in degrees.}
\vspace{-22pt}
\label{tab:comparison2}
\setlength{\tabcolsep}{2pt} 
\begin{center}
\begin{tabular}{c|*{14}{c}|c}
\toprule
  \multirow{2}{*}{Method} & \multicolumn{13}{c}{Object ID}&\\
 & 1 & 2 & 3 & 4 & 5 & 6 & 7 & 8 & 9 & 10 & 11 & 12 & 13 & 14 & Ave.\\ 
\hline
Ours & \textbf{10.1}&\textbf{11.5}&\textbf{13.5}&\textbf{9.0}&\textbf{12.8}&\textbf{7.0}&\textbf{10.9}&\textbf{5.5}&\textbf{6.3}&\textbf{9.3}&\textbf{12.7}&\textbf{5.3}&\textbf{9.8}&\textbf{14.7}&\textbf{10.0}\\
Lv~\etal~\cite{Lv2023} & 22.8&26.0&27.0&19.4&30.3&19.5&22.1&18.9&14.3&19.8&23.5&20.8&19.0&21.4&21.7\\
Guo~\etal~\cite{Guo2022} & 31.1&27.5&29.2&20.1&38.0&19.0&33.4&23.5&13.6&26.6&32.7&14.7&17.1&39.3&25.7\\
\bottomrule
\end{tabular}
\end{center}
\caption{Comparison in condition 3. The values are mean angular errors in degrees.}
\vspace{-22pt}
\label{tab:comparison3}
\setlength{\tabcolsep}{2pt} 
\begin{center}
\begin{tabular}{c|*{14}{c}|c}
\toprule
  \multirow{2}{*}{Method} & \multicolumn{14}{c}{Object ID}&\\
  & 1 & 2 & 3 & 4 & 5 & 6 & 7 & 8 & 9 & 10 & 11 & 12 & 13 & 14 & Ave.\\ 
\hline
Ours & \textbf{12.0}&\textbf{12.8}&\textbf{15.8}&\textbf{11.0}&\textbf{16.5}&\textbf{5.9}&\textbf{12.3}&\textbf{9.8}&\textbf{6.9}&\textbf{9.8}&\textbf{14.6}&\textbf{6.3}&\textbf{7.5}&\textbf{20.1}&\textbf{11.6}\\
Lv~\etal~\cite{Lv2023} & 38.5&34.1&38.2&32.4&40.1&38.2&36.6&29.6&38.2&35.1&38.2&36.6&38.4&30.9&36.0\\
Guo~\etal~\cite{Guo2022} & 46.0&42.6&56.3&37.6&57.1&45.6&76.0&48.0&29.9&62.1&49.8&50.2&52.8&73.7&51.7\\
\bottomrule
\end{tabular}
\end{center}
}
{\scriptsize
\caption{Comparison in condition 4. The values are mean angular errors in degrees.}
\vspace{-22pt}
\label{tab:comparison4}
\setlength{\tabcolsep}{2pt} 
\begin{center}
\begin{tabular}{c|*{14}{c}|c}
\toprule
  \multirow{2}{*}{Method} & \multicolumn{13}{c}{Object ID}&\\
 & 1 & 2 & 3 & 4 & 5 & 6 & 7 & 8 & 9 & 10 & 11 & 12 & 13 & 14 & Ave.\\ 
\hline
Ours & \textbf{10.9}&\textbf{11.3}&\textbf{9.7}&\textbf{6.3}&\textbf{10.8}&\textbf{5.3}&\textbf{12.4}&\textbf{4.3}&\textbf{7.8}&\textbf{6.6}&\textbf{9.1}&\textbf{4.0}&\textbf{7.4}&\textbf{10.6}&\textbf{8.4}\\
Lv~\etal~\cite{Lv2023} & 24.1&19.8&20.7&12.0&19.8&12.4&15.1&18.7&11.4&17.1&21.2&17.5&19.1&16.7&17.5\\
Guo~\etal~\cite{Guo2022} & 30.7&16.0&18.3&9.9&29.6&8.4&23.1&25.7&10.2&14.6&24.5&13.9&29.6&13.8&19.1\\
\bottomrule
\end{tabular}
\end{center}
\caption{Comparison in condition 5. The values are mean angular errors in degrees.}
\vspace{-22pt}
\label{tab:comparison5}
\setlength{\tabcolsep}{2pt} 
\begin{center}
\begin{tabular}{c|*{14}{c}|c}
\toprule
  \multirow{2}{*}{Method} & \multicolumn{13}{c}{Object ID}&\\
 & 1 & 2 & 3 & 4 & 5 & 6 & 7 & 8 & 9 & 10 & 11 & 12 & 13 & 14 & Ave.\\ 
\hline
Ours & \textbf{10.9}&\textbf{11.0}&\textbf{9.9}&\textbf{7.3}&\textbf{11.0}&\textbf{4.7}&\textbf{13.3}&\textbf{4.8}&\textbf{7.9}&\textbf{6.5}&\textbf{9.8}&\textbf{4.2}&\textbf{7.4}&\textbf{10.4}&\textbf{8.6}\\
Lv~\etal~\cite{Lv2023} & 29.8&26.2&28.9&22.8&28.9&25.8&21.9&24.6&18.7&24.7&27.6&24.8&22.3&27.6&25.0\\
Guo~\etal~\cite{Guo2022} & 40.0&27.3&29.2&23.2&33.3&25.0&29.3&30.1&20.0&26.7&31.6&26.3&25.0&27.2&27.8\\
\bottomrule
\end{tabular}
\end{center}
}
\vspace{-20pt}
\end{table}
The results are illustrated in~\cref{tab:comparison1,tab:comparison2,tab:comparison3,tab:comparison4,tab:comparison5} and~\cref{fig:spectram14}. Despite the fact that all existing spectrally multiplexed photometric stereo methods assume calibrated light sources and known directional light source conditions, our proposed method significantly outperformed them. This is because most of the real objects used in our experiment are neither Lambertian nor convex, and do not conform to their assumptions. However, our non-physical-based method successfully restored the normals very stably for these objects. Furthermore, our proposed method enabled robust reconstruction for all objects, despite having been trained only with RGB color images. This result supports the efficacy of our approximation. Furthermore, unlike existing methods that suffer from reduced estimation accuracy with increasing spectral crosstalk, our approach demonstrates only minimal performance degradation. Remarkably, our method excels in recovering a more realistic structure with spatially-varying surface materials. This breakthrough implies that our network can effectively achieve dynamic surface reconstruction across video frames in a universal setting. Due to space constraints, not all results can be included here. However, all results are comprehensively presented in the supplementary materials. Additionally, the supplementary materials evaluate the impact of the spatial distribution of light sources on the performance of the proposed method. We also offer an in-depth discussion of each experimental condition therein.

\section{Conclusion}
\label{sec:conclusion}
In this work, we introduce an innovative approach to spectrally multiplexed photometric stereo under unknown spatial/spectral composition. Turning spectral ambiguity into a benefit, our method allows for the creation of training data without the need for complex multispectral rendering. Our work significantly broadens the scope for dynamic surface analysis, establishing a critical advancement in the utilization of photometric stereo across multiple sectors. Our proposed method exhibits several limitations. Firstly, there is unstable temporal variation in the normal maps reconstructed by our method for dynamic surface reconstruction. This instability arises from factors such as motion blur in certain frames, image noise, or the influence of cast/attached shadows, which become more pronounced compared to conventional photometric stereo methods that utilize numerous images. To recover clean and temporally stable normal maps, we may need to consider temporal consistency and more actively utilize monocular cues. Additionally, while our method targets dynamic surfaces, it currently requires several seconds to up to ten seconds per RGB image, which is far from real-time processing. Considering industrial applications in the future, accelerating the processing speed is a crucial challenge.


%
%
\bibliographystyle{splncs04}
\bibliography{main}

\appendix
\section{Details of SpectraM14 Benchmark} \label{sec:supp_SpectraM14}
In this section, we detail SpectraM14, the first benchmark dataset for spectrally multiplexed photometric stereo. Firstly, we provide a detailed explanation of the five task conditions included in the benchmark. Then, we offer comprehensive information about data acquisition. Finally, we discuss the spectral characteristics of the sensors and light sources.
\subsection{Details of Benchmark Tasks}\label{subsec:tasks}
\begin{figure}[!t]
	\begin{center}
		\includegraphics[width=120mm]{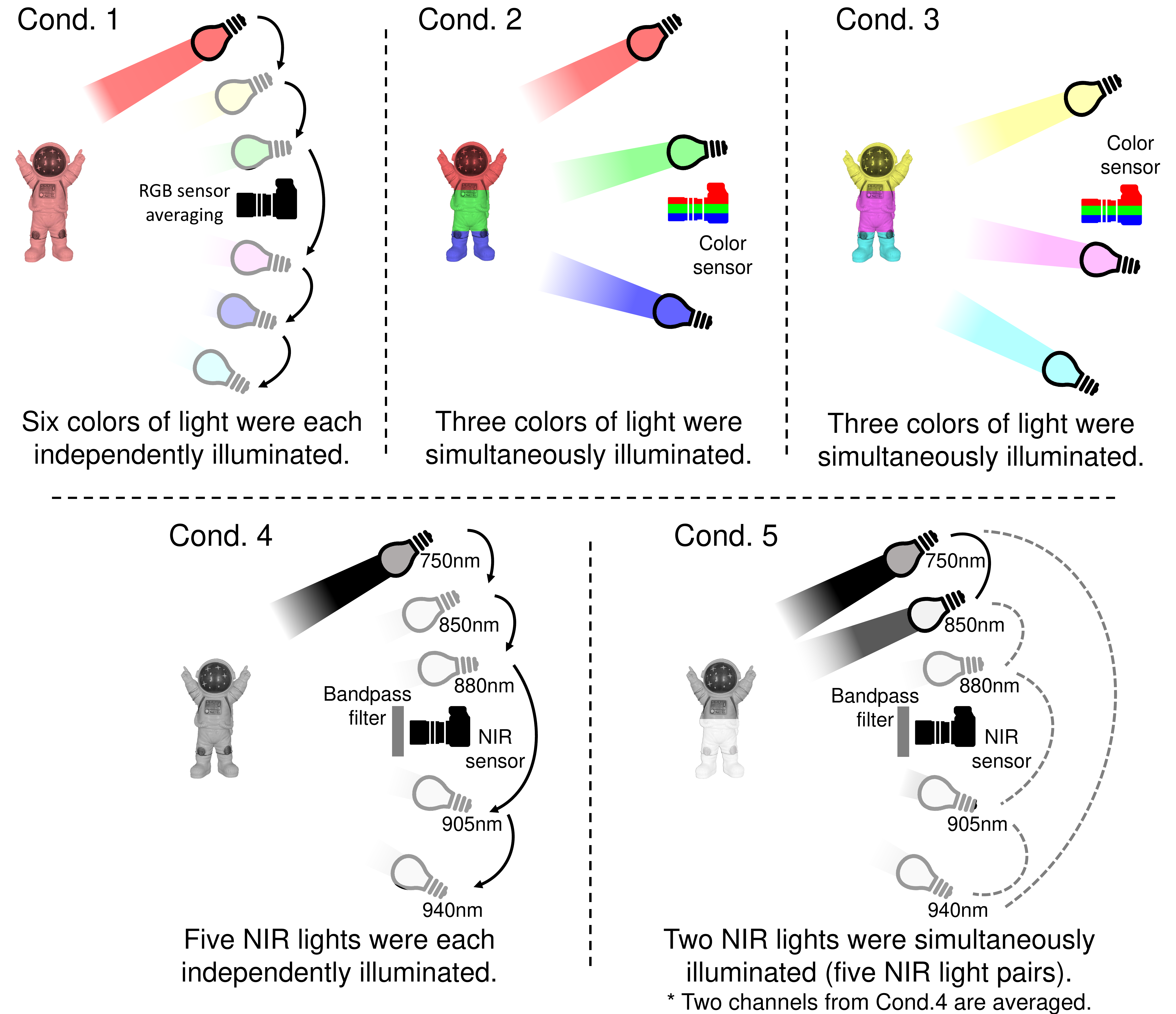}
	\end{center}
	\vspace{-10pt}
	\caption{Illustration of task conditions.}
	\label{fig:supp_conditions}
	\vspace{-5pt}
\end{figure}

Our SpectraM14 benchmark encompasses tasks under five distinct conditions, as illustrated in~\cref{fig:supp_conditions}. The details of these conditions are as follows.
\newpage
\begin{mdframed}[backgroundcolor=gray!20] 
\noindent {\large{\textbf{Condition 1} (six channels)}} 
\begin{itemize}[leftmargin=*,label={}]
    \item \textit{Setup.} 
    \begin{itemize}[leftmargin=*]
        \item Color sensor, no crosstalk condition: Six colors of light (red, green, blue, cyan, yellow, magenta) were each independently illuminated and observed with an RGB sensor. Afterward, the channels of RGB were averaged. 
    \end{itemize}
    \item \textit{Motivation.}
    \begin{itemize}[leftmargin=*]
        \item This condition evaluates the robustness to differences in channel numbers during training (\ie, three) and testing (\ie, six). It also simulates an idealized multiplexing scenario without any channel crosstalk.
    \end{itemize}
\end{itemize}
\end{mdframed}

\begin{mdframed}[backgroundcolor=gray!20] 
\noindent {\large{\textbf{Condition 2} (three channels)}} 
\begin{itemize}[leftmargin=*,label={}]
    \item \textit{Setup.} 
    \begin{itemize}[leftmargin=*]
        \item Color sensor, weak crosstalk condition: Three colors of light (red, green, blue) were simultaneously illuminated and observed through each channel of the RGB sensor. 
    \end{itemize}
    \item \textit{Motivation.}
    \begin{itemize}[leftmargin=*]
        \item Actual multiplexing is employed using RGB LEDs and a color sensor, with LEDs' spectral peaks generally aligning with sensor responses, albeit not narrowband, leading to weak channel crosstalk. This setup tests the method's ability to handle real-world multiplexing scenarios within a typical RGB setup.
    \end{itemize}
\end{itemize}
\end{mdframed}

\begin{mdframed}[backgroundcolor=gray!20] 
\noindent {\large{\textbf{Condition 3} (three channels)}} 
\begin{itemize}[leftmargin=*,label={}]
    \item \textit{Setup.} 
    \begin{itemize}[leftmargin=*]
        \item Color sensor, strong crosstalk condition: Three colors of light (cyan, yellow, magenta) were simultaneously illuminated and observed through each channel of the RGB sensor.
    \end{itemize}
    \item \textit{Motivation.}
    \begin{itemize}[leftmargin=*]
        \item The light source's spectral distribution no longer uniquely matches the RGB channels' sensitivity, leading to strong channel crosstalk and invalidating the assumption of a single directional light source. This setup tests the method under more complex lighting scenarios than those assumed by most existing spectrally multiplexed photometric stereo methods~\cite{Guo2022, Lv2023}.
    \end{itemize}
\end{itemize}
\end{mdframed}
\newpage
\begin{mdframed}[backgroundcolor=gray!20] 
\noindent {\large{\textbf{Condition 4} (five channels)}} 
\begin{itemize}[leftmargin=*,label={}]
    \item \textit{Setup.} 
    \begin{itemize}[leftmargin=*]
        \item NIR sensor, no crosstalk condition: Light at wavelengths of 750 nm, 850 nm, 880 nm, 905 nm, and 940 nm were each independently illuminated and observed with a monochrome sensor corresponding to each wavelength.
    \end{itemize}
    \item \textit{Motivation.}
    \begin{itemize}[leftmargin=*]
        \item Evaluating spectral characteristics beyond visible light can address the concern that learning-based methods trained on specific narrowband wavelengths (\eg, RGB images) may struggle to effectively handle characteristics of unknown wavelengths.
    \end{itemize}
\end{itemize}
\end{mdframed}

\begin{mdframed}[backgroundcolor=gray!20] 
\noindent {\large{\textbf{Condition 5} (five channels)}} 
\begin{itemize}[leftmargin=*,label={}]
    \item \textit{Setup.} 
    \begin{itemize}[leftmargin=*]
        \item NIR sensor, spatially-varying lighting condition: New images were created by averaging two images taken under the conditions mentioned above. The combinations were (750 nm, 850 nm), (850 nm, 880 nm), (880 nm, 905 nm), (905 nm, 940 nm), and (940 nm, 750 nm).
    \end{itemize}
    \item \textit{Motivation.}
    \begin{itemize}[leftmargin=*]
        \item This setup evaluates methods on NIR images under multiple light sources, causing spatially-varying illumination where the assumption of a single directional light source is no longer valid. From the other perspective, this simulates strong spectral multiplexing under NIR lighting and sensors. Due to the practical difficulties of real multiplexing under NIR light, a pseudo-environment is created by averaging multiple NIR images.
    \end{itemize}
\end{itemize}
\end{mdframed}
\subsection{Details of Data Acquisition}
\begin{figure}[!t]
	\begin{center}
		\includegraphics[width=120mm]{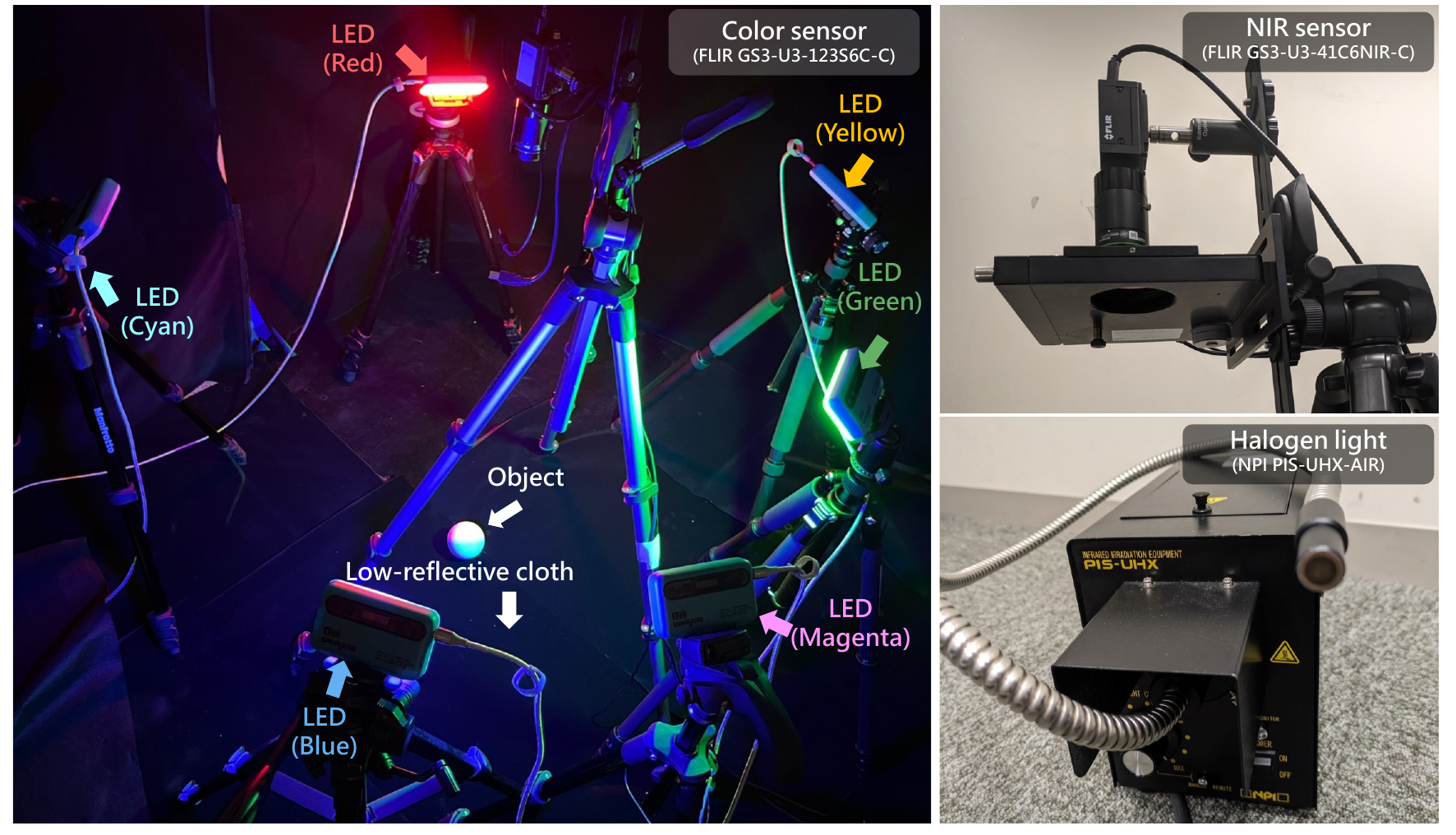}
	\end{center}
	\vspace{-10pt}
	\caption{Imaging setup for SpectraM14 dataset for condition 1--3. In condition 4--5, replace the LED light source with a halogen light and install a bandpass filter in front of the camera lens.}
	\label{fig:setup_asano}
	\vspace{-5pt}
\end{figure}

In our imaging setup, six LED color light sources (Weeylite S05 RGB Pocket Lamp) were positioned around the camera for conditions 1 to 3, and three halogen light sources (NPI PIS-UHX-AIR), whose wavelengths range from visible to NIR, were used for conditions 4 to 5, as shown in~\cref{fig:setup_asano}. The walls and floor were covered with cloth made from low-reflectance material to minimize the effects of inter-reflection. For conditions 4 and 5, an NIR image at each specific wavelength (\ie, 750 nm, 850 nm, 880 nm, 905 nm, and 940 nm) was captured using halogen light as the light source and placing a bandpass filter (Edmund Optics~\cite{EdmundOptics}) in front of the camera lens. In condition 5, synthetic data was generated by averaging two images from those captured in condition 4. Note that we utilized three NIR lights for our convenience to smoothly change lighting conditions. In reality, only a single NIR light was turned on for each capture.

\begin{figure}[h]
	\begin{center}
		\includegraphics[width=120mm]{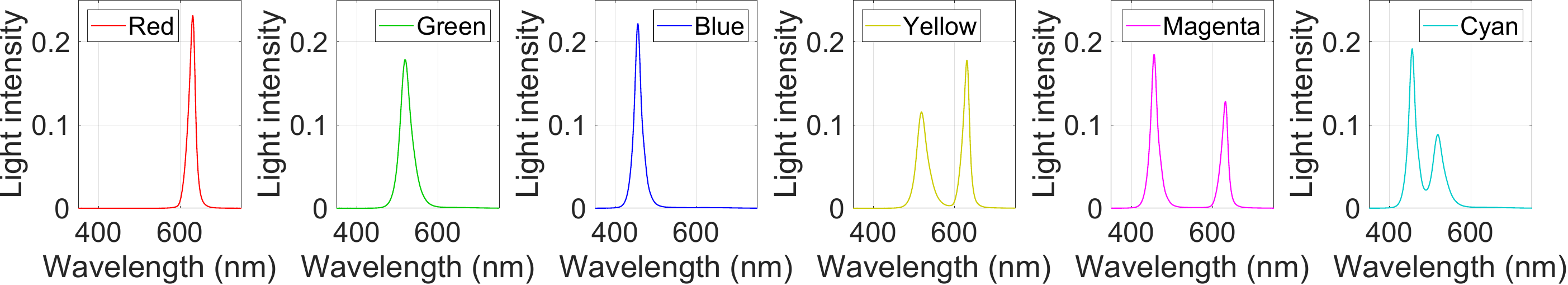}
	\end{center}
	\vspace{-10pt}
	\caption{Spectra of six LED light sources.}
	\label{fig:light_spe_VIS}
	\vspace{-5pt}
\end{figure}

\begin{figure}[h]
	\begin{center}
		\includegraphics[width=120mm]{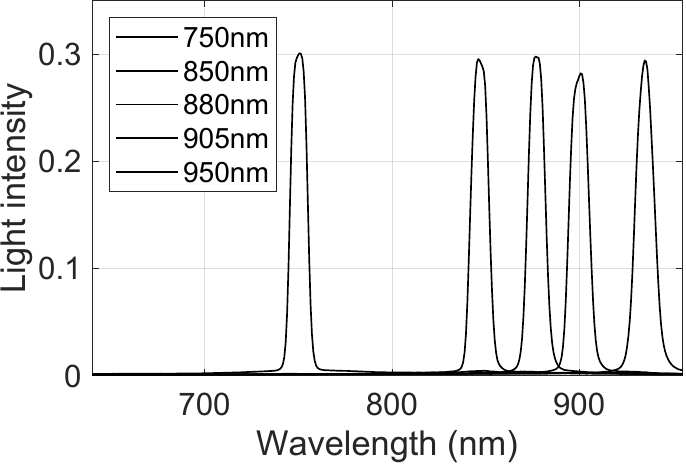}
	\end{center}
	\vspace{-10pt}
	\caption{Spectra of five NIR light sources with bandpass filters.}
	\label{fig:light_spe_NIR}
	\vspace{-5pt}
\end{figure}

The colors of the six LEDs used in our dataset—red, green, blue, yellow, magenta, and cyan—were remotely controlled by the vendor's software~\cite{WeeyliteApp2023}, and their respective spectra, measured by the Hamamatsu Photonics Multichannel Analyzer C10027-01, are shown in~\cref{fig:light_spe_VIS}.
The spectrum of a NIR light source is also shown in~\cref{fig:light_spe_NIR}. Halogen lights emit NIR light strongly, in addition to visible light. By placing a rotating filter holder with multiple bandpass filters in front of the camera lens, multiple near-infrared spectral images can be efficiently acquired without the need for a NIR light source with a specific wavelength. The bandpass filters have a full width-half maximum of 10 nm, and there is minimal spectral crosstalk in the NIR image.

We utilized a color sensor (FLIR GS3-U3-123S6C-C) and an NIR sensor (FLIR GS3-U3-41C6NIR-C), both equipped with a 50 mm lens and having a linear radiometric response function. For each lighting condition, we captured five images with different exposure times (50ms, 100ms, 150ms, 200ms, 300ms, and 500ms), which we then combined to produce a single HDR image. This process enables precise observation of strong specular reflections and shadowed areas, and improves the intensity resolution of the image.

\begin{figure}[h]
  \begin{minipage}[b]{0.33\linewidth}
    \centering
\includegraphics[keepaspectratio, scale=0.11]{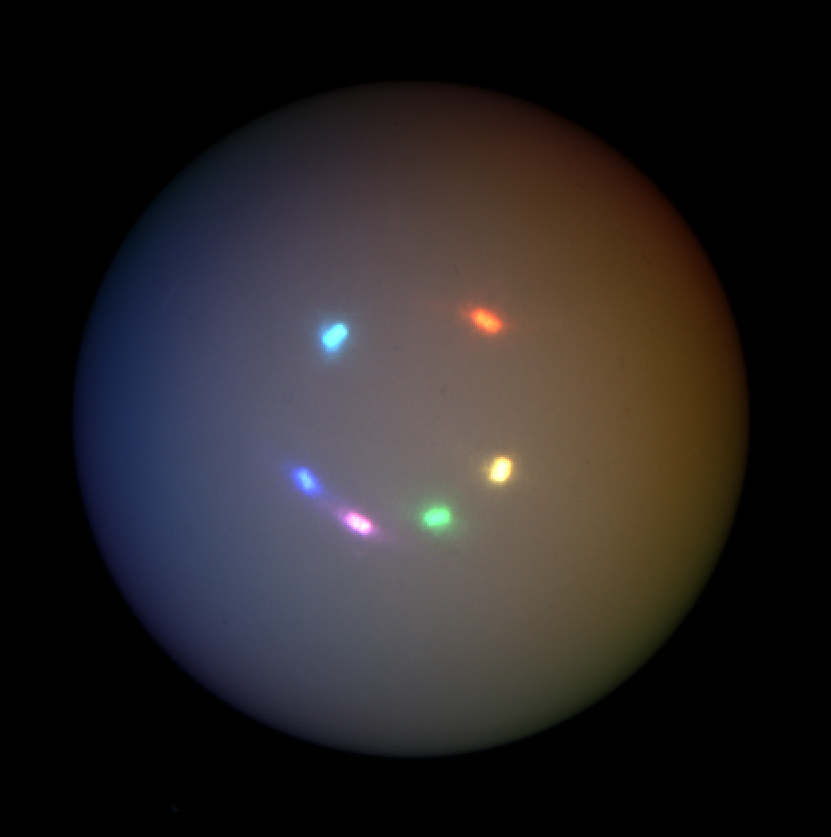}
    \subcaption{A sphere with strong specular characteristics.}
  \end{minipage}
  \begin{minipage}[b]{0.32\linewidth}
    \centering
\includegraphics[keepaspectratio, scale=0.10]{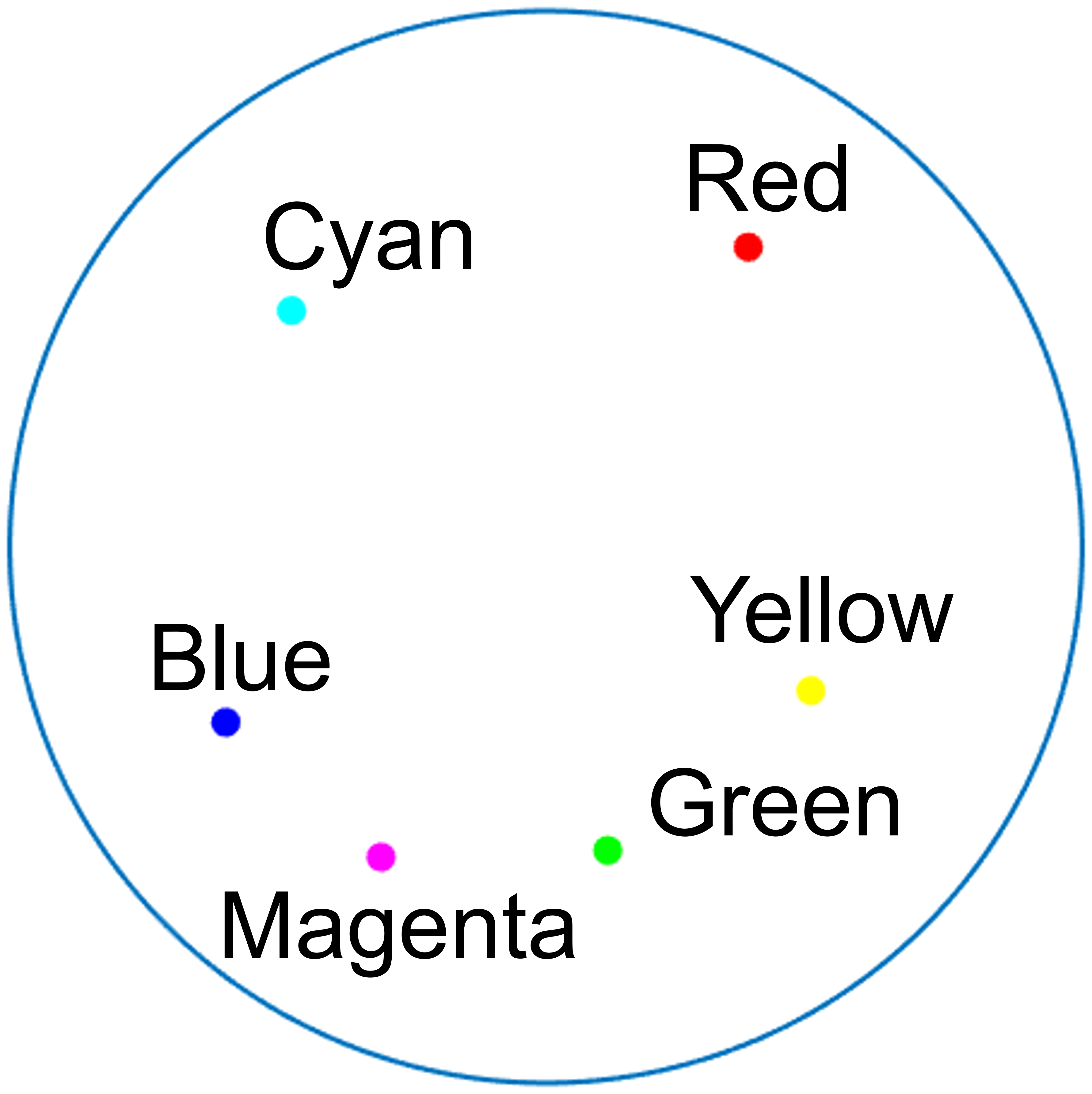}
    \subcaption{Light source directions in condition 1--3. }
  \end{minipage}
  \begin{minipage}[b]{0.32\linewidth}
    \centering
\includegraphics[keepaspectratio, scale=0.10]{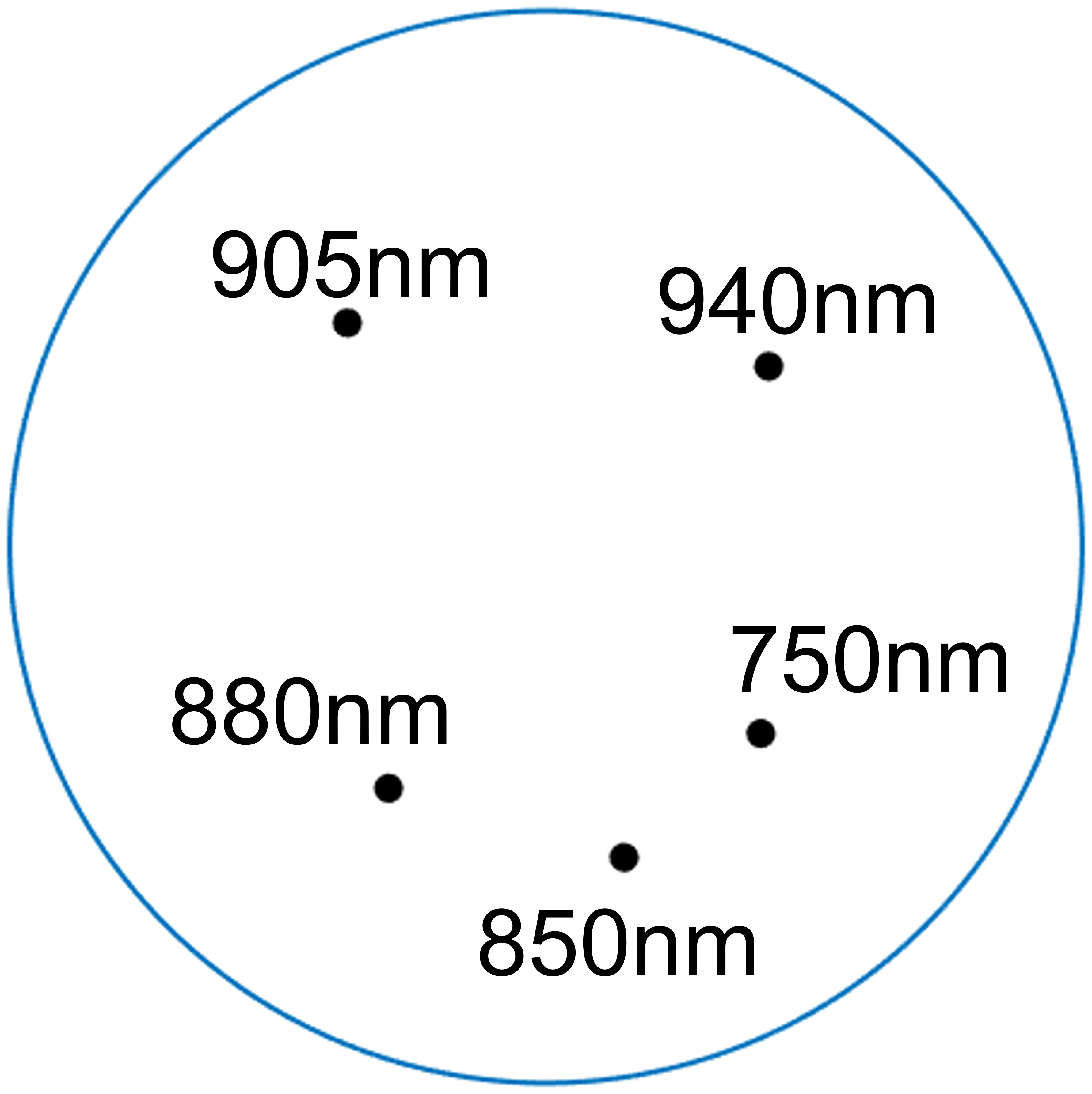}
    \subcaption{Light source directions in conditions 4–5 (with averaging).}
  \end{minipage}
  \caption{Light source distribution of LED  and halogen light sources.}
  \label{fig:light_dir}
\end{figure}

To apply the calibrated methods (\ie, including all prior methods \textit{except for ours}), the direction of each light source is measured based on the specular highlight on a plastic sphere (See~\cref{fig:light_dir}). Note that LED/halogen lights are placed approximately 1m from the object center (the object size is less than 10cm) in roughly uniform directions, maintaining a light-object distance that is about 10 times larger than the object size to practically approximate the directional lighting setup.

\subsection{Spectral Analysis}
\begin{figure}[h]
  \begin{minipage}[b]{0.49\linewidth}
    \centering
\includegraphics[keepaspectratio, scale=0.30]{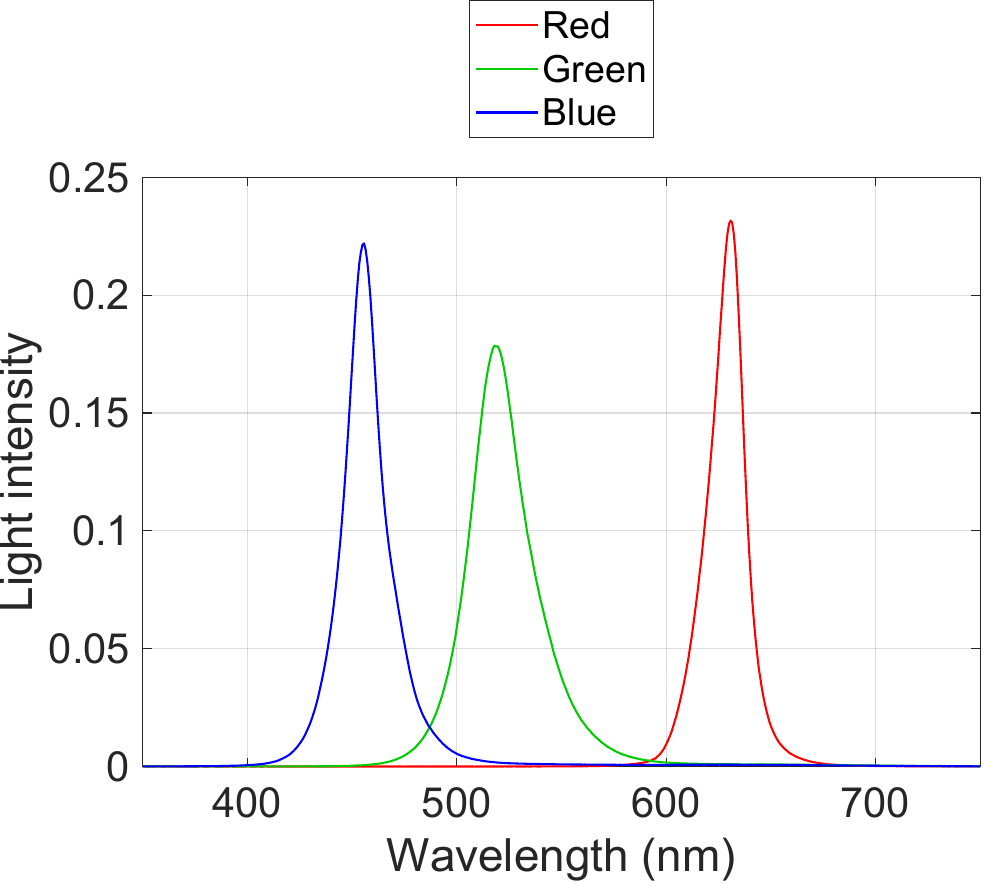}
    \subcaption{Condition 2.}
  \end{minipage}
  \begin{minipage}[b]{0.49\linewidth}
    \centering
\includegraphics[keepaspectratio, scale=0.30]{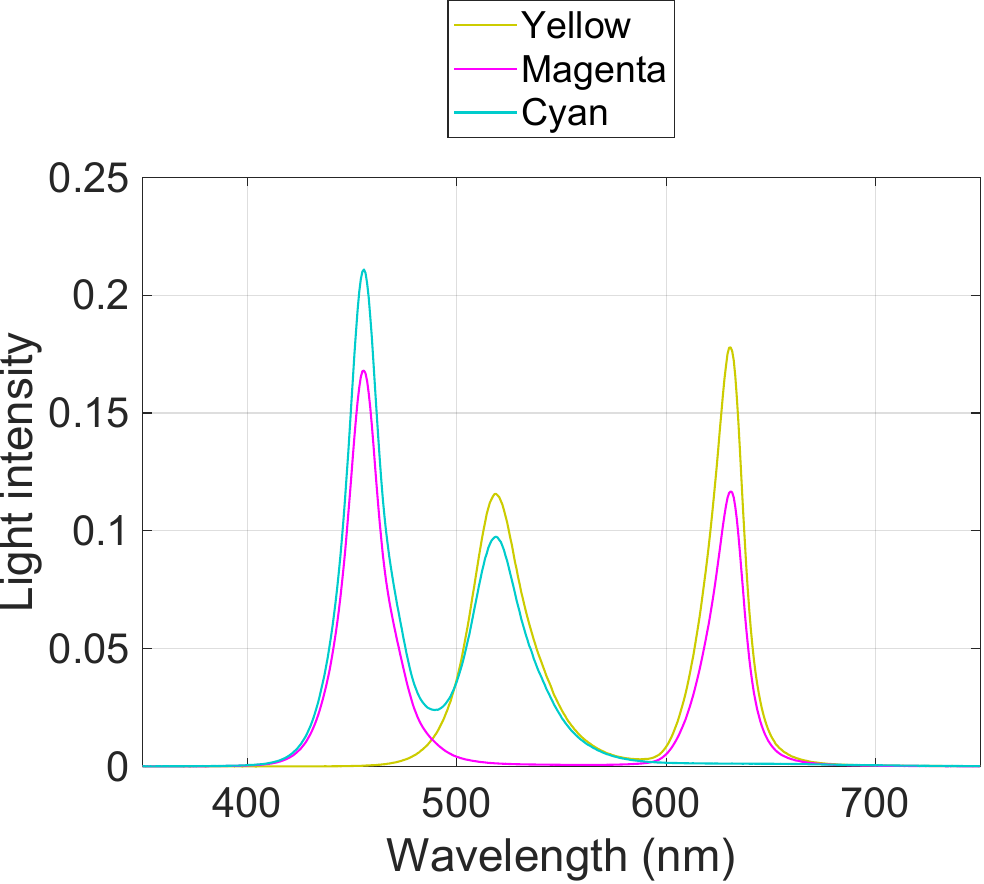}
    \subcaption{Condition 3.}
  \end{minipage}
  \caption{Spectrum of LEDs used in condition 2--3.}
  \label{fig:spe_multi}
\end{figure}

\begin{figure}[h]
  \begin{minipage}[b]{0.49\linewidth}
    \centering
\includegraphics[keepaspectratio, scale=0.30]{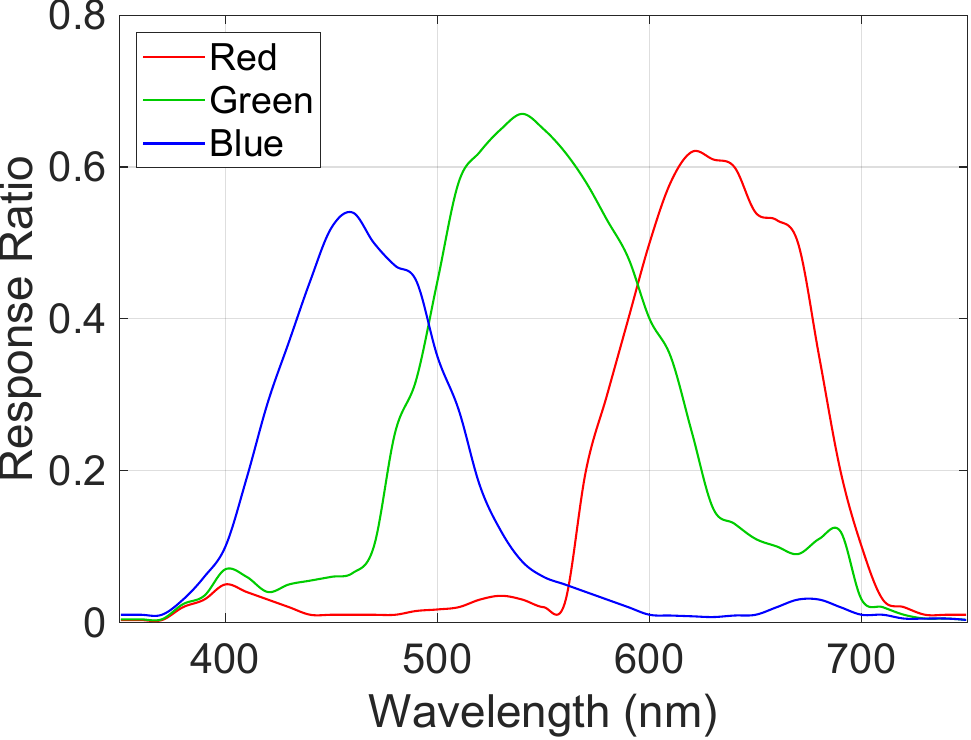}
    \subcaption{Color sensor.}
  \end{minipage}
  \begin{minipage}[b]{0.49\linewidth}
    \centering
\includegraphics[keepaspectratio, scale=0.30]{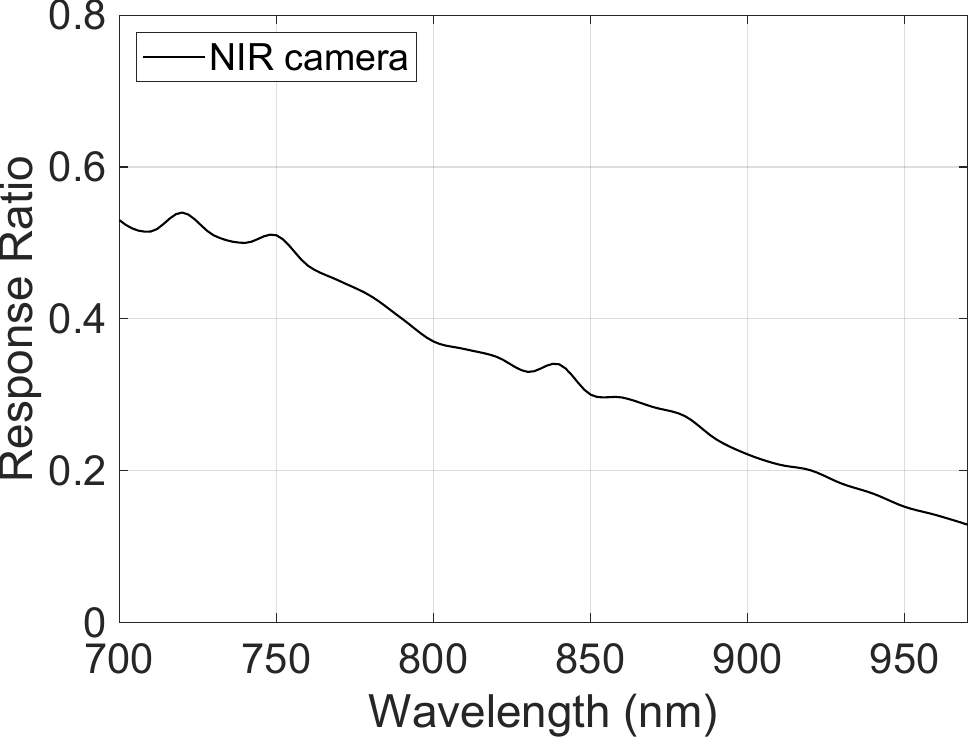}
    \subcaption{NIR sensor.}
  \end{minipage}
  \caption{Sensor spectral response function.}
  \label{fig:camera_spe}
\end{figure}

In spectral multiplexing, channel crosstalk refers to the phenomenon where light from sources of different wavelengths is observed in the same channel of a sensor. This occurs when the sensor's wavelength response characteristics are not sufficiently discriminatory across channels, or when the spectra of the light sources overlap. If channel crosstalk occurs, each channel will observe light from multiple sources, thus disrupting the single lighting assumption. Conditions 2 and 3 are settings designed to compare performance based on the degree of channel crosstalk in actual spectral multiplexing scenarios. Indeed,~\cref{fig:spe_multi} demonstrates that while the spectra of the red, green, and blue LEDs hardly overlap, the spectra of the cyan, yellow, and magenta LEDs significantly overlap, causing substantial channel crosstalk. It is noteworthy that the occurrence of channel crosstalk depends on both the spectrum of the light source and the spectral sensitivity of the sensor. As shown in\cref{fig:camera_spe}, there is also an overlap in sensor spectral response, which makes the influence of crosstalk between images more pronounced. All the spectral properties of lights and sensors were measured using the Hamamatsu Photonics Multichannel Analyzer C10027-01.

\section{Details of Main Results}\label{sec:supp_results}
\begin{figure}[t]
    \begin{center}
        \includegraphics[width=120mm]{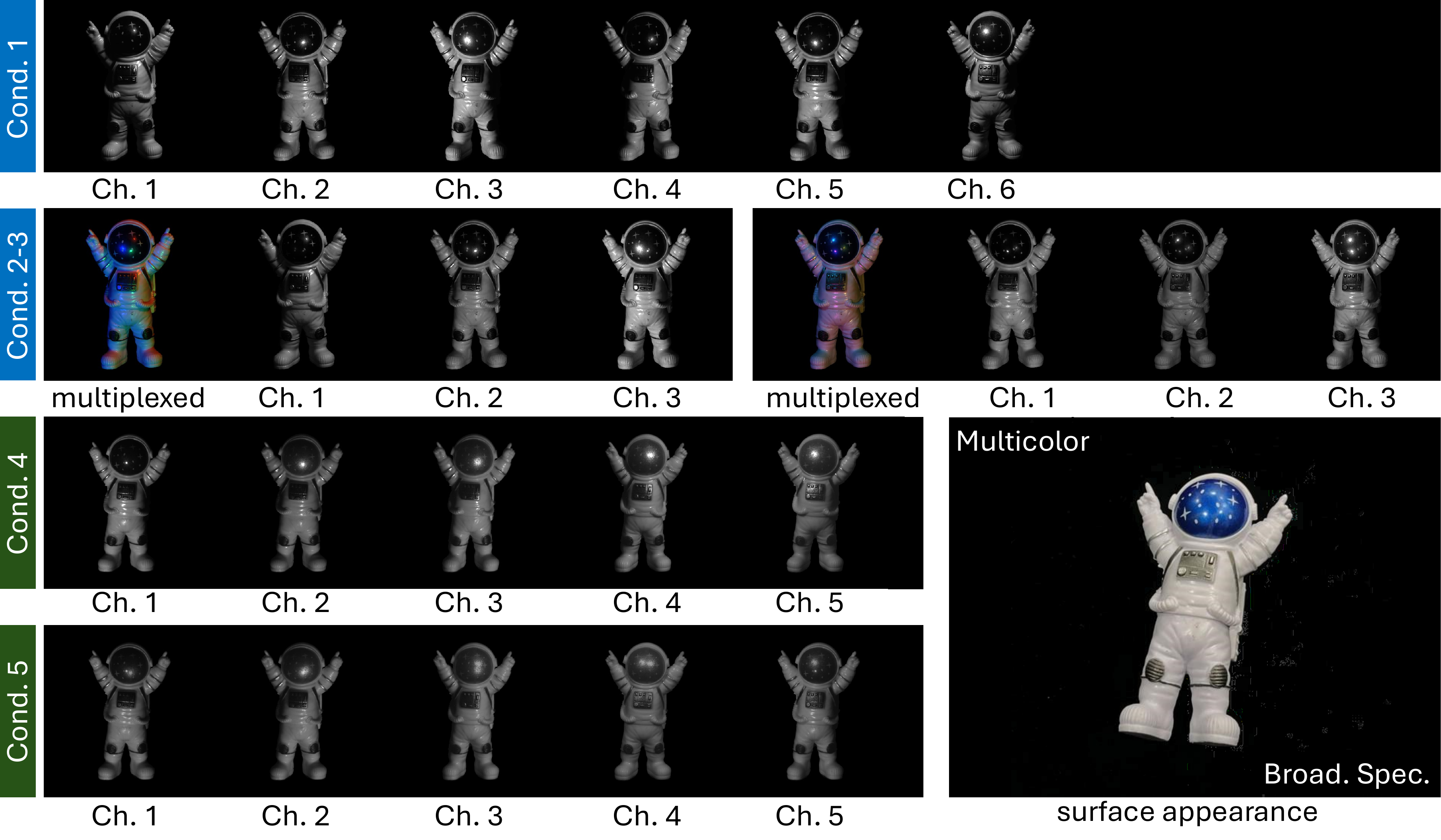}
    \end{center}
    \vspace{-15pt}
        \caption{Input of Object ID 1. A toy astronaut with a simplistic yet recognizable space suit design made from a glossy plastic material. The dark blue helmet has a starry design. The object overall lacks intricate details.}
    \label{fig:input1}
    \vspace{-5pt}
    \begin{center}
        \includegraphics[width=120mm]{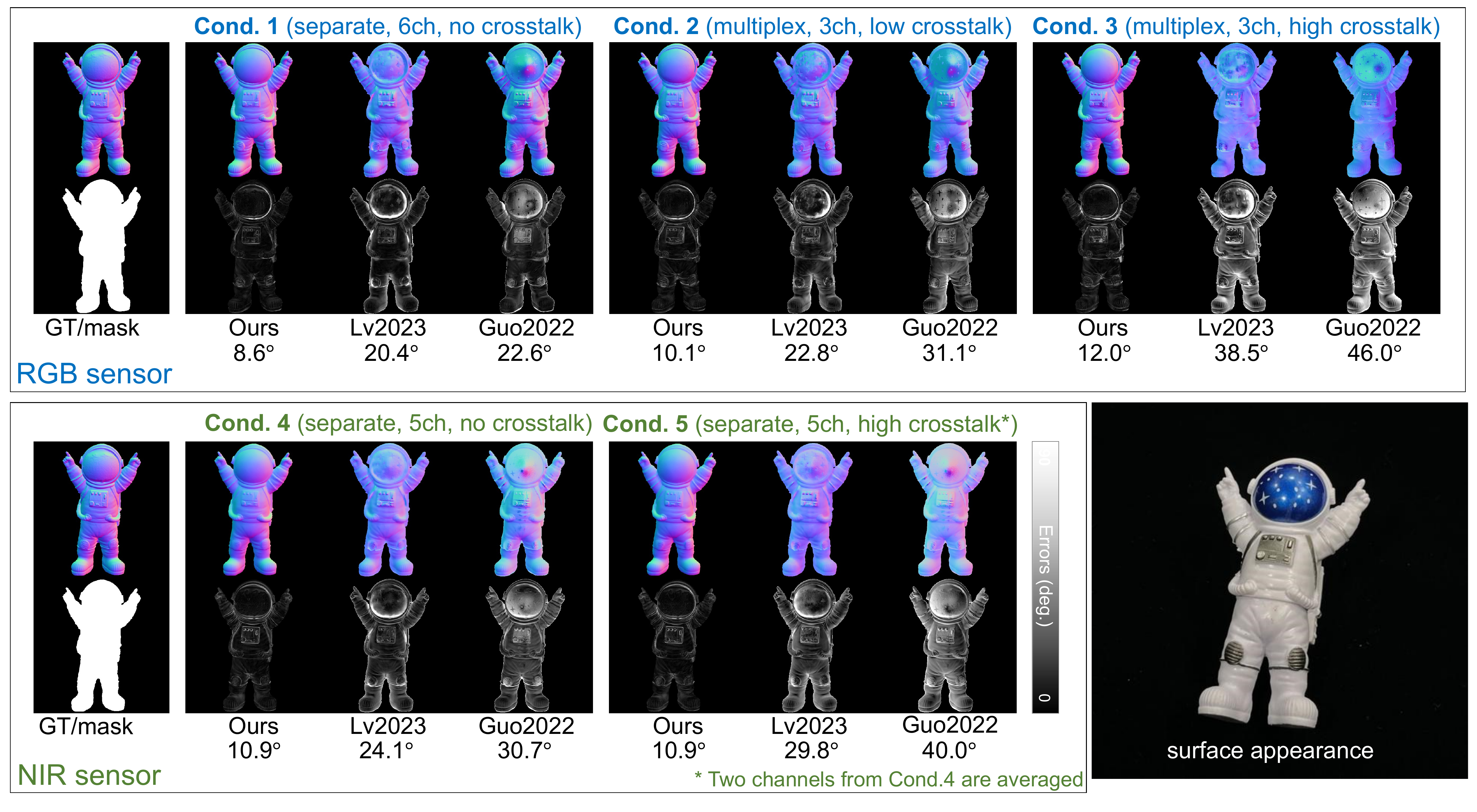}
    \end{center}
    \vspace{-15pt}
	\caption{Output of Object ID 1. The dark blue helmet exhibits very low reflectivity, showing low brightness across all wavelength ranges. The brightness values decrease further as a result of multiplexing in Cond. 2, Cond. 3 and Cond. 5, making the problem more challenging due to the lack of uniformity in quality. The proposed method has successfully recovered the difficult helmet section under any condition, in contrast to prior methods~\cite{Lv2023,Guo2022} which have shown significant errors, especially in this helmet area.}
	\label{fig:output1}
	\vspace{-5pt}
\end{figure}

\begin{figure}[t]
    \begin{center}
    \includegraphics[width=120mm]{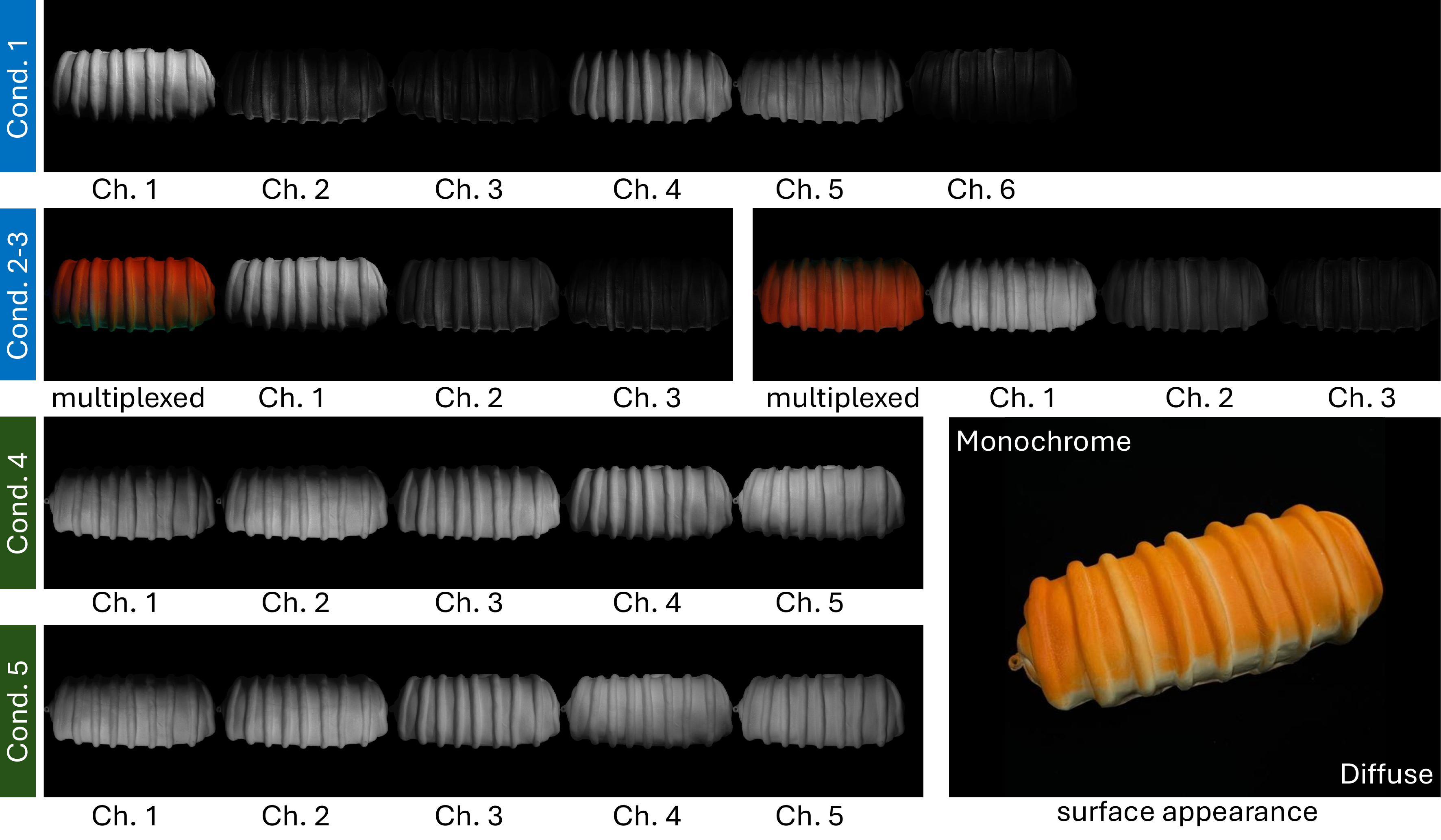}
    \end{center}
    \vspace{-15pt}
    \caption{Input of Object ID 2. A keychain resembling a piece of bread. It is made from a soft, foam-like material, giving it a spongy texture, and has a matte finish. The shape is cylindrical with a series of horizontal indentations that mimic the appearance of sliced bread. It has a mostly uniform light orange color.}
    \label{fig:input2}
    \vspace{-5pt}
    \begin{center}
    \includegraphics[width=120mm]{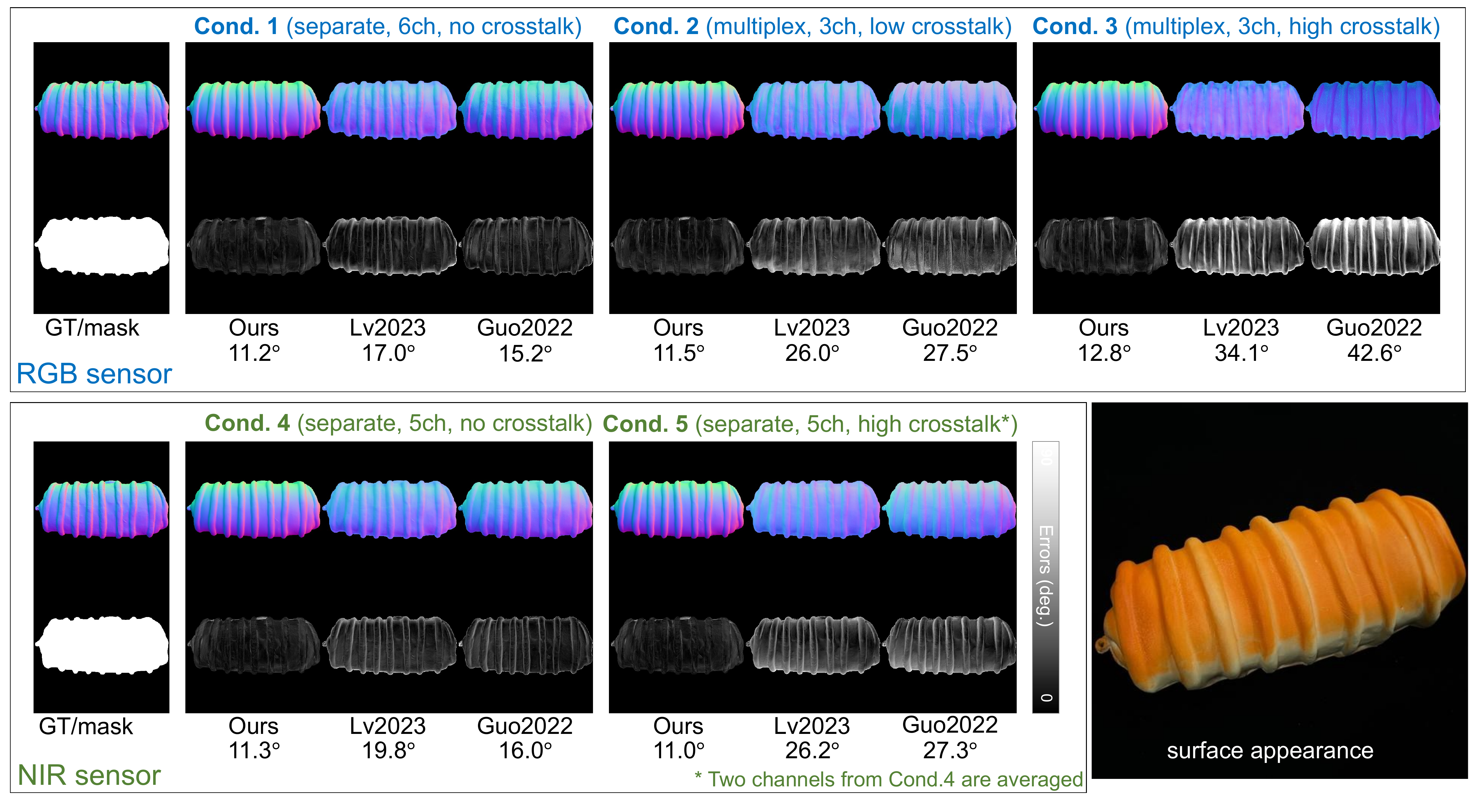}
    \end{center}
    \vspace{-15pt}
    \caption{Output of Object ID 2. The surface reflectivity is high in the red wavelength, resulting in high observed intensity in the red channel, but relatively low in other channels. Therefore, in Cond. 2 and 3, the number of reliable channels per pixel is reduced, making it a more challenging object than it appears. The diffuse reflection is dominant, making it poorly compatible with Lv2023~\cite{Lv2023} that rely on specular reflection. The proposed method shows stable recovery even under such challenging conditions.}
    \label{fig:output2}
    \vspace{-5pt}
\end{figure}

\begin{figure}[t]
    \begin{center}
    \includegraphics[width=120mm]{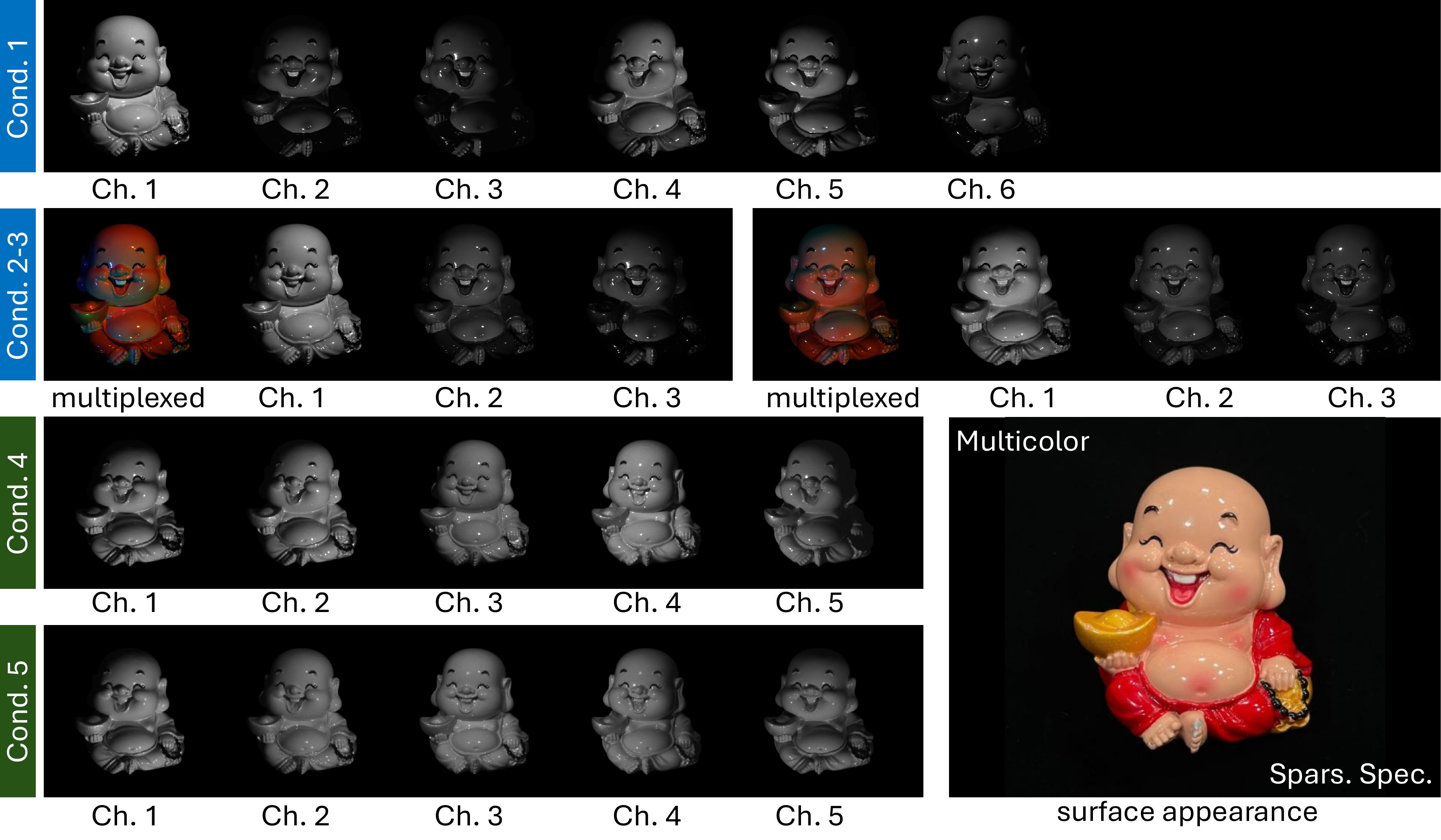}
    \end{center}
    \vspace{-15pt}
    \caption{Input of Object ID 3. A statue of Buddha. The material is glossy and reflective, indicative of a glazed ceramic finish. It's painted in vibrant colors, with the figure dressed in a bright red robe and holding what appears to be a gold ingot. The figure is seated, with exposed belly and feet, adding to the intricate detail of the statuette.}
    \label{fig:input3}
    \vspace{-5pt}
    \begin{center}
    \includegraphics[width=120mm]{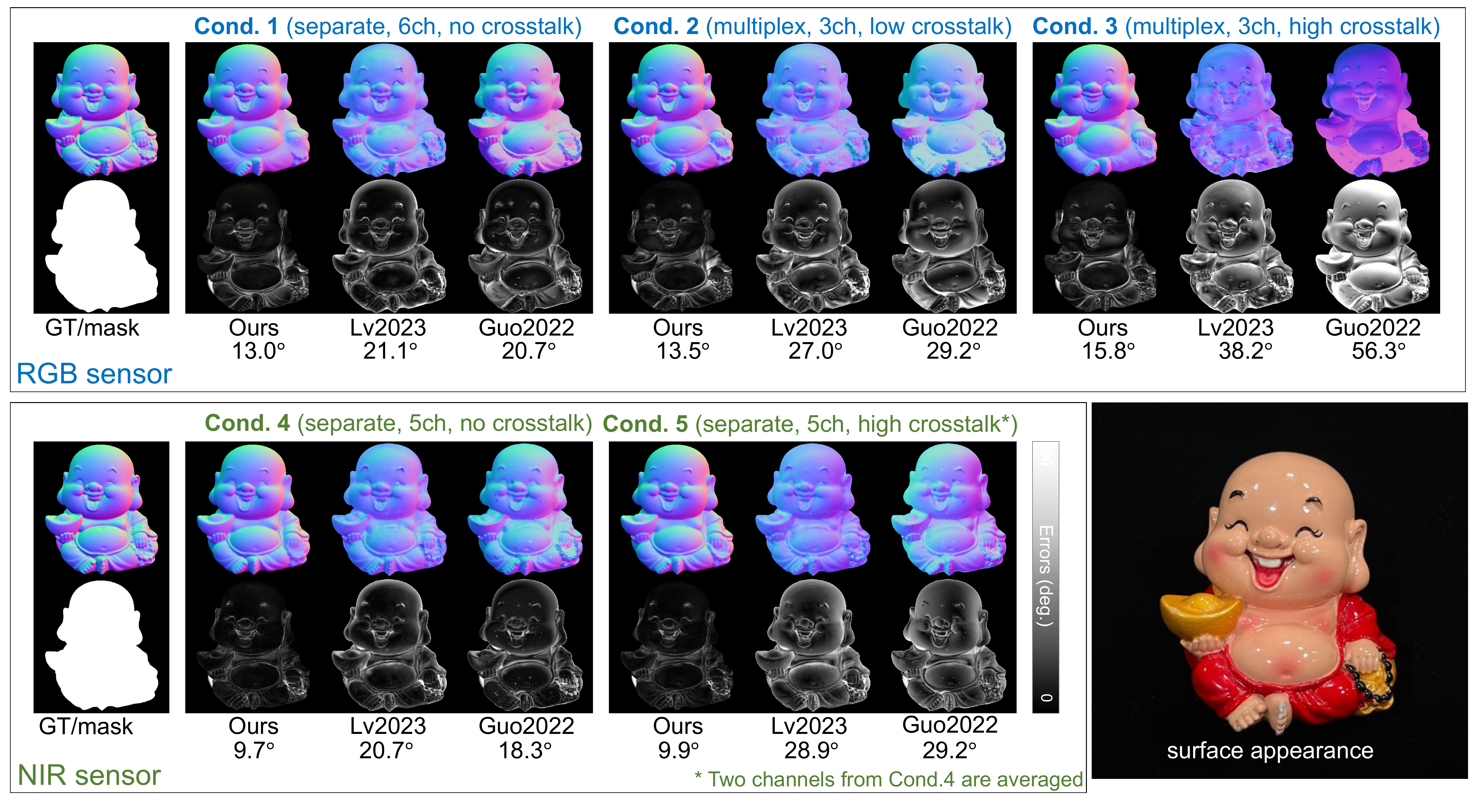}
    \end{center}
    \vspace{-15pt}
    \caption{Output of Object ID 3. The surface has a high red reflectance, appearing glossy. Its concave shape enhances shadows and reflections. Contrary to expectations, Lv2023~\cite{Lv2023} underperforms compared to Guo2022~\cite{Guo2022} for surfaces with specular reflections. The proposed method sees a minor drop in red area accuracy across different conditions with a color sensor, but using an NIR sensor improves results due to high reflectance. Prior methods~\cite{Lv2023,Guo2022} struggle with inter-reflections and cast shadows, leading to accuracy loss.}
    \label{fig:output3}
    \vspace{-5pt}
\end{figure}

\begin{figure}[t]
	\begin{center}
		\includegraphics[width=120mm]{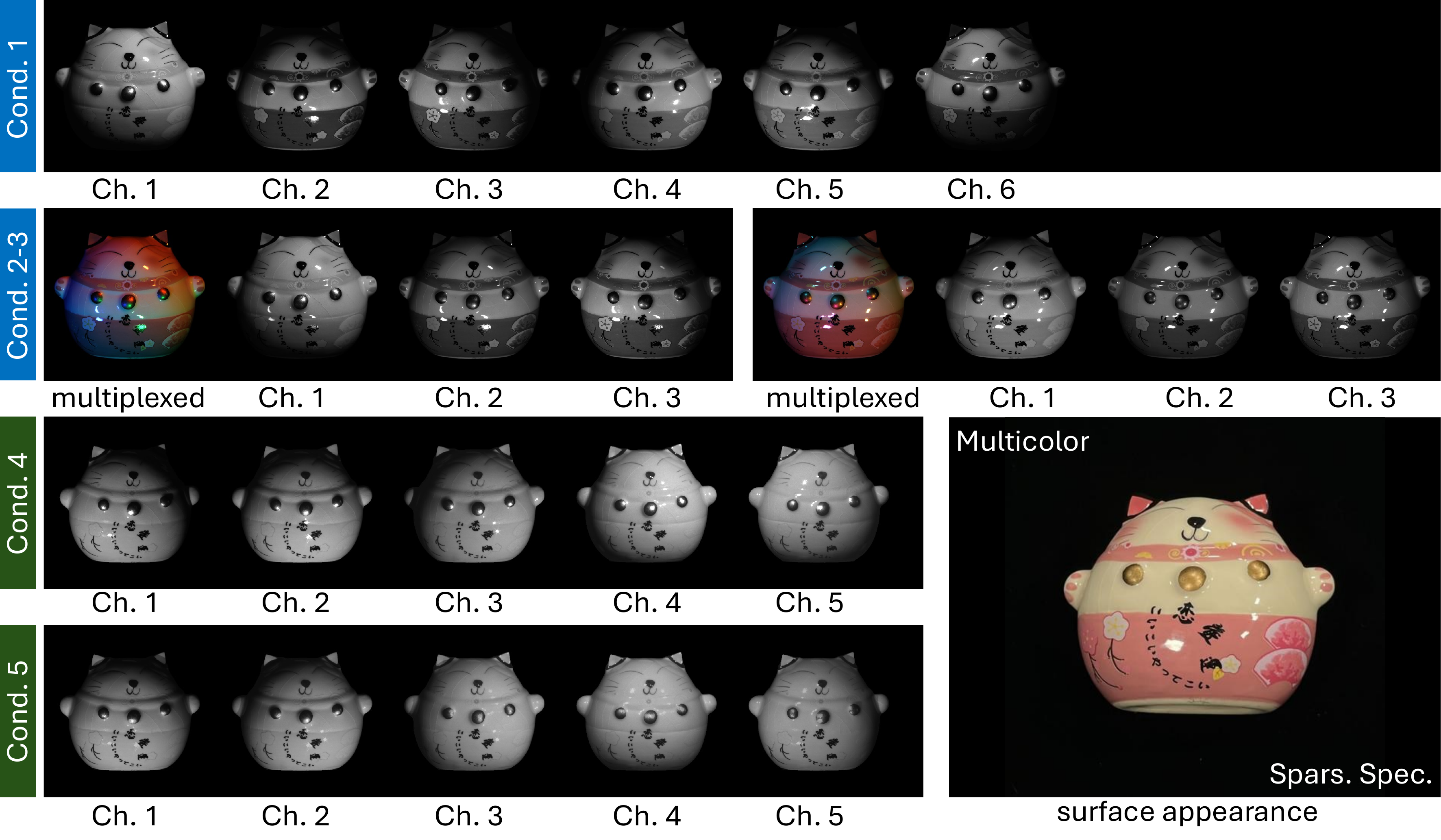}
	\end{center}
	\vspace{-15pt}
	\caption{Input of Object ID 4. A ceramic cat. It has a glossy finish indicative of glazed ceramic. The cat is stylized with a rounded, simplified form. The colors are soft, with pastel pinks and whites, and there are golden accents on the ears, paws, and a medallion on its chest. 
}
	\label{fig:input4}
	\vspace{-5pt}
        \begin{center}
		\includegraphics[width=120mm]{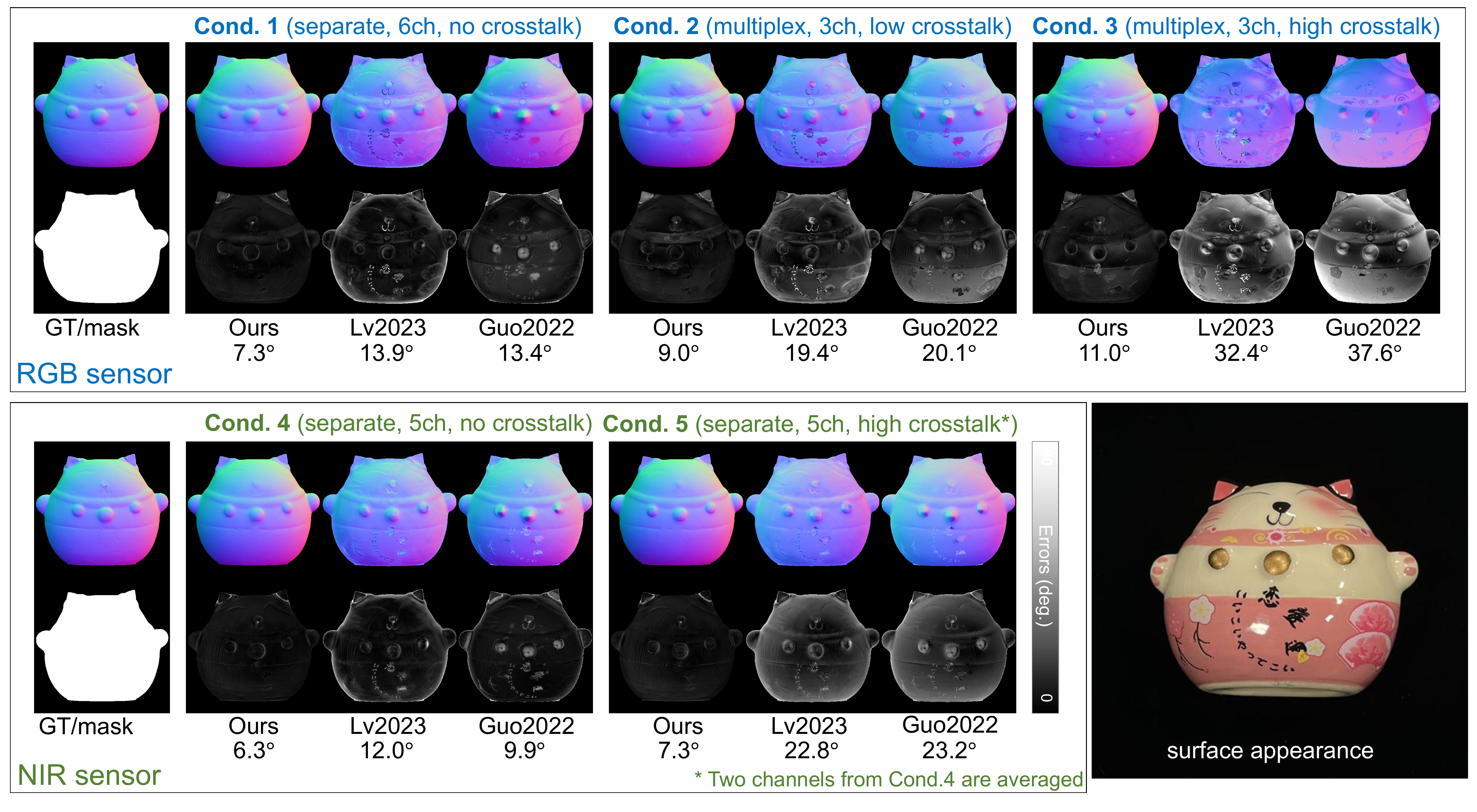}
	\end{center}
	\vspace{-15pt}
	\caption{Output of Object ID 4. This object is characterized by a discrepancy between the continuity of the geometry and material. Networks trained to align texture boundaries with geometric boundaries should produce artifacts on such objects. The proposed method achieves very stable recovery of surface normals under all conditions except for a slight error increase in Cond. 2 and 3. On the other hand, Lv2023~\cite{Lv2023} and Guo2022~\cite{Guo2022} struggle significantly with texture boundaries, showing the detrimental effects of their assumption of uniform material.}
	\label{fig:output4}
	\vspace{-5pt}  
\end{figure}

\begin{figure}[t]
	\begin{center}
		\includegraphics[width=120mm]{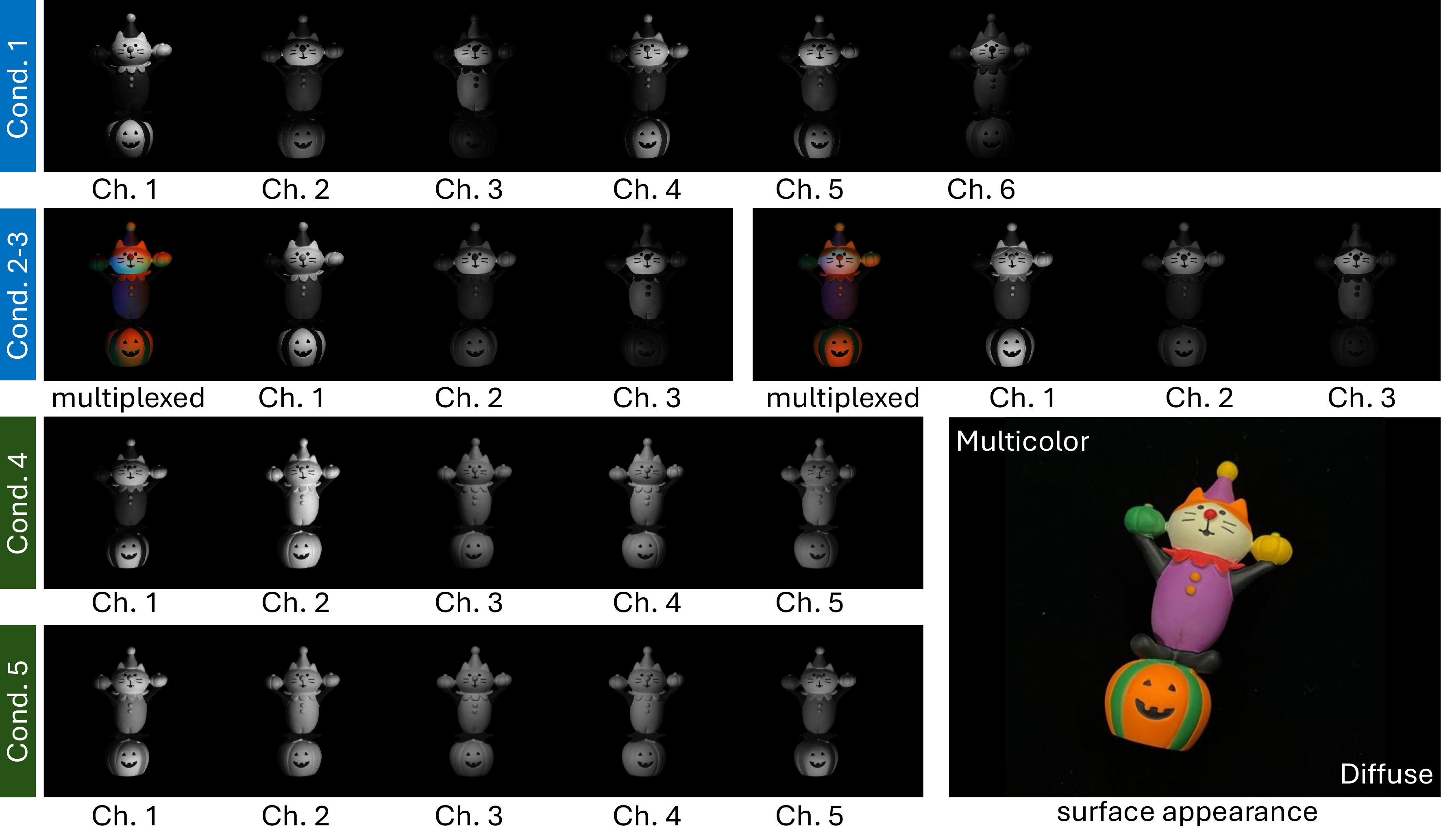}
	\end{center}
	\vspace{-15pt}
	\caption{Input of Object ID 5. A plastic figurine depicting a cat dressed as a clown, set in a playful pose atop a pumpkin. The cat is adorned with a clown's hat and collar, painted in vibrant colors such as purple, green, yellow, and red.}
	\label{fig:input5}
	\vspace{-5pt}
 	\begin{center}
		\includegraphics[width=120mm]{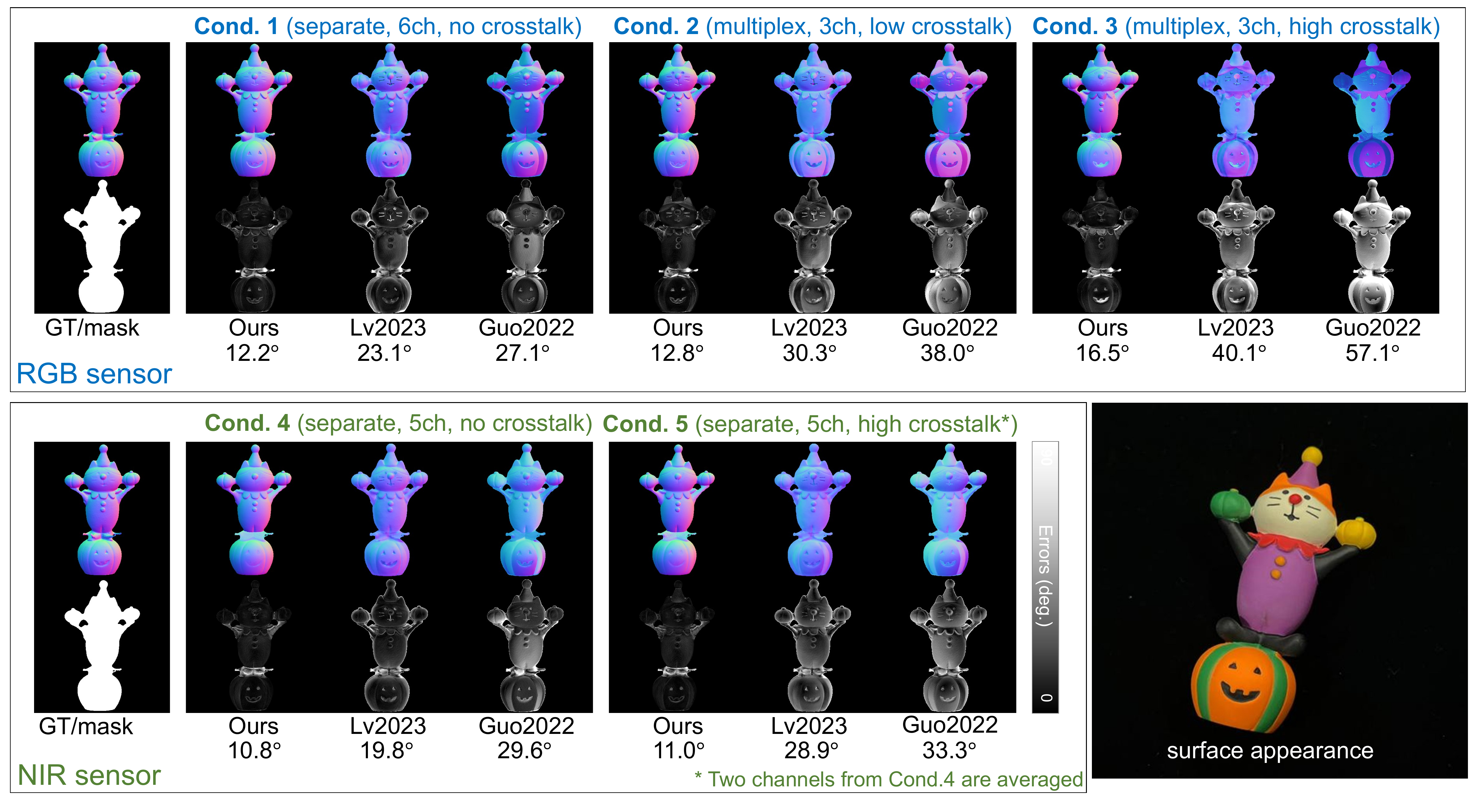}
	\end{center}
	\vspace{-15pt}
    	\caption{Output of Object ID 5. This object poses challenges due to its particularly complex reflectance distribution. The high reflectance in the red and green wavelength ranges, combined with a black surface that lowers the signal-to-noise ratio, and its complex shape, make it one of the most challenging objects among the 14 objects. Indeed, Lv2023~\cite{Lv2023} and Guo2022~\cite{Guo2022} exhibit significant errors under all conditions. In contrast, the proposed method, while experiencing some accuracy degradation in non-convex areas under Cond. 2 and 3, does not encounter significant issues with non-uniform materials.}
	\label{fig:output5}
	\vspace{-5pt}
\end{figure}

\begin{figure}[t]
	
	\begin{center}
		\includegraphics[width=120mm]{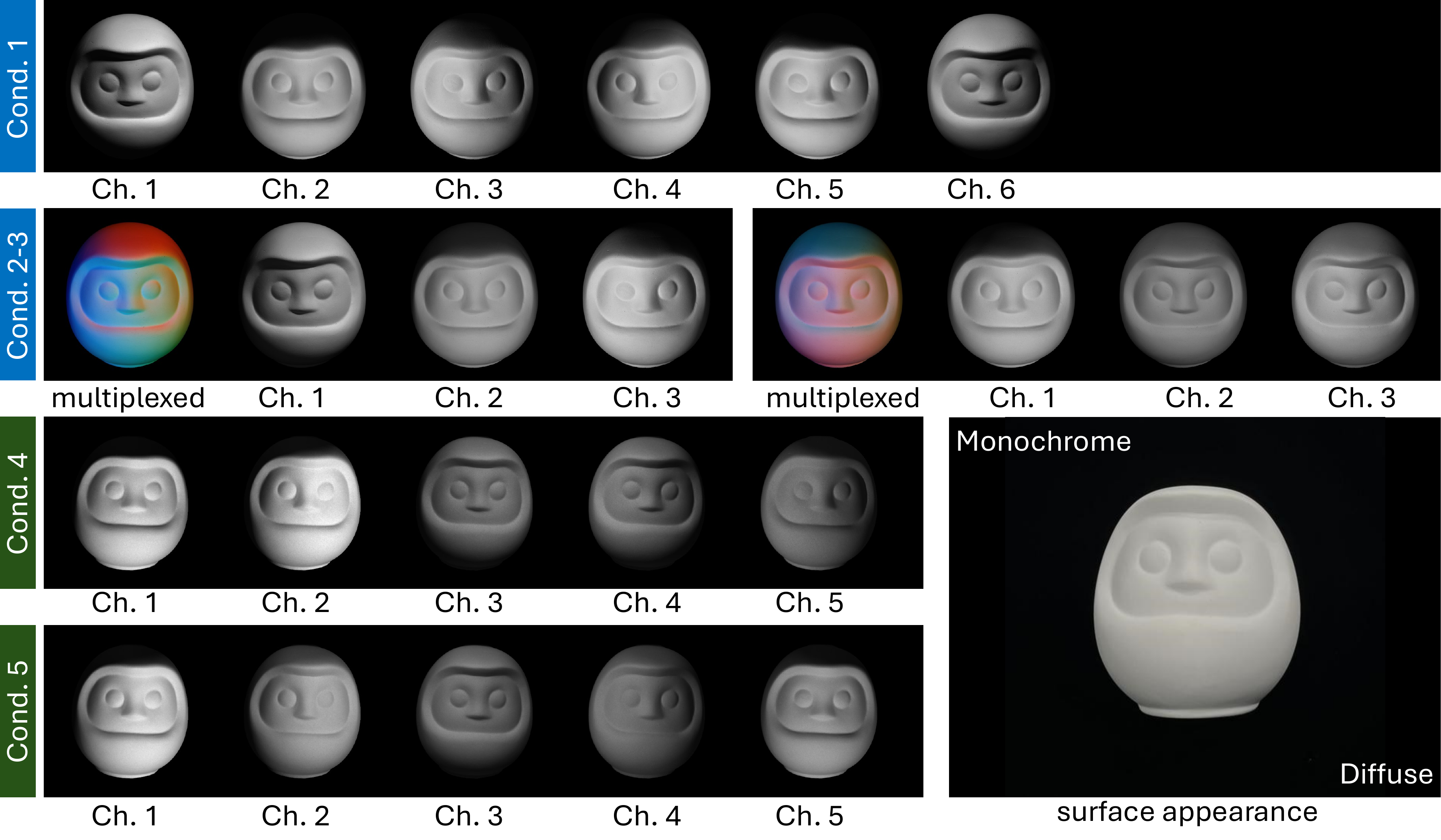}
	\end{center}
	\vspace{-15pt}
	\caption{Input of Object ID 6. A plaster Daruma doll. The doll is characterized by a round shape, uniform white color, and a face with simplistic features. The material, being plaster, has a matte finish.
        }
	\label{fig:input6}
	\vspace{-5pt}
        \begin{center}
		\includegraphics[width=120mm]{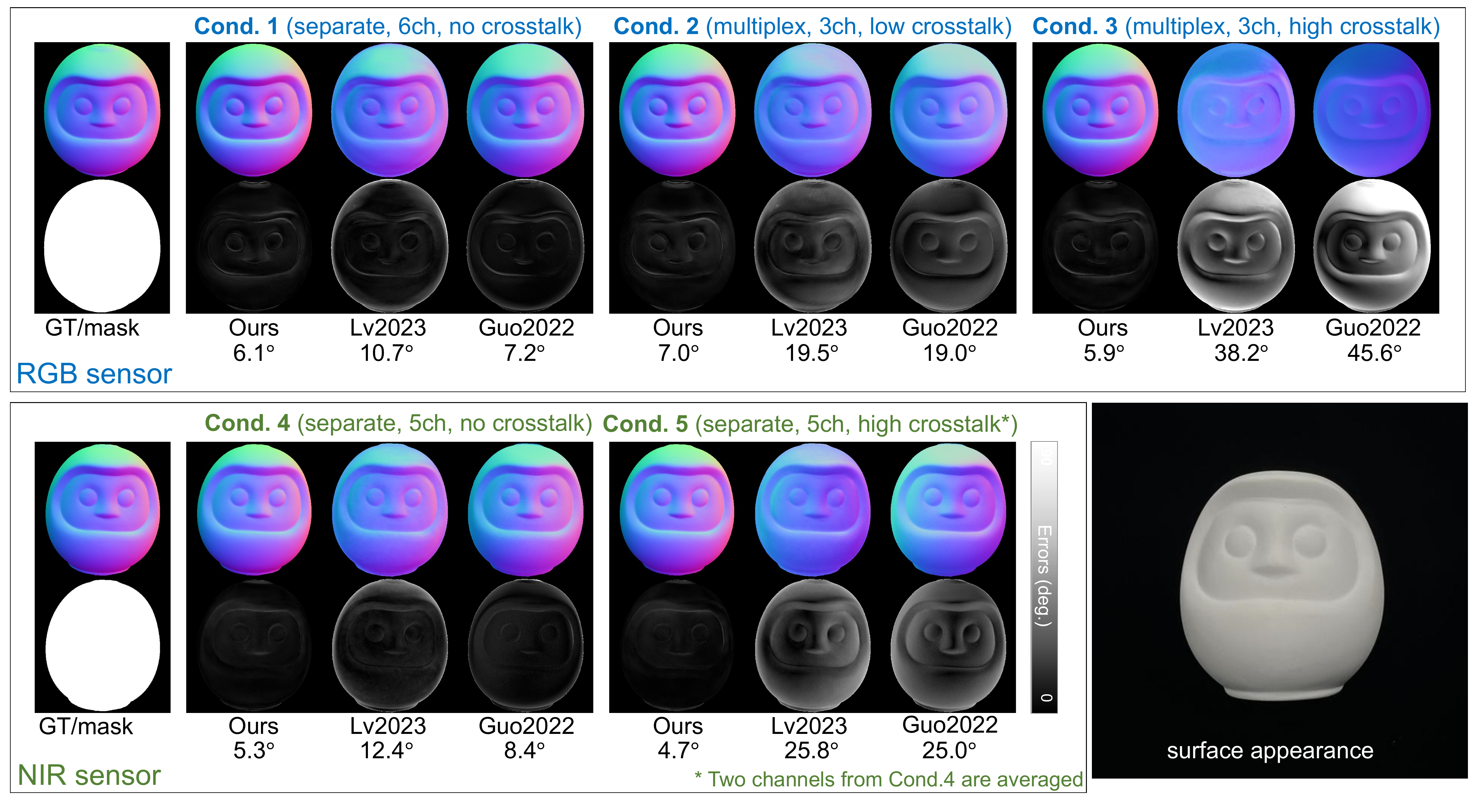}
	\end{center}
	\vspace{-15pt}
	\caption{Output of Object ID 6. This object is ideal for photometric stereo due to its uniform color, Lambertian material, and mostly convex shape. Indeed, prior methods such as Lv2023~\cite{Lv2023} and Guo2022~\cite{Guo2022} demonstrate stable surface normal recovery. However, challenges still arise under conditions with channel crosstalk, such as Conditions 2, 3, and 5. The difficulty in obtaining accurate results, even under relatively ideal circumstances, highlights the potentially unrealistic constraints under which previous methods have been developed.}
	\label{fig:output6}
	\vspace{-5pt}
\end{figure}

\begin{figure}[t]
	
	\begin{center}
		\includegraphics[width=120mm]{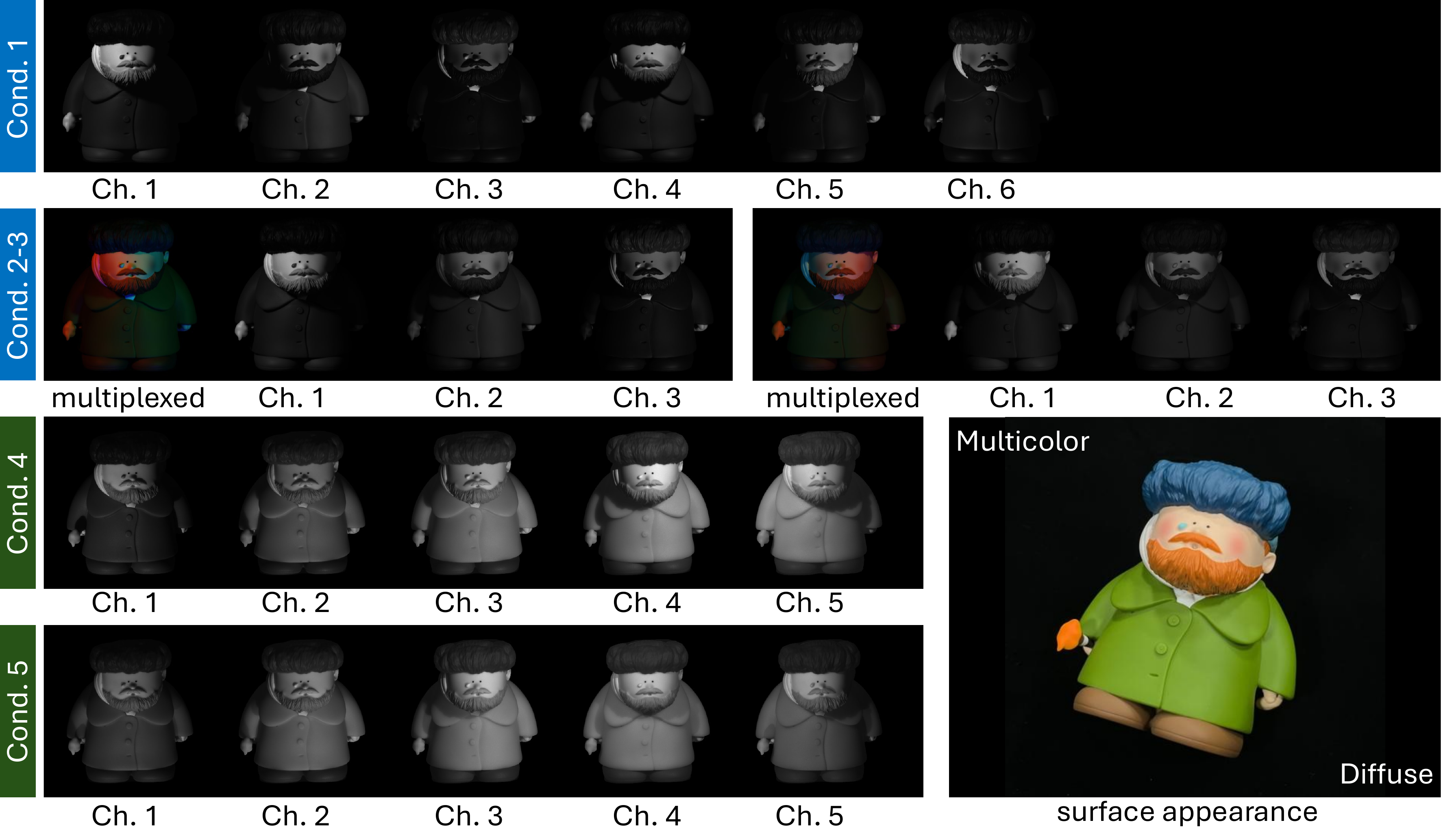}
	\end{center}
	\vspace{-15pt}
	\caption{Input of Object ID 7. A figurine of the artist Vincent van Gogh. This stylized representation features a orange beard, blue hair and a green coat. The material is diffusive.}
	\label{fig:input7}
	\vspace{-5pt}
        \begin{center}
		\includegraphics[width=120mm]{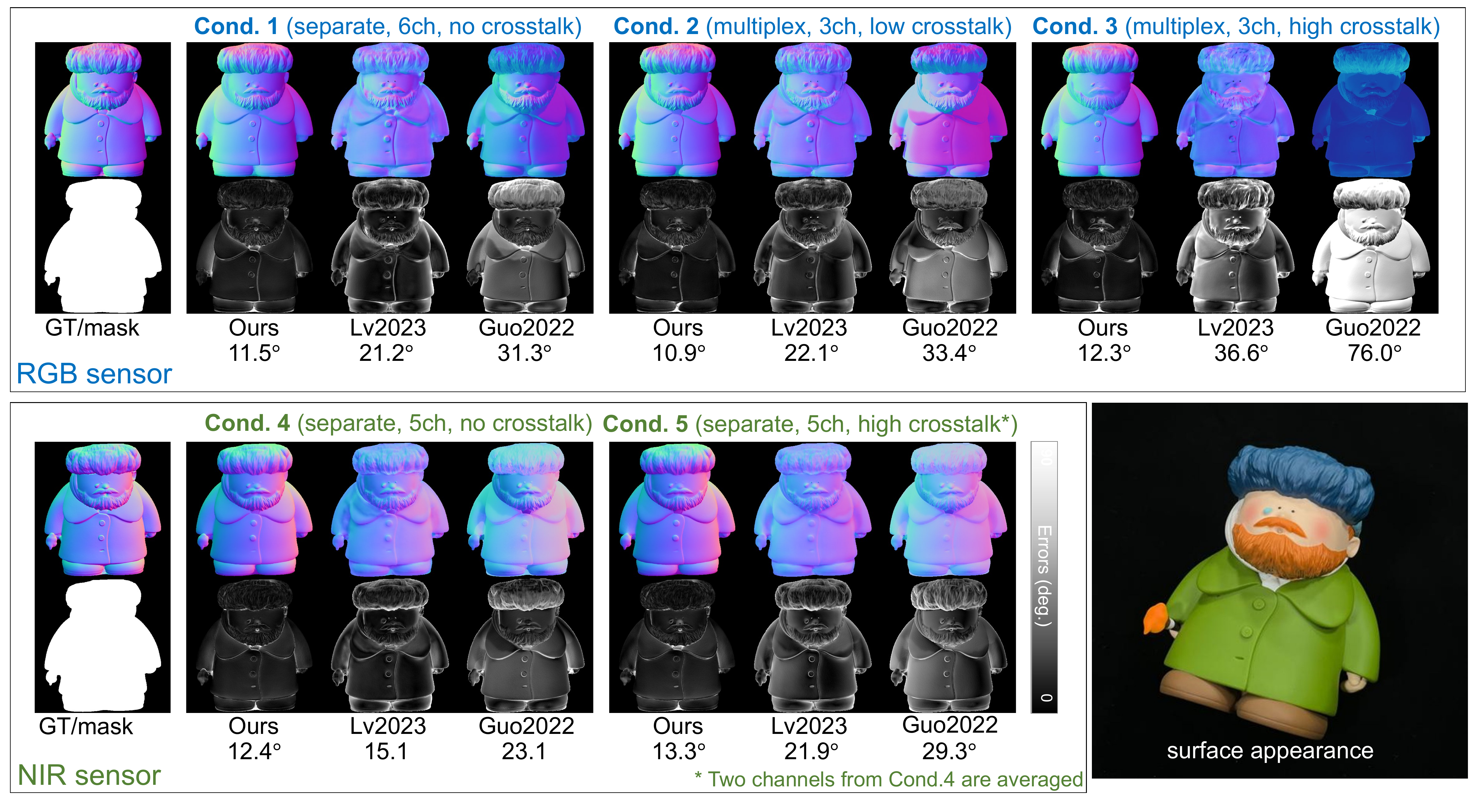}
	\end{center}
	\vspace{-15pt}
	\caption{Output of Object ID 7. At first glance, this object appears to be highly sensitive in the blue and green channels. However, visualization under Cond. 1, 2, and 3 reveals that the reflectivity of the clothes is very low for a color sensor, posing challenges for Lv2023~\cite{Lv2023} and Guo2022~\cite{Guo2022}. While our method successfully recovers the details in the beard and hat, prior methods, such as Lv2023~\cite{Lv2023}, struggle with these fine details.}
	\label{fig:output7}
	\vspace{-5pt}
\end{figure}

\begin{figure}[t]

	\begin{center}
		\includegraphics[width=120mm]{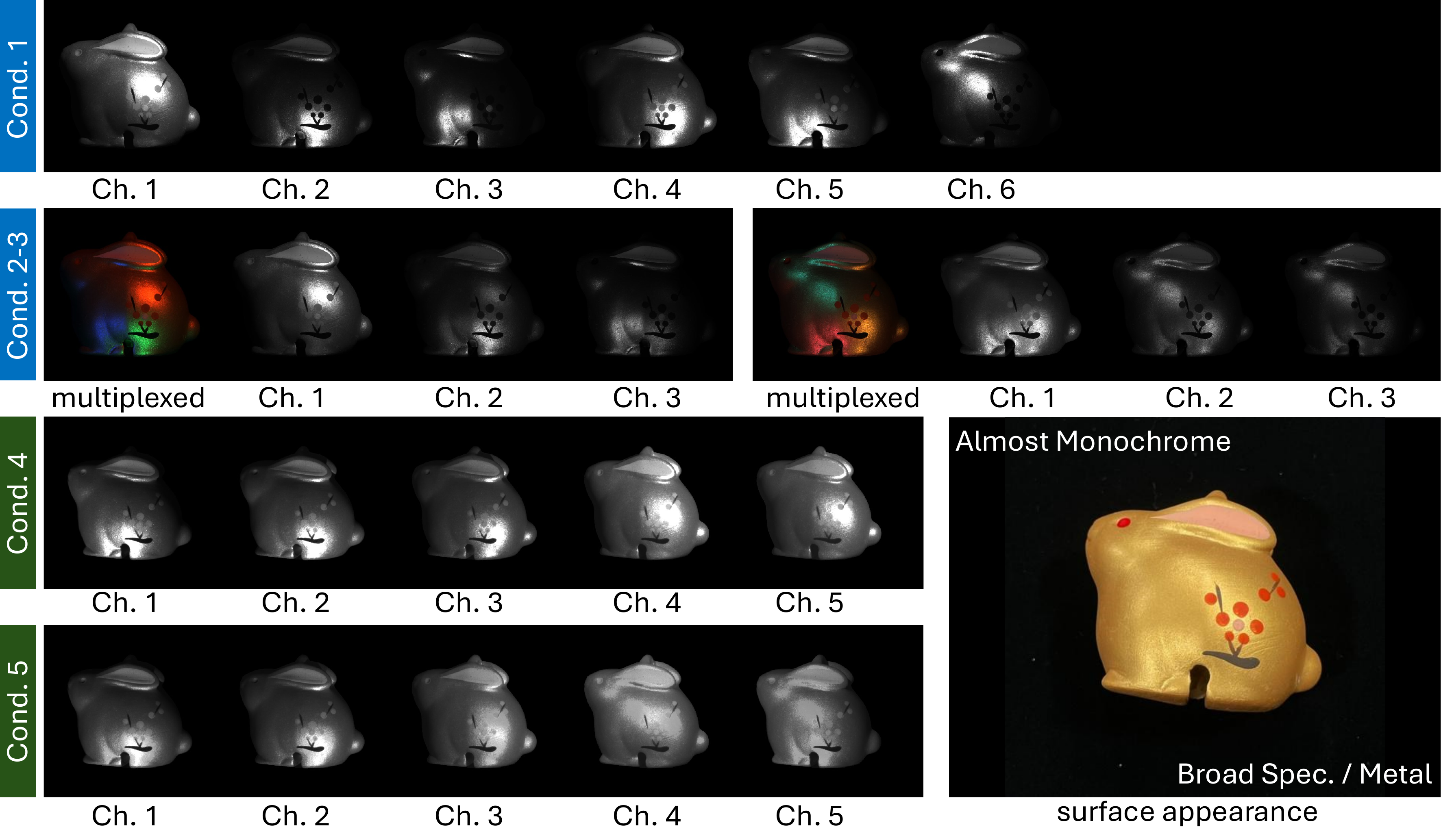}
	\end{center}
	\vspace{-15pt}
	\caption{Input of Object ID 8. A gold-painted rabbit figurine. It features a simplified, stylized form with minimal details. The golden finish results in broad specular reflections. 
}
	\label{fig:input8}
	\vspace{-5pt}
 	\begin{center}
		\includegraphics[width=120mm]{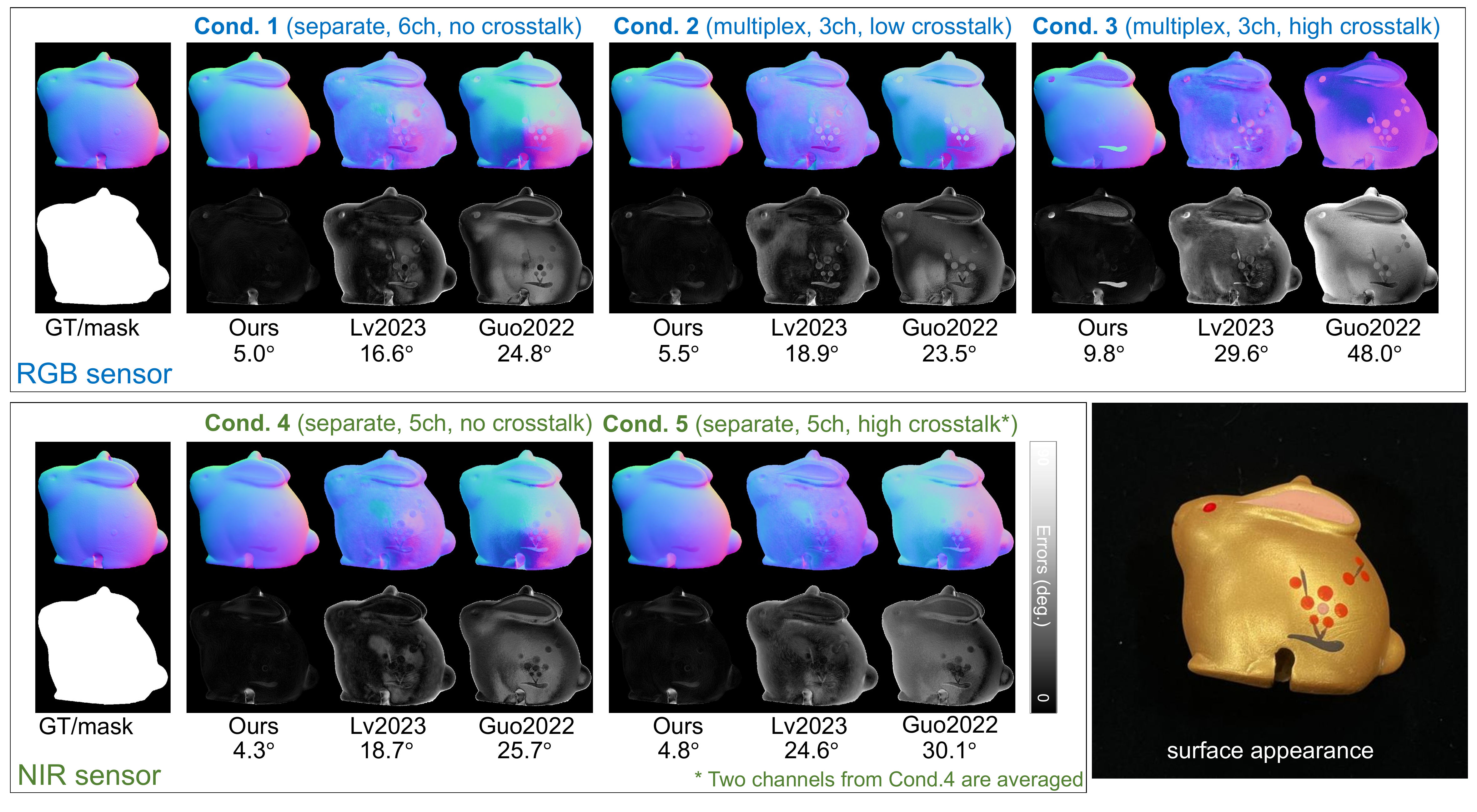}
	\end{center}
	\vspace{-15pt}
	\caption{Output of Object ID 8. The material is almost uniform, yet slight non-uniformity is observed due to the floral pattern. In Cond. 1, 4, and 5, where the number of channels is relatively high, our method achieves particularly high reconstruction accuracy among the objects. In spectral multiplexing setups like Cond. 2 and 3 with a color sensor, slight decreases in accuracy are observed in areas where texture changes exist. Notably, due to its non-uniform and non-Lambertian nature, the accurate recovery of surface normals poses challenges for methods like Guo2022~\cite{Guo2022} and even Lv2023~\cite{Lv2023}.
}
	\label{fig:output8}
	\vspace{-5pt}
\end{figure}

\begin{figure}[t]

	\begin{center}
		\includegraphics[width=120mm]{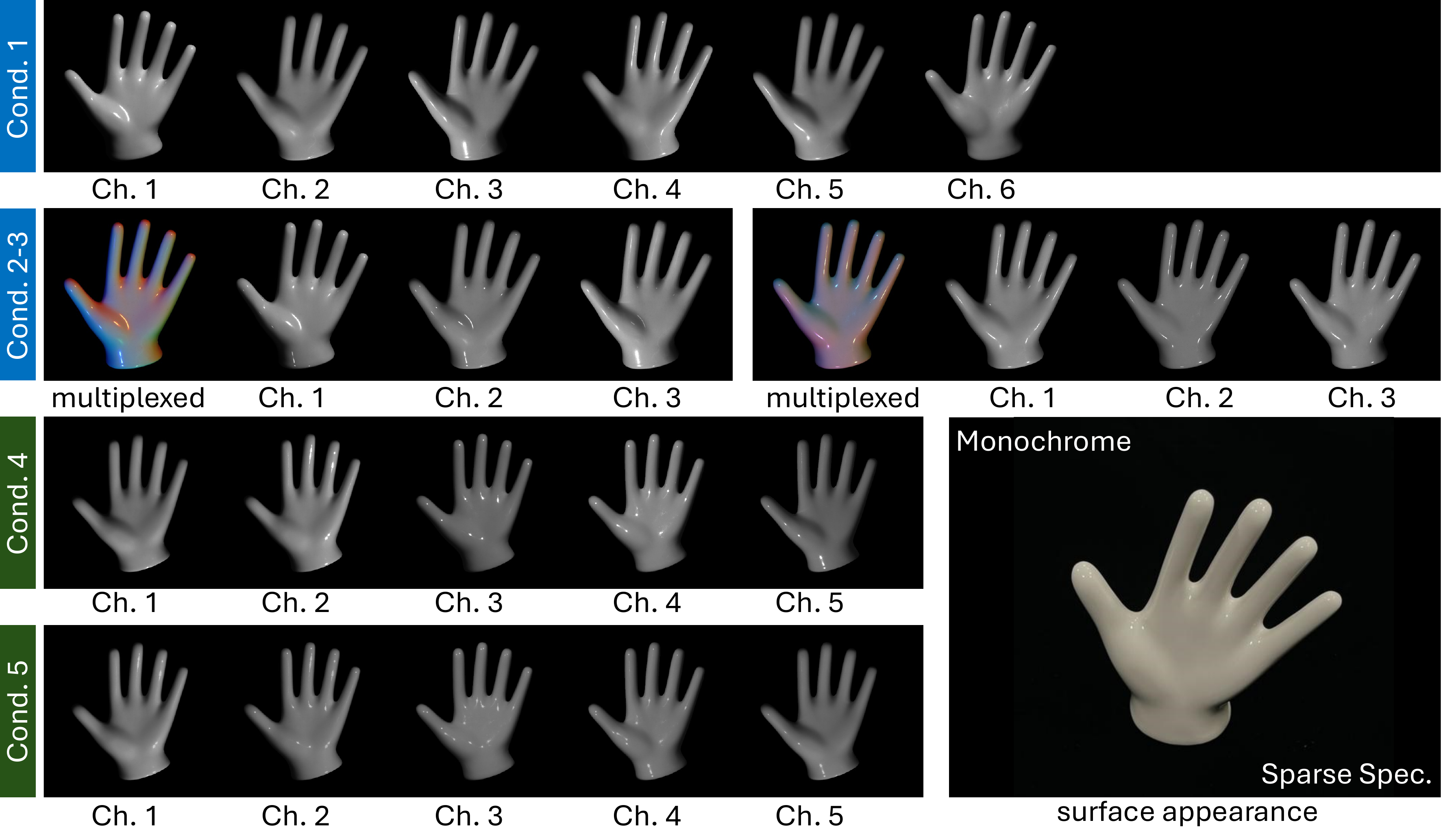}
	\end{center}
	\vspace{-15pt}
	\caption{Input of Object ID 9. A figurine in the shape of a hand. It is made of a white, smooth and glossy ceramic. The hand is displayed in an open position, with fingers slightly spread.}
	\label{fig:input9}
	\vspace{-5pt}
	\begin{center}
		\includegraphics[width=120mm]{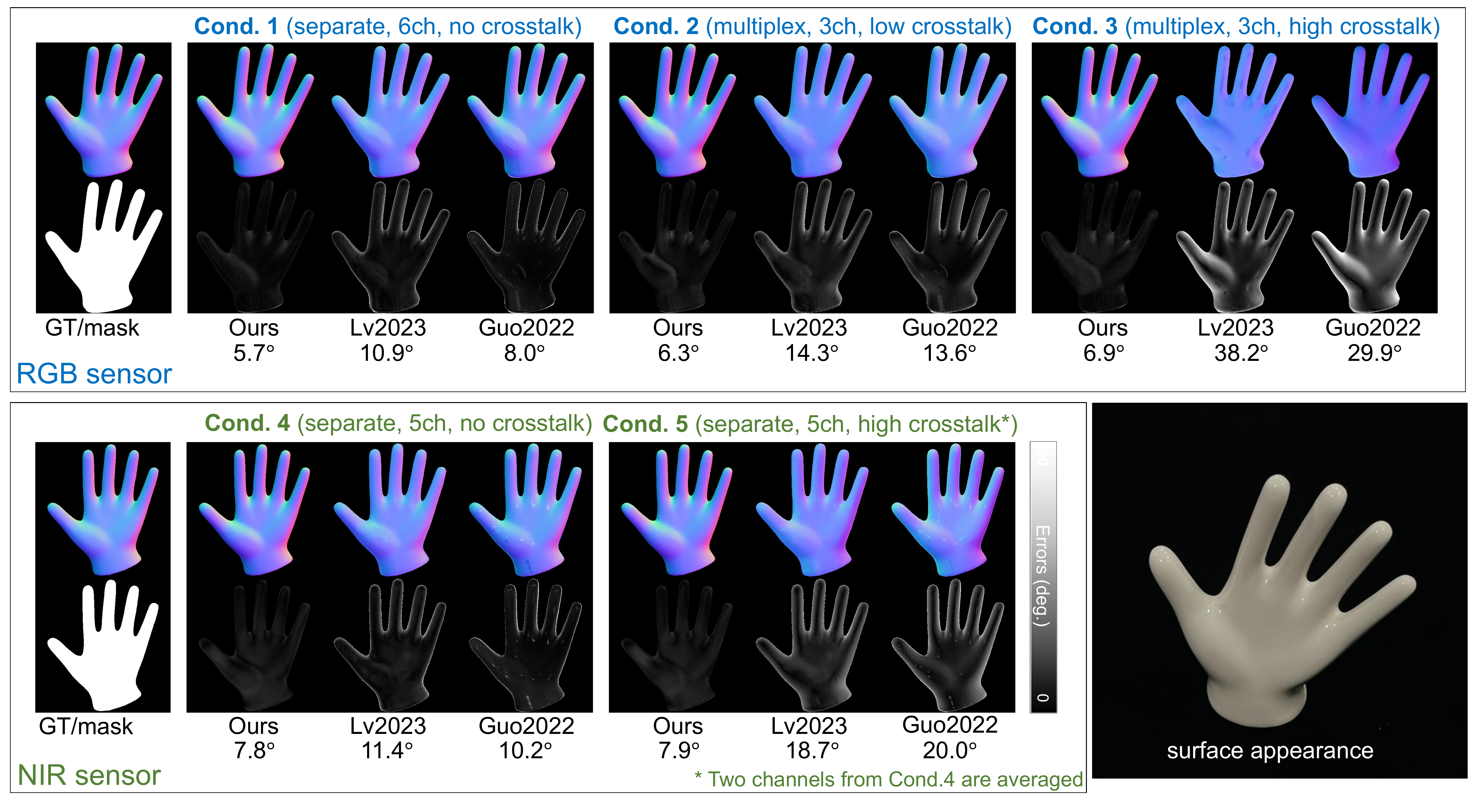}
	\end{center}
	\vspace{-15pt}
        \caption{Output of Object ID 9. This object features a completely uniform material and exhibits peaky specular reflections due to its smooth surface. Not only does the proposed method recover surface normals with reasonable accuracy, but prior methods also perform well if a sufficient number of channels (e.g., 5, 6) are provided (\ie, Cond. 1, 4). However, a clear degradation in accuracy is observed for them in the presence of channel crosstalk, as demonstrated in Cond. 2, 3, and 5.}
	\label{fig:output9}
	\vspace{-5pt}
\end{figure}

\begin{figure}[t]
	\begin{center}
		\includegraphics[width=120mm]{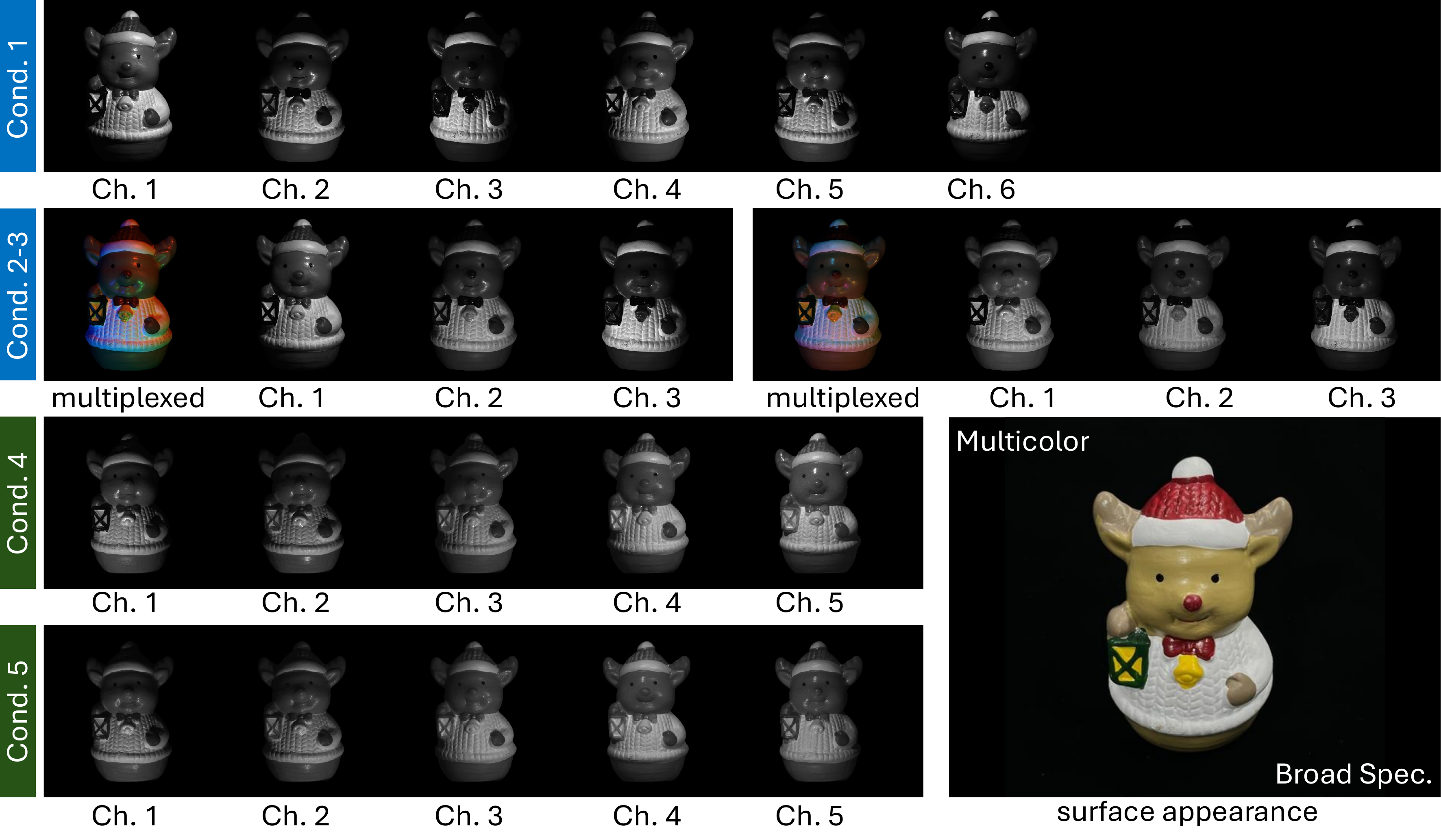}
	\end{center}
	\vspace{-15pt}
	\caption{Input of Object ID 10. A figurine of a reindeer adorned with a red and white Christmas hat. It holds a green ornament and wears a red bow tie. The body of the reindeer is white with a quilted texture, and there's a yellow bell at the center. The material is ceramic with a glossy finish.}
	\label{fig:input10}
	\vspace{-5pt}
 	\begin{center}
		\includegraphics[width=120mm]{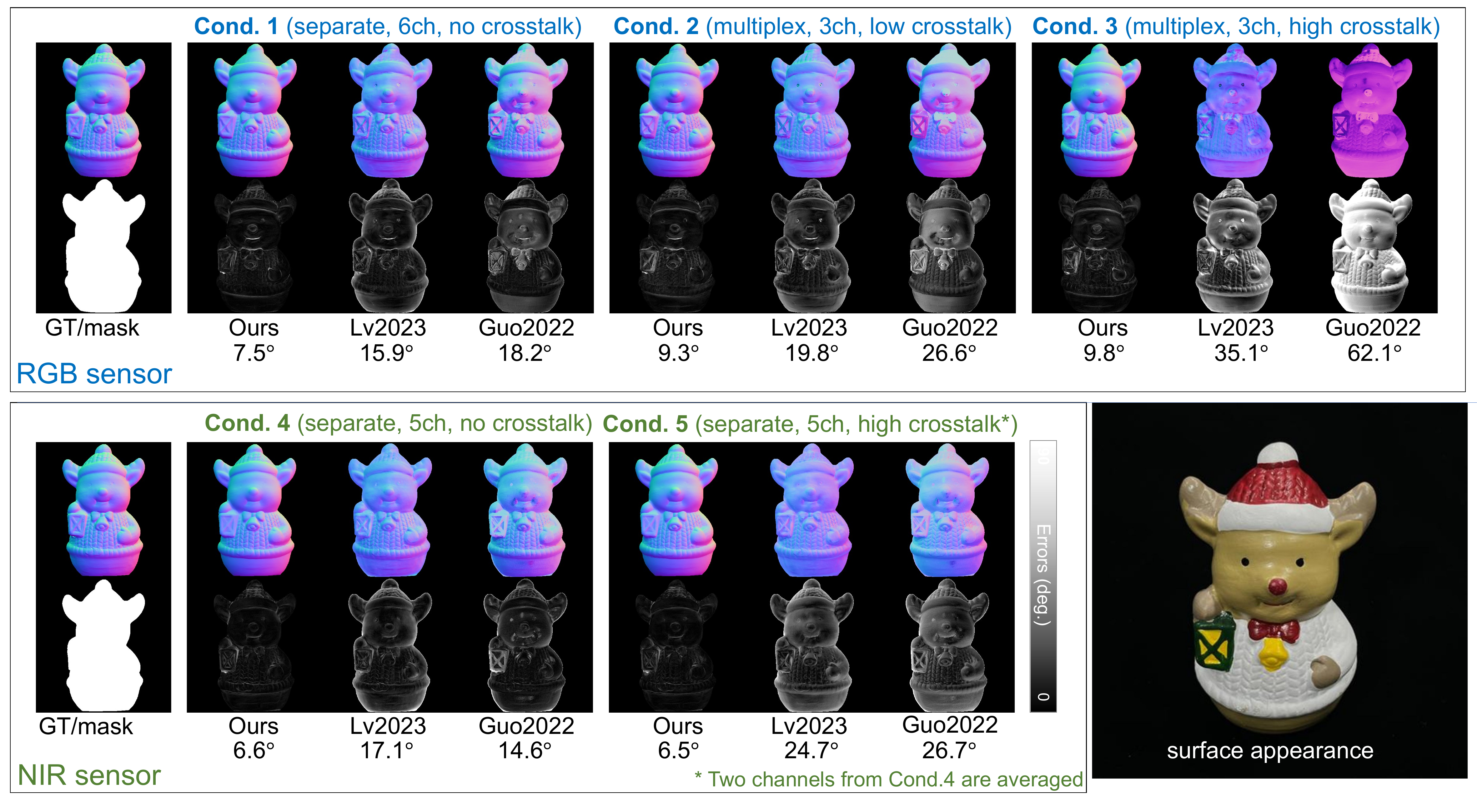}
	\end{center}
	\vspace{-15pt}
        \caption{Output of Object ID 10. This object is distinguished by its complex geometric shape, notably the red hat and the white knitted sweater, and its overall specular reflection. The hat, in particular, shows low brightness values even when observed with NIR sensors, posing a challenge for recovery. However, the proposed method demonstrates good performance even in Cond. 2 and 3, where the effective channels for the hat section are limited. In contrast, prior methods manage to recover relatively well in Cond. 1 and 4, but significant degradation in accuracy is observed under conditions of channel crosstalk.}
	\label{fig:output10}
	\vspace{-5pt}
\end{figure}

\begin{figure}[t]

	\begin{center}
		\includegraphics[width=120mm]{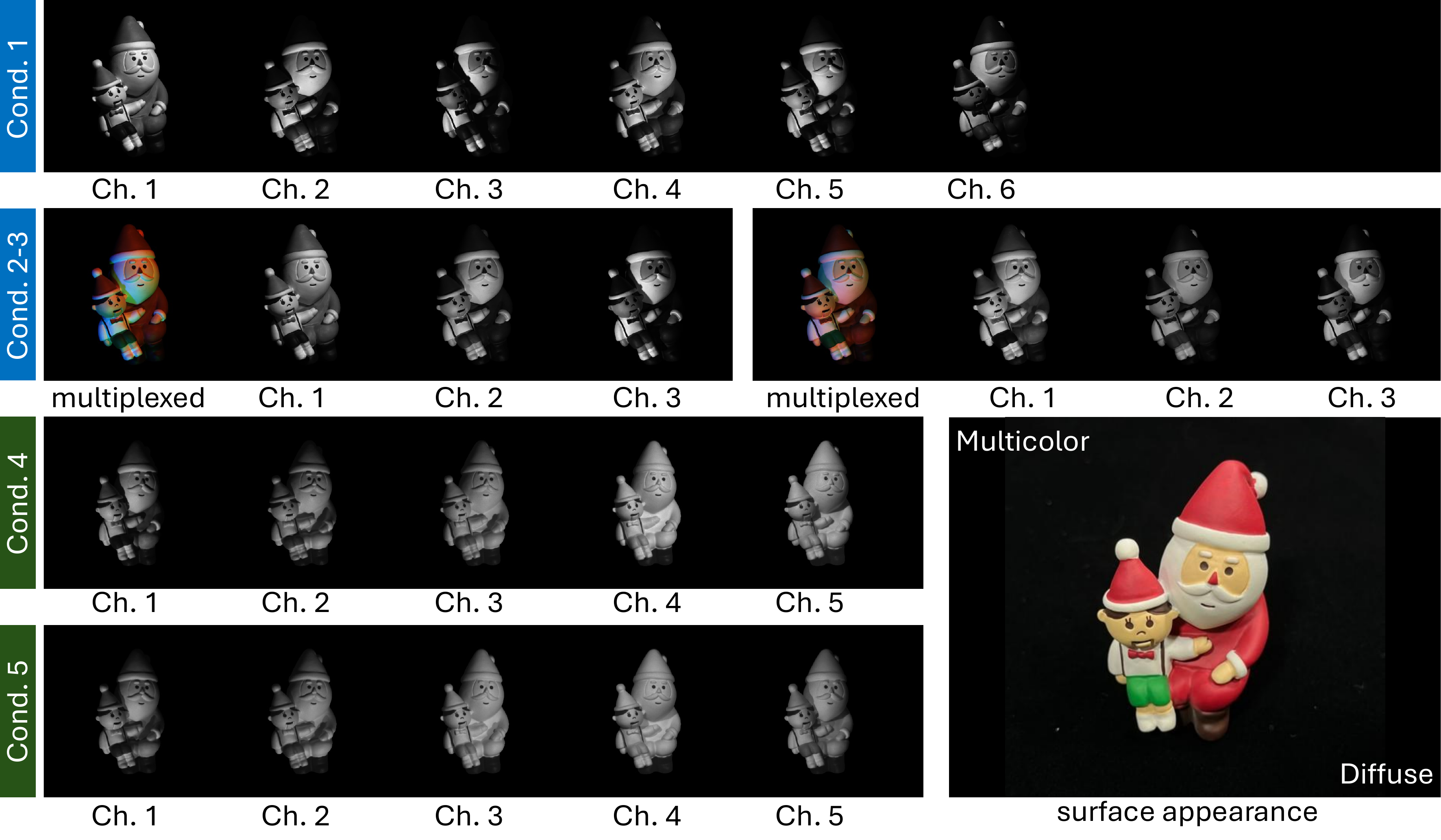}
	\end{center}
	\vspace{-15pt}
	\caption{Input of Object ID 11. A figurine depicting Santa Claus with a child. The materials consist of painted plastic, providing a matte finish and vibrant colors.}
	\label{fig:input11}
	\vspace{-5pt}
 	\begin{center}
		\includegraphics[width=120mm]{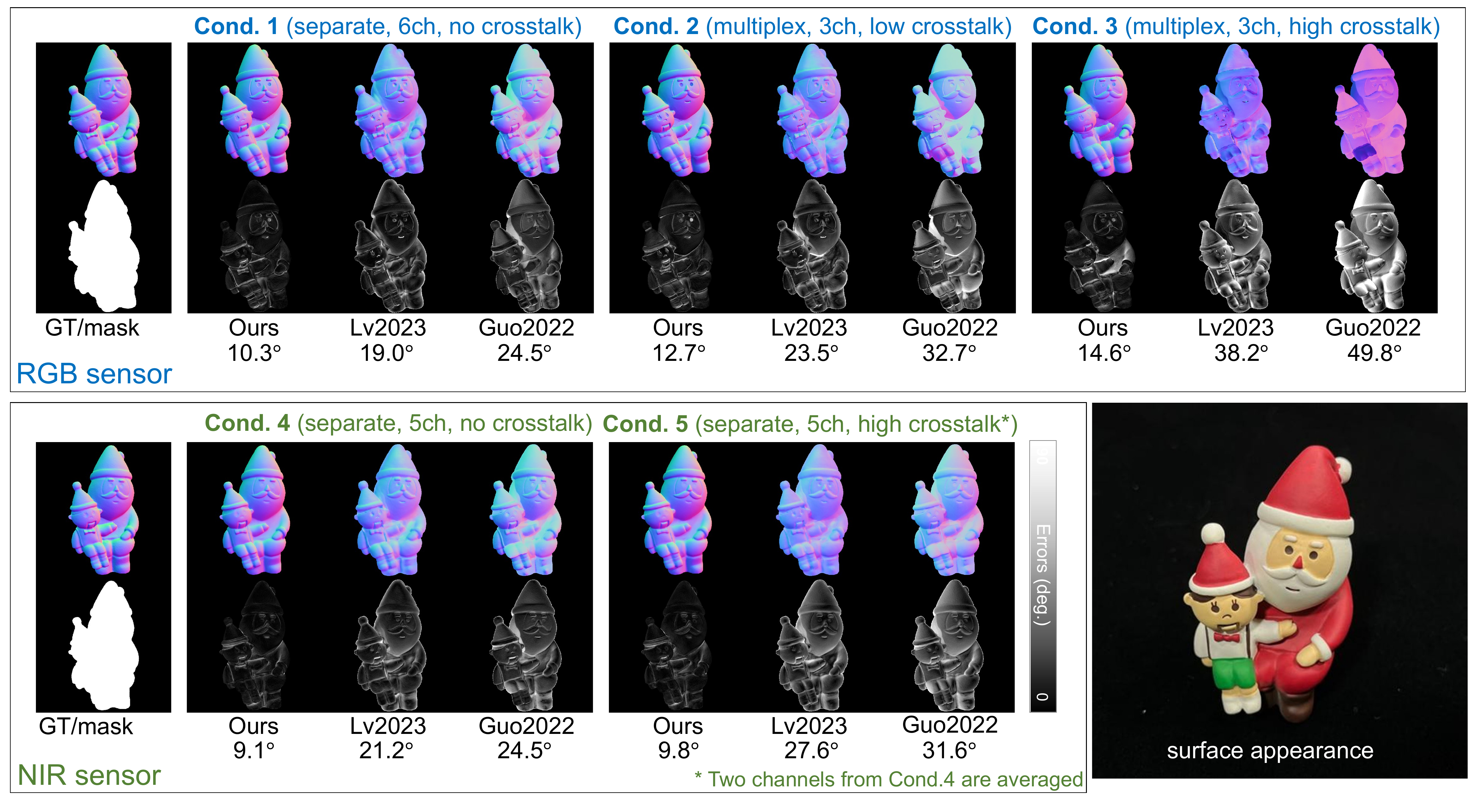}
	\end{center}
	\vspace{-15pt}
	\caption{Output of Object ID 11. This object ranks among the most challenging of the 14 objects due to its non-uniform surface reflectance and its complex non-convex shape. In particular, the knee part in Cond. 2 and 3 poses significant recovery challenges due to the combined effects of limited effective channels and cast shadows from the non-convex shape. The proposed method experiences some accuracy degradation in Cond. 2 and 3 compared to other conditions. However, it still achieves sufficiently accurate results, unlike Lv2023~\cite{Lv2023} and Guo2022~\cite{Guo2022}, which exhibit almost unsatisfactory performance.}
	\label{fig:output11}
	\vspace{-5pt}
\end{figure}

\begin{figure}[t]

        \begin{center}
		\includegraphics[width=120mm]{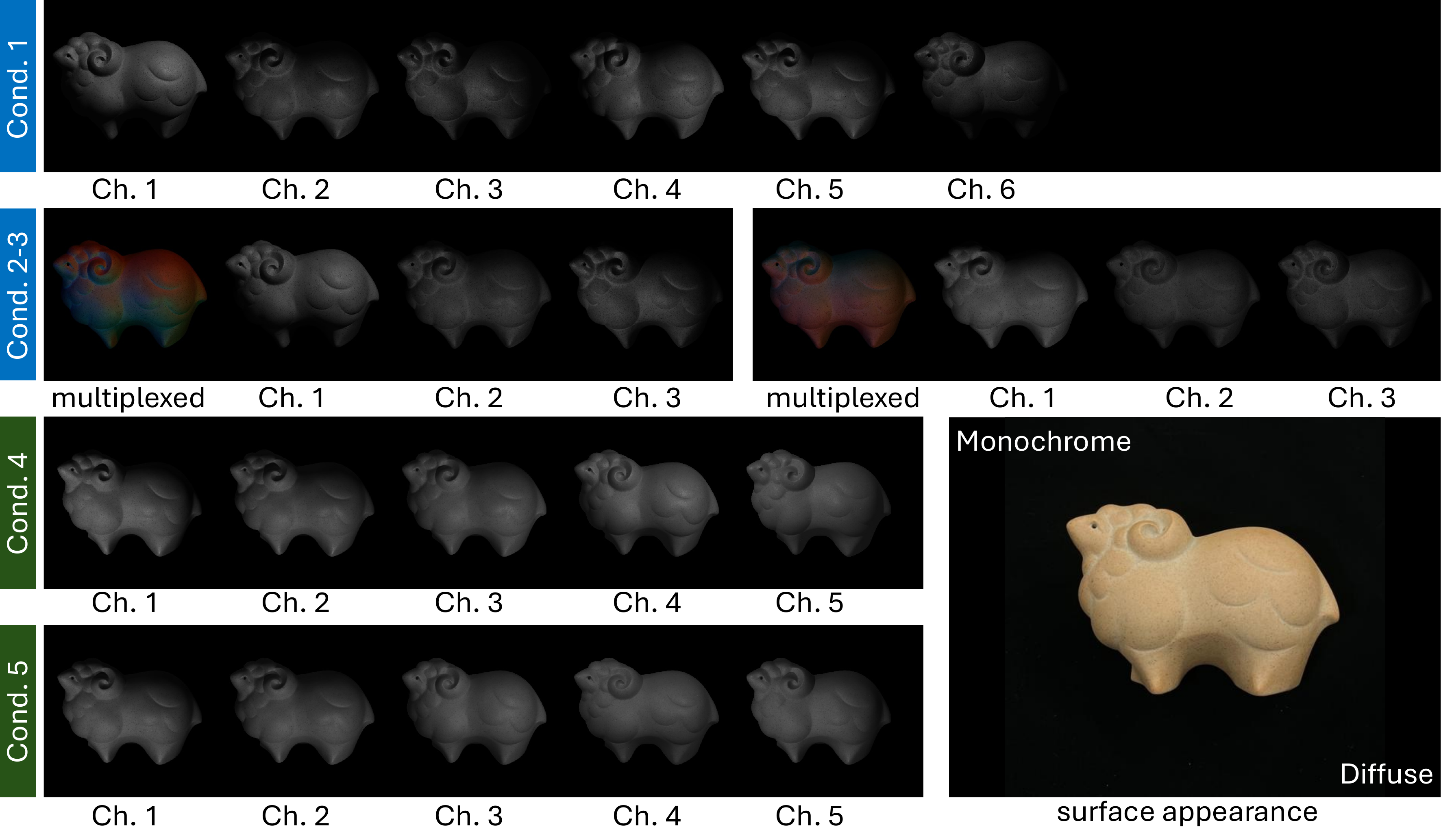}
	\end{center}
	\vspace{-15pt}
	\caption{Input of Object ID 12. A sheep figurine crafted from wood. The sheep's shape is stylized and simplistic, featuring soft curves. The surface material is nearly uniform, though the wood grain is slightly visible.}
	\label{fig:input12}
	\vspace{-5pt}
 	\begin{center}
		\includegraphics[width=120mm]{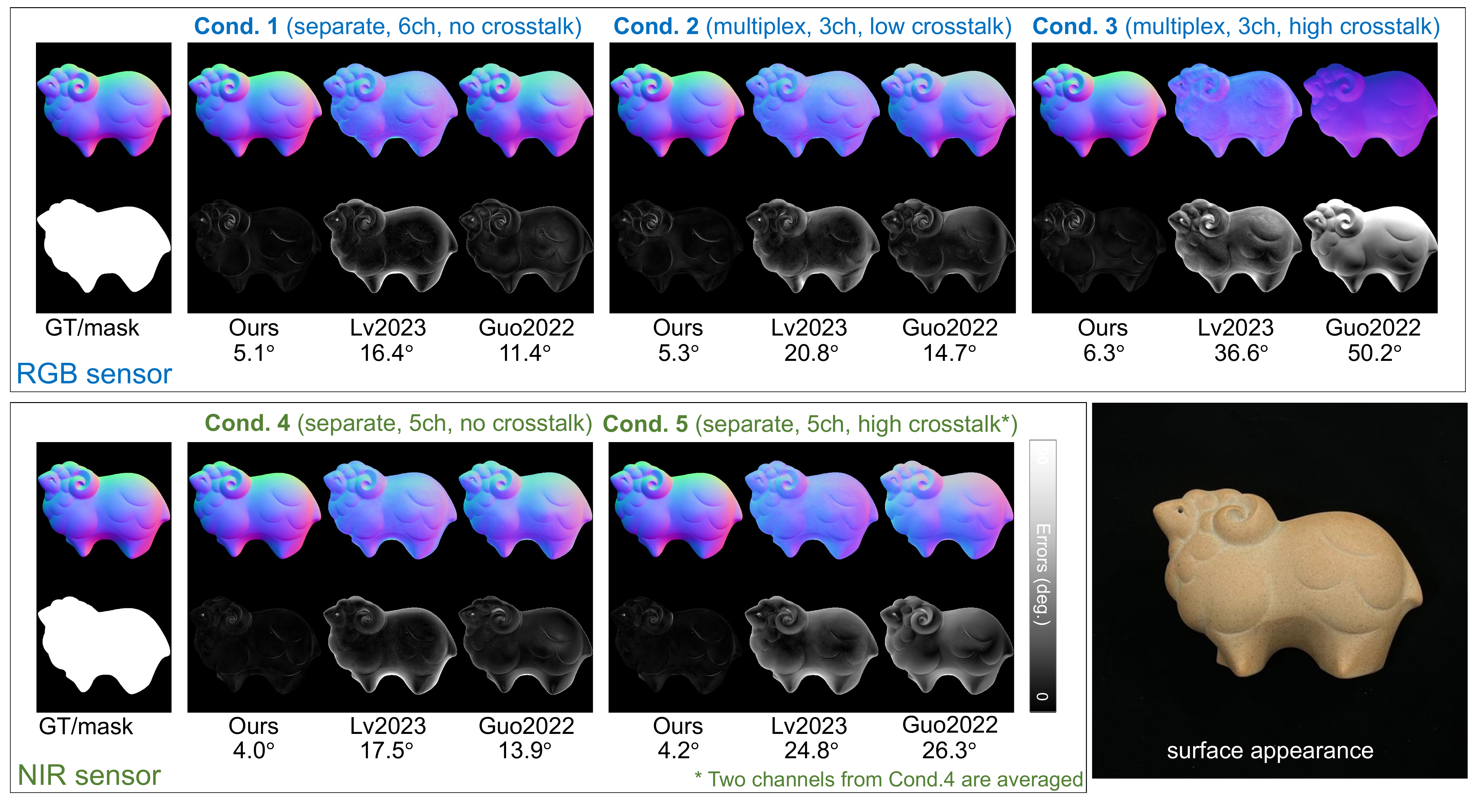}
	\end{center}
	\vspace{-15pt}
	\caption{Output of Object ID 12. This object, a wooden sheep, is characterized by a mostly uniform diffuse surface. Its shape is relatively simple. In fact, in Cond. 1 and 4, even Lv2023~\cite{Lv2023} and Guo2022~\cite{Guo2022} manage to achieve somewhat accurate results. However, they exhibit significant performance degradation under conditions with channel crosstalk. In contrast, the proposed method consistently achieves very accurate recovery under all conditions.}
	\label{fig:output12}
	\vspace{-5pt}
\end{figure}

\begin{figure}[t]

	\begin{center}
		\includegraphics[width=120mm]{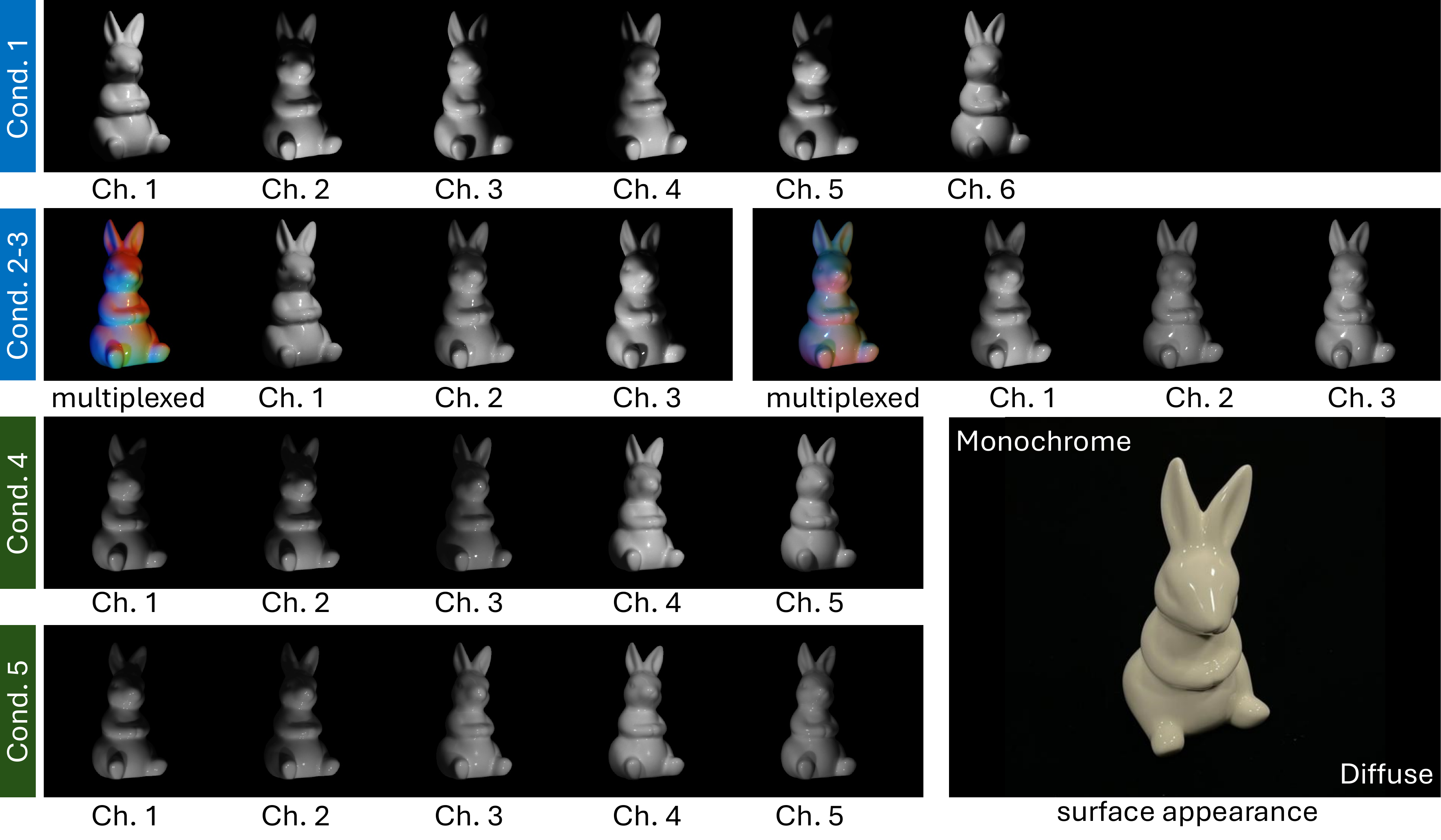}
	\end{center}
	\vspace{-15pt}
	\caption{Input of Object ID 13. A ceramic rabbit figurine featuring a smooth and glossy finish and uniform material. While the design lacks intricate details, it includes some non-convex structures.}
	\label{fig:input13}
	\vspace{-5pt}
 	\begin{center}
		\includegraphics[width=120mm]{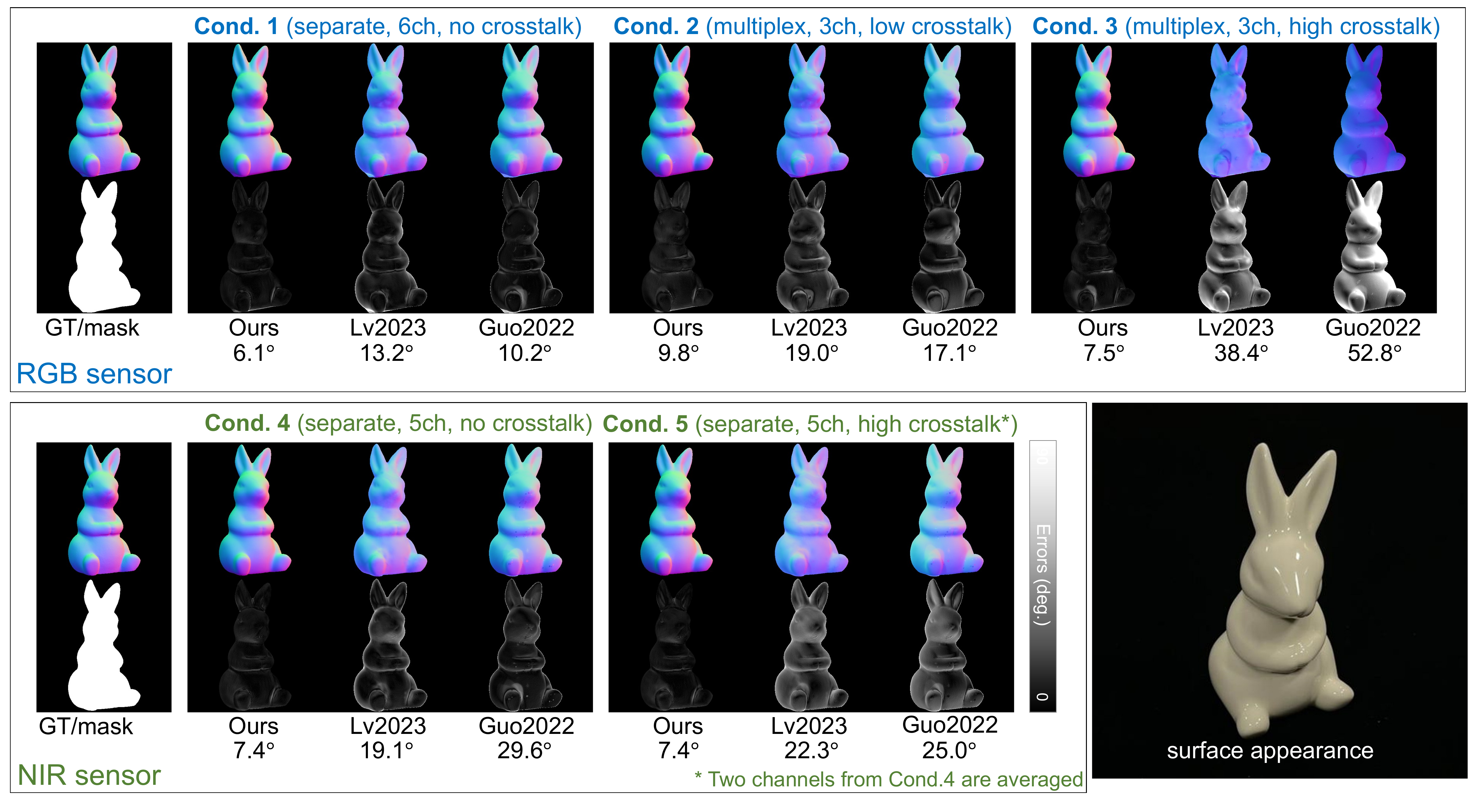}
	\end{center}
	\vspace{-15pt}
	\caption{Output of Object ID 13. This object is made from a material that is almost uniform and exhibits very peaky highlights. The proposed method demonstrated very high recovery performance under all conditions, while prior methods achieved reasonable reconstruction performance in Cond. 1. However, the shape of the object is complex, often producing shadows or interreflections that pose challenges for Lv2023~\cite{Lv2023} and Guo2022~\cite{Guo2022} under other conditions.}
	\label{fig:output13}
	\vspace{-5pt}
\end{figure}

\begin{figure}[t]

	\begin{center}
		\includegraphics[width=120mm]{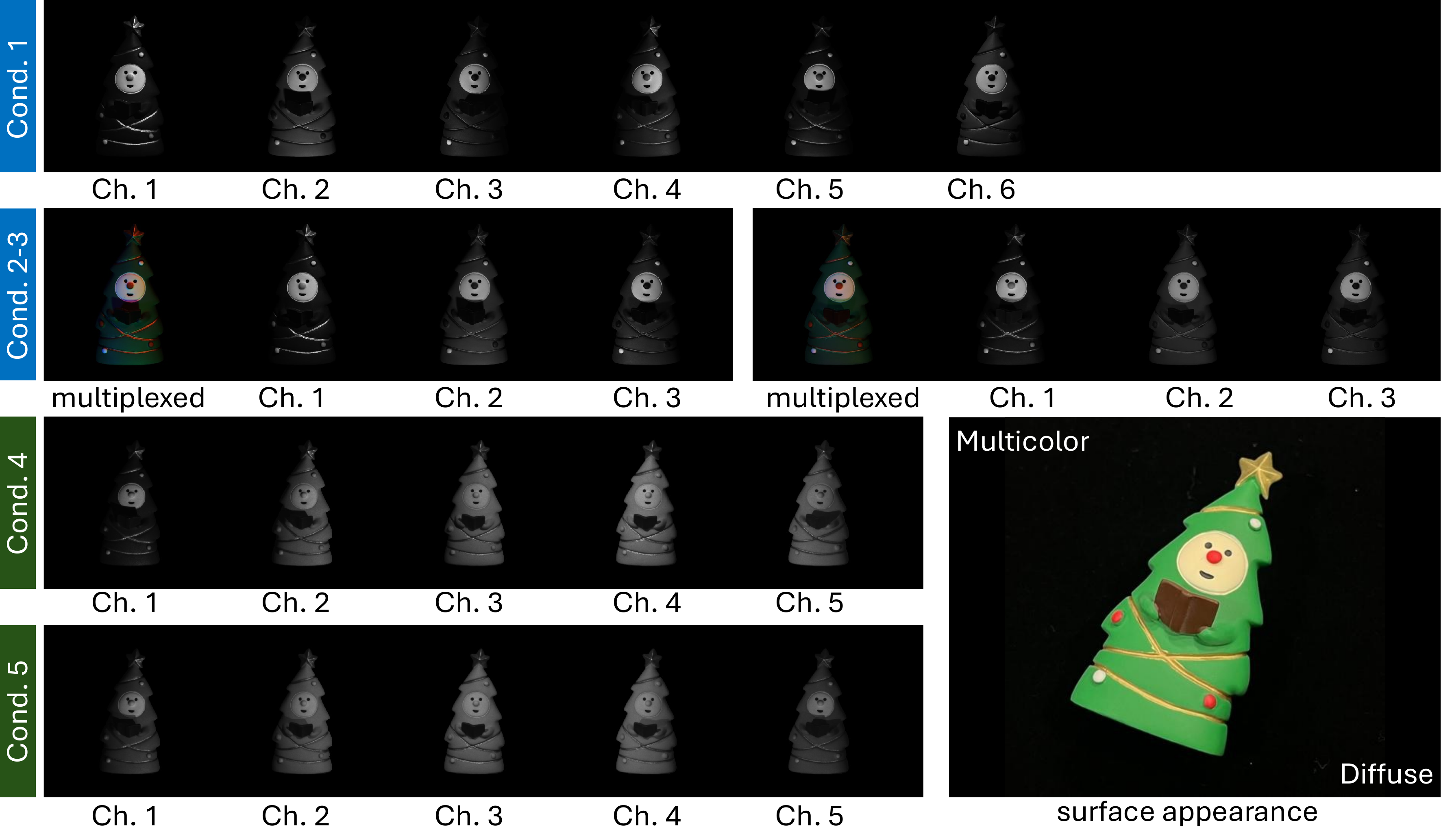}
	\end{center}
	\vspace{-15pt}
	\caption{Input of Object ID 14. A figurine depicting a Christmas tree with a face, stylized with a green body. It features decorative elements such as red ornaments and gold trim, topped with a star. The object is crafted from painted plastic.}
	\label{fig:input14}
	\vspace{-5pt}
 	\begin{center}
		\includegraphics[width=120mm]{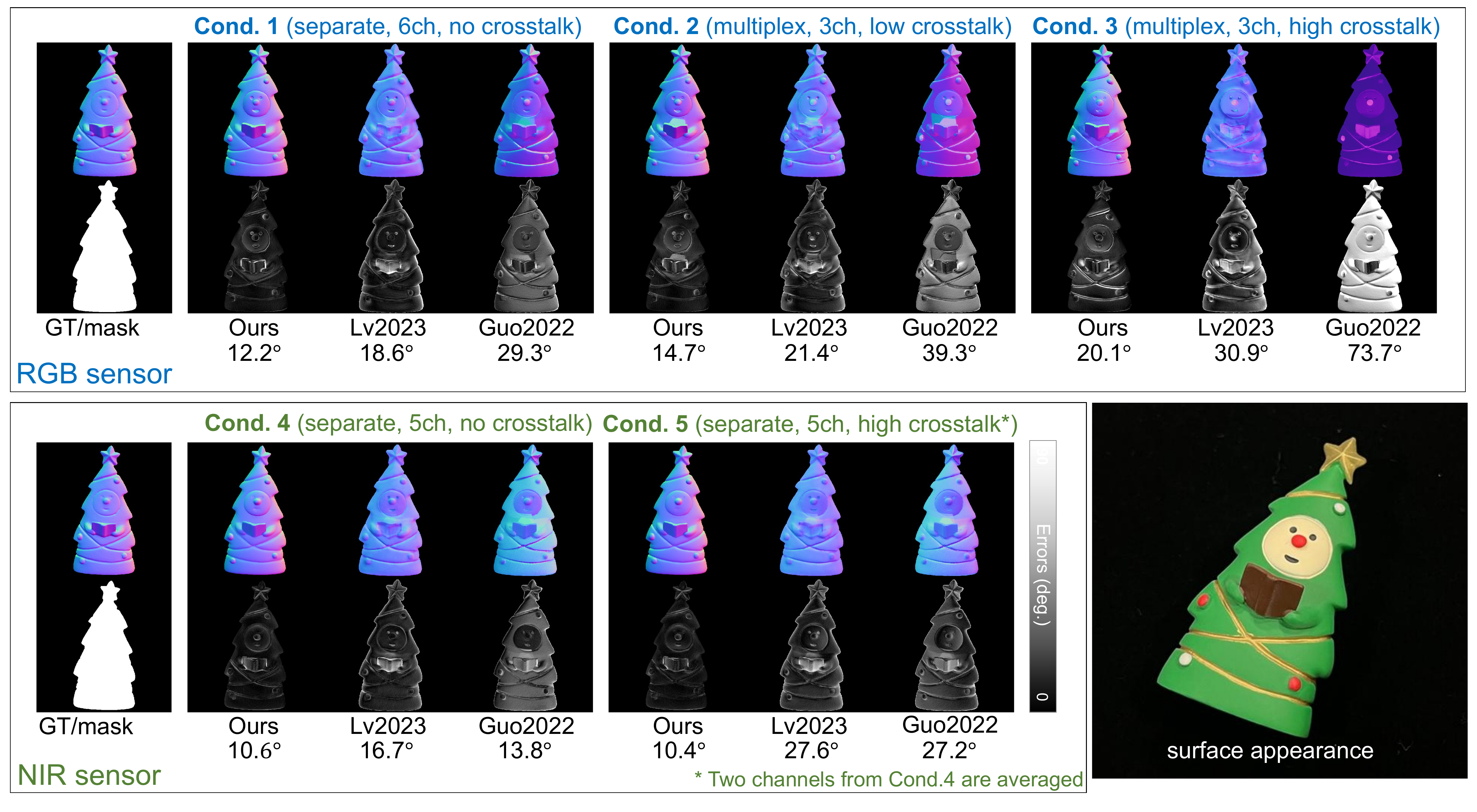}
	\end{center}
	\vspace{-15pt}
        \caption{Output of Object ID 14. A distinctive feature is the generally flat tree part, setting it apart from other shapes that are predominantly rounded. The proposed method typically incurs errors around 10 degrees. In contrast, Lv2023~\cite{Lv2023} and Guo2022~\cite{Guo2022} consistently face significant challenges, with errors exceeding 20 degrees. Notably, given the differing materials of the face and the tree, accuracy degradation in these areas is particularly pronounced in Conditions 2 and 3 for prior methods.}
	\label{fig:output14}
	\vspace{-5pt}
\end{figure}

In this section, we detail the main results of our paper. All inputs are illustrated in~\cref{fig:input1,fig:input2,fig:input3,fig:input4,fig:input5,fig:input6,fig:input7,fig:input8,fig:input9,fig:input10,fig:input11,fig:input12,fig:input13,fig:input14}, and outputs are illustrated in~\cref{fig:output1,fig:output2,fig:output3,fig:output4,fig:output5,fig:output6,fig:output7,fig:output8,fig:output9,fig:output10,fig:output11,fig:output12,fig:output13,fig:output14}. For each figure, we provide a detailed discussion about the object and the results.
\newpage

\section{Analysis on Lighting Distributions}
\label{sec:supp_analysis}
In this section, we investigate the impact of light source distribution on the proposed method. Unlike conventional temporally multiplexed problems, spectrally multiplexed photometric stereo requires consideration not only of the spatial distribution of light sources but also their spectral distribution. When the spectra of different light sources are similar, each sensor is more likely to be influenced by multiple sources (\ie, channel crosstalk). Consequently, the image captured by each sensor shifts from ones based on directional light sources to those based on spatially more complex lighting. Furthermore, when light sources are spatially proximate, the shading variations in images by different sources decreases, complicating the process of photometric stereo. On the other hand, in cases where light sources are in completely different directions or have no wavelength overlap, the system becomes more susceptible to the effects of cast and attached shadows, thus not necessarily improving performance. To investigate these non-trivial relationships, we varied the light source distribution both spatially and spectrally.

\begin{figure}[!t]
	\begin{center}
		\includegraphics[width=130mm]{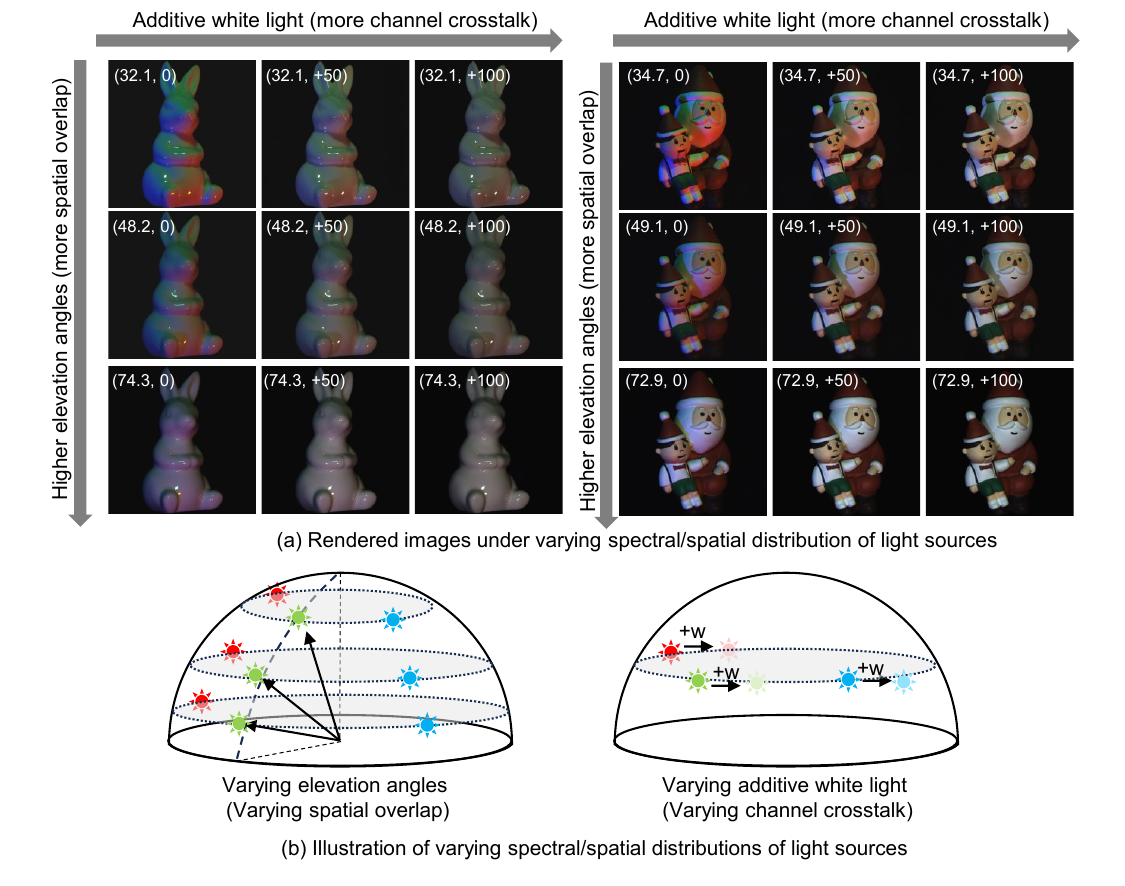}
	\end{center}
	\vspace{-10pt}
	\caption{Illustration of varying spatial/spectral lighting distribution. }
	\label{fig:limitation_data}
	\vspace{-5pt}
\end{figure}

The process of controlling the spatial/spectral light source distribution is illustrated in~\cref{fig:limitation_data}. In our experiment, three LED sources were used, with spectral multiplexing. For varying spatial distribution, the azimuth angle of the light sources was kept constant, while the elevation angle was manually adjusted in seven increments from roughly 0 to 90 degrees (See~\cref{fig:limitation_data}-(b)-left). At lower elevation angles, the light is projected horizontally relative to the object, while at higher angles, it is more vertically oriented, reducing the directional diversity between sources. Additionally, each LED light source was initially set with maximum brightness for one of the primary colors (Red, Green, Blue) and minimum for the others. Then, the intensity of white LEDs was incrementally increased, thereby shifting the spectrum from the initial state to include other spectral components (See~\cref{fig:limitation_data}-(b)-right). This alteration transformed the light source from a directional to a more spatially complex distribution. The intensity of the white LEDs was linearly increased from zero, reaching maximum at the eleventh increment. Here, we captured two different objects (ID11 and ID13) from our SpectraM14 dataset. The captured images are illustrated in~\cref{fig:limitation_data}-(a). 

\begin{figure}[!t]
	\begin{center}
		\includegraphics[width=130mm]{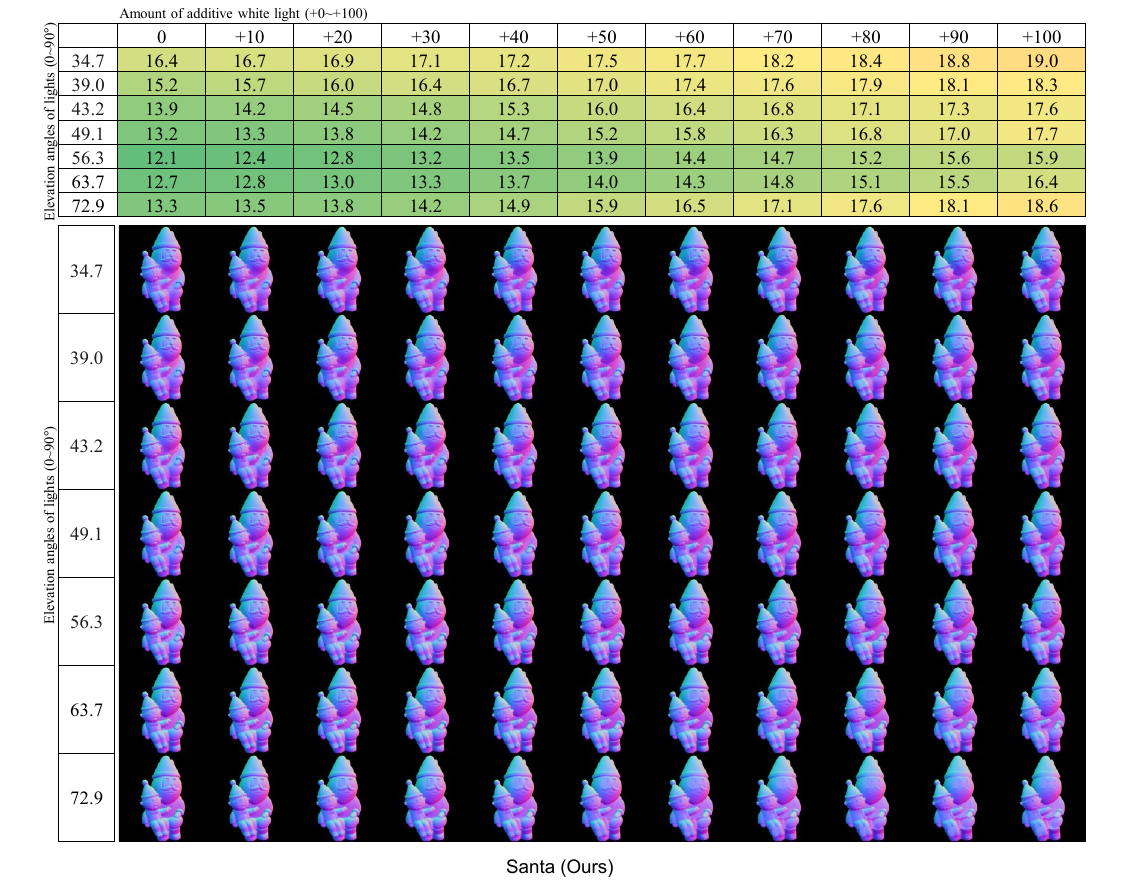}
	\end{center}
	\vspace{-10pt}
	\caption{Our method on Object ID 11 under varying lighting distributions.}
	\label{fig:limitation_ours_santa}
	\vspace{-5pt}
\end{figure}

\begin{figure}[!t]
	\begin{center}
		\includegraphics[width=130mm]{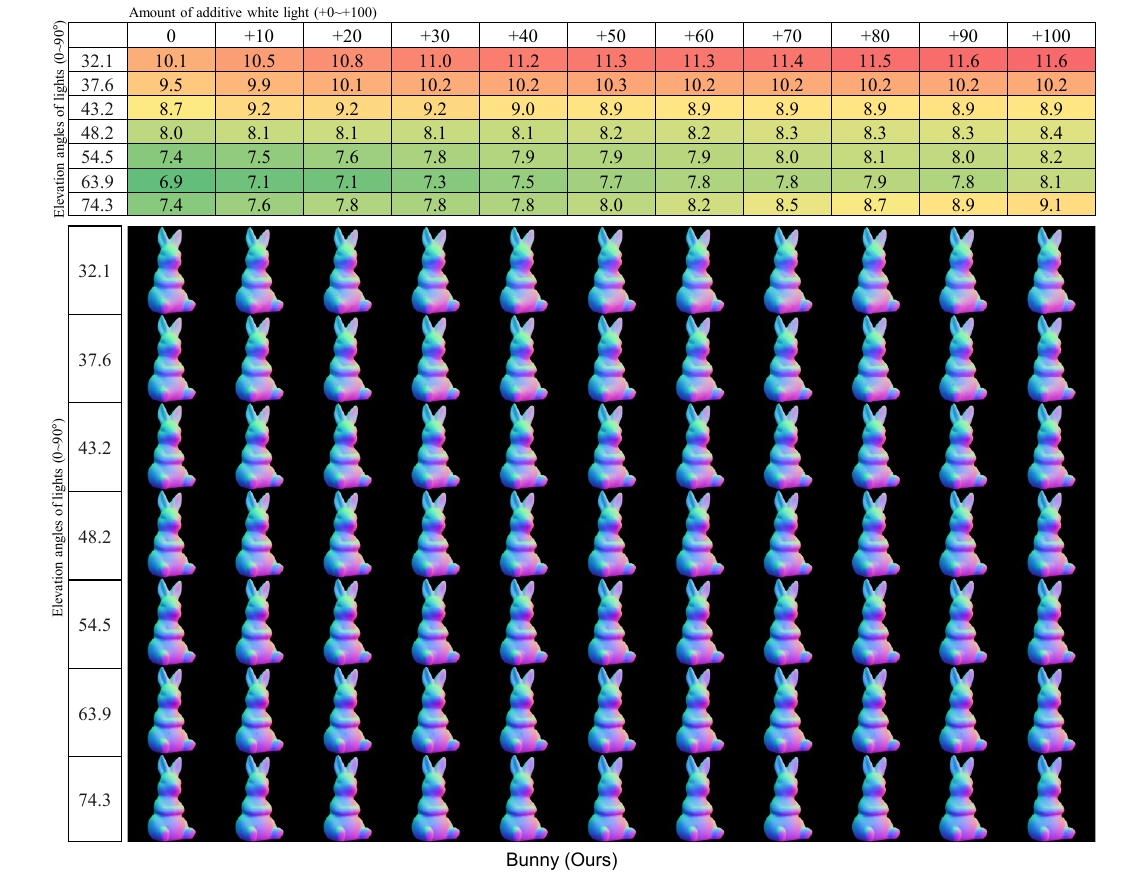}
	\end{center}
	\vspace{-10pt}
	\caption{Our method on Object ID 13 under varying lighting distributions.}
	\label{fig:limitation_ours_bunny}
	\vspace{-5pt}
\end{figure}

The results are illustrated in~\cref{fig:limitation_ours_bunny,fig:limitation_ours_santa}. The rows of the tables represent the elevation angles of the light sources (all three light sources share the same elevation angles), and the columns indicate the amount of additive white LED light. Initially, higher accuracy is observed at higher elevation angles of the light source, contrary to lower angles. However, it becomes clear that the optimal elevation angle is not the highest but slightly lower, indicating a performance decline when the overlap between light sources is maximized. Furthermore, the increasing MAE towards the right side of the table demonstrates that wavelength overlap between light sources also leads to performance degradation. Consequently, better accuracy is more likely when light sources are distinctly separated into RGB and multiplexed. This trend remains consistent even in scenarios like Object ID 11, which predominantly reflects red light. Although a general trend in spatial and spectral distribution is observable, the optimal combination of light source distribution varies with each object, complicating the identification of the best setup consistently. As a future research direction, automatically determining the optimal light source distribution poses a significant and worthwhile challenge.

\section{Application: Dynamic Surface Recovery}
\label{sec:supp_dynamic}
In our project website\footnote{https://github.com/satoshi-ikehata/SpectraM-PS-ECCV2024}, We demonstrate dynamic surface recovery by applying our method to each frame of a video captured under spectrally multiplexed illumination. We captured dynamic scenes using a Grasshopper3 RGB color sensor under multiplexed illumination and applied our proposed method to individual video frames. We employed four newer RGB 168 lights to create random non-uniform spatial and spectral distributions for illumination and captured the scenes against a black, low-reflective background (but not in the dark room). 

It's important to mention that we use a color sensor for the demonstration as we could not find a budget-friendly multispectral sensor in our vicinity suitable for dynamic surface reconstruction using spectrally multiplexed photometric stereo. Although multispectral or hyperspectral cameras provide broader channel measurements, their high cost (\eg, \$50,000 for the EBA NH7 as noted in~\cite{Guo2022}) and slow frame rates (less than 1fps) restrict their application in dynamic scenes. Consequently, optimizing for affordable sensors with fewer channels yet greater sensitivity and speed is crucial for practical applications.
\end{document}